\documentclass[AMA,STIX1COL]{WileyNJD-v2}

\usepackage{enumitem}
\usepackage{subcaption}
\usepackage{subfiles}
\usepackage{caption} 

\articletype{Research Article}%

\received{}
\revised{}
\accepted{}

\begin{document}

\title{Automatic Map Update Using Dashcam Videos}

\author[1]{Aziza Zhanabatyrova}

\author[1]{Clayton Frederick Souza Leite}

\author[1]{Yu Xiao}

\authormark{Zhanabatyrova \textsc{et al}}

\address{\orgdiv{Department of Communications and Networking}, \orgname{Aalto University}, \orgaddress{\state{Espoo}, \country{Finland}}}

\corres{Yu Xiao. Address: Konemiehentie 2, Espoo, Finland. Email: \email{yu.xiao@aalto.fi}}
\abstract[Summary]{Autonomous driving requires 3D maps that provide accurate and up-to-date information about semantic landmarks. Due to the wider availability and lower cost of cameras compared with laser scanners, vision-based mapping solutions, especially the ones using crowdsourced visual data, have attracted much attention from academia and industry. However, previous works have mainly focused on creating 3D point clouds, leaving automatic change detection as open issues. We propose in this paper a pipeline for initiating and updating 3D maps with dashcam videos, with a focus on automatic change detection based on comparison of metadata (e.g., the types and locations of traffic signs). To improve the performance of metadata generation, which depends on the accuracy of 3D object detection and localization, we introduce a novel deep learning-based pixel-wise 3D localization algorithm. The algorithm, trained directly with SfM point cloud data, can locate objects detected from 2D images in a 3D space with high accuracy by estimating not only depth from monocular images but also lateral and height distances. In addition, we also propose a point clustering and thresholding algorithm to improve the robustness of the system to errors. We have performed experiments on two distinct areas - a campus and a residential area - with different types of cameras, lighting, and weather conditions. The changes were detected with 85\% and 100\% accuracy in the campus and residential areas, respectively. The errors in the campus area were mainly due to traffic signs seen from a far distance to the vehicle and intended for pedestrians and cyclists only. We also conducted cause analysis of the detection and localization errors to measure the impact from the performance of the background technology in use.}


\maketitle


\section{Introduction} \label{sec:intro}

High-definition (HD) 3D maps are an important component in current autonomous driving solutions as they provide essential information for safe maneuvering in complex urban environments. Several mapping companies - such as HERE and TOMTOM - have already allocated diverse efforts to build, maintain and distribute HD maps. The creation of HD maps involves vehicles equipped with high-precision LiDAR sensors driving through different areas to collect point cloud data of the environment. Due to the high costs of LiDAR sensors, the vehicle fleet in charge of this task is limited to a few units and, therefore, considerably outnumbered by the number of roads. In addition, roads that were previously mapped need to be maintained, i.e., constantly monitored for changes and updated when necessary. Hence, the efficiency of building and maintaining HD maps is a bottleneck for autonomous driving. Compared with LiDAR sensors, visual sensors, such as dashcams, have a considerably lower cost; they are also widely available in the market and are easy to be utilized. Previous works \cite{8416714}\cite{7780814} have also shown the feasibility of creating accurate 3D point clouds from unordered images using structure from motion (SfM) techniques, which allows the input to be collected through crowdsourcing. However, due to the high computational complexity of SfM-based point cloud generation, it is too costly to frequently reconstruct point clouds from crowdsourced visual data. Therefore, it becomes essential to effectively detect and localize changes in the environment, and remap only the regions where the changes occurred. Currently, this is an under-explored topic and still remains an open issue.

We propose a pipeline based on SfM techniques for initiating and updating a semantic 3D map, with a focus on automatically detecting changes based on a comparison of the metadata (i.e. types and locations of traffic signs). Firstly, our pipeline generates a sparse point cloud that, combined with image-based semantic segmentation and object detection, enables the automatic generation of semantic map data - termed as metadata and consisted of types and 3D locations of traffic signs - assisted with a clustering algorithm that we devise. Secondly, it provides a novel method for utilizing SfM-based point clouds to train a deep learning model for online pixel-wise 3D localization from monocular RGB data. This method allows localizing traffic signs online with respect to the camera poses with high accuracy. Compared with our method, previous works\cite{9009796, lee2019big} provide only depth information, neglecting the lateral and height information necessary for 3D localization, and require LiDAR solutions to serve as annotations (ground truth). By utilizing data extracted from SfM-based point clouds to serve as ground truth, we discard the need for LiDAR annotations and additionally provide lateral and height distances. Thirdly, our system supports lightweight change detection by comparing the semantic map data with a thresholding algorithm to induce robustness to errors. With the multi-layer design, the dynamic map data representing temporary changes are stored on separate layers. 


We collected dashcam videos from two urban areas - campus and residential - in February and December, 2019, and used the data to evaluate system performance and conduct cause analysis of detection and localization errors. The change detection results in the residential area showed that our method was able to hit 100\% accuracy. In the campus area, the change detection accuracy was 85\%, where the errors were mainly due to traffic signs seen from a far distance to the vehicle and intended to pedestrians and cyclists only. We also provided an analysis of the errors of each component of the pipeline and how they can be improved to enable a more accurate change detection and localization. The advantages of our work are summarized as follows.

\begin{itemize}
    \item Our pipeline only requires RGB images and GPS information to generate the metadata (i.e. information on the location and types of traffic signs) of the initial state of the environment and to detect changes from subsequent iterations of the environment. Moreover, even though our proposal can benefit from multiple camera views, a sole camera view is enough to obtain high change detection accuracy. Also, our proposal has been evaluated under complex urban scenarios with different weather, lighting conditions, and camera types.
    \item Our pipeline reduces manual effort by eliminating the need for labels of changes and their locations since it does not utilize an end-to-end neural network trained on a dataset specifically designed for change detection\cite{transcd, scdmultiscale, diffnet}.
\end{itemize}

The rest of the paper is organized as follows. Section \ref{background} reviews the background technology used for building our solution. In Section \ref{overview} we provide an overview of the system architecture, followed by the detailed designs described in Section \ref{metadatagen} and Section \ref{changedetectionmethod}. The evaluation of these designs using the datasets presented in Section \ref{implementation} are discussed in Sections \ref{eval} and \ref{eval_our}. The related studies and future work are summarized in Sections \ref{related} and \ref{discussion} before concluding the paper in Section \ref{conclusion}.

\section{Background} \label{background}

This section introduces the technical background and prior works on structure from motion, semantic segmentation and object detection. 

\subsection{Structure from Motion}

Structure from Motion (SfM) has been used in our system to create 3D point clouds from 2D images. A typical SfM pipeline consists of 3 steps: feature extraction, feature matching, and bundle adjustment. The first step extracts highly distinctive and invariant features from the images, whereas the second step tries to match these features between image pairs. The matches are input for the last step that jointly produces optimal estimates of camera poses and locations of 3D points. Such a pipeline has been implemented in several SfM softwares, such as COLMAP \cite{7780814} and VisualSfM \cite{visualsfm}. 

How to apply SfM to reconstruct 3D point clouds from crowdsourced 2D images on a city scale was first demonstrated in \cite{agarwal2011building}. Another method that used SfM on a large scale for creating dense point clouds from stereo imagery was presented in \cite{alcantarilla2013large}, with the assumption that the stereo camera calibration parameters and the camera motion are known. COLMAP, presented in  \cite{7780814}, took one step further towards a general-purpose SfM system. We implement our system based on COLMAP since it offers improved robustness, accuracy, completeness, scalability, and has been released as an open-source software. Our system allows crowdsourced visual data as input, without camera calibration and motion information.


Regarding feature matching, several methods include different options, such as exhaustive, sequential, spatial, vocabulary-tree-based \cite{schonberger2016vote}, and custom feature matching. In the case of exhaustive feature matching, each image is matched against all others. Since it can result in an excessive processing duration, it is only indicated for small datasets of unordered images. Exhaustive matching is, for this reason, not utilized in this work. When the images are ordered in a sequence (such as when they are extracted from a video), sequential matching is recommended. In this matching method, the images are matched only against their closest ones. Hence, the benefit is a shorter processing duration. The spatial matching method utilizes spatial data - e.g. the GPS coordinates of all images - as additional input for faster processing. However, in our preliminary experiments, it often led to model fragmentation possibly due to inaccurate location information. Hence, it is discarded in this work. In vocabulary-tree-based matching, each image is matched against its visually nearest neighbors using a vocabulary-tree with spatial re-ranking, which is recommended for large image collections. Finally, custom matching is a method where the user defines a list of pairs of images to be matched. It is recommended for unordered datasets and requires manual labor. For the present work, we selected the vocabulary-tree-based, sequential and custom matching methods. Since the system is supposed to work with a large amount of unordered crowdsourced data, the vocabulary-tree matching proves to be more efficient and accurate than its counterparts in our experiments. However, it has been observed that when the desired area to be mapped does not have enough visual features in some portions of it, it is more relevant to incorporate the sequential matching method. This is because it avoids the matching between images located distant from each other, which can result in an inaccurate reconstruction.

The SfM pipeline outputs a 3D point cloud with random scale and orientation. Therefore, geo-registration (or geo-referencing) - which consists of a similarity transformation - is typically performed afterwards to re-scale and align the model with respect to the real world. The geo-registration function provided by COLMAP requires the real-world Cartesian coordinates of at least three distinct images uniformly spaced across the map. The simplest way to do this is by utilizing positioning services - e.g. GPS or even real-time kinematic (RTK) positioning - during the visual data collection to obtain geodetic coordinates for each image and then transforming these coordinates into Cartesian ones. In case positioning services are not available during the data collection, a solution is to manually select three locations in the map and obtain their GPS coordinates with the help of tools, such as Google Maps and, transform their GPS coordinates into Cartesian ones. 

COLMAP does not provide methods for change detection or automatic point cloud update. However, it provides functions for deleting images and their related 3D information from a point cloud, as well as registering new images into an existing point cloud. When a change in the scene has been detected and localized, it is possible to utilize these functions to delete the corresponding 3D points from the existing point cloud, and then register the images capturing the new scene into the point cloud. Therefore, the focus of this paper is placed on change detection and localization rather than the implementation of the point clouds update.  

\subsection{Semantic Segmentation and Object Detection}
\label{sem_objdet}
Semantic segmentation and object detection are computer vision tasks employed to detect objects in an image and assign to them an appropriate class label. In the case of semantic segmentation, a class label is assigned to each pixel in an image. However, multiple objects of the same class are not recognized as separate objects unless a more complex form of semantic segmentation - i.e. instance semantic segmentation - is performed. Differently from semantic segmentation, object detection provides an individual bounding box around each detected object. In case the boundaries of objects are not precisely defined, it may reduce the accuracy of the following object localization step. For example, if a segment that represents a traffic sign by accident covers part of a building from the background, the location of the traffic sign may be set to the location of the building, which is away from the ground truth. 

Our system combines both approaches as object detection spots different instances of the same object individually while semantic segmentation gives a more precise boundary around the detected object. The algorithms are later used to detect traffic signs in images on a pixel-wised level and localize them in the 3D map with the other methods discussed in Section \ref{overview}. By projecting the points of the point cloud back to the images that generated the SfM model - thus transforming the points back into pixels - and running the objection detection and semantic segmentation neural networks on these images, we can classify each individual pixel and consequently its corresponding 3D point from the point cloud. In this work we use the deep learning-based semantic segmentation solution, Seamseg, proposed by Porzi \textit{et al.}\cite{8954334}. As for the object detection we use method proposed by Xin \textit{et al.} \cite{ssdresnet}, (SSDResNet), available in the TensorFlow Object Detection API. The details of the both methods are given in Section \ref{metadatagen} and Section \ref{eval}. We do not utilize any instance segmentation method as it has not been possible to find any appropriate open-source dataset for training. Also, creating such a dataset would demand an extensive amount of time.




\section{System Overview} \label{overview}

\begin{figure*}[bt!] 
  \centering
    \includegraphics[width=\textwidth]{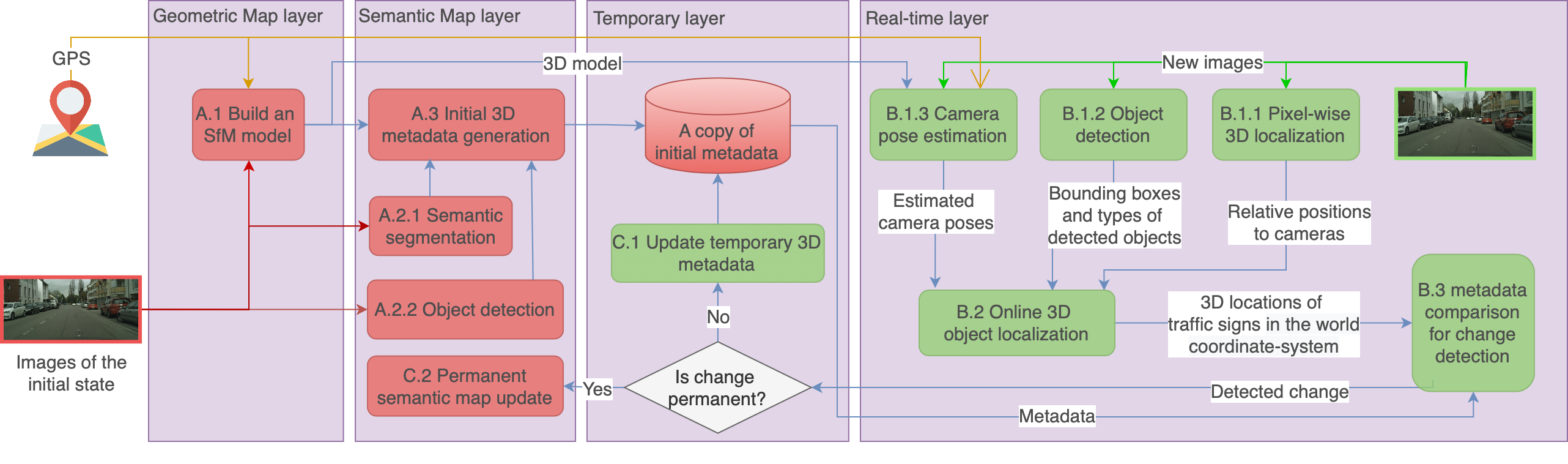}
    \caption{Architecture of a multi-layer map and the pipeline of creating and updating the map based on crowdsourced visual data. Blocks in red demonstrate the generation of initial map, whereas those in green show the update of the map using newly collected data.}
\label{fig:system}
\end{figure*}

HD maps commonly have multiple layers, where each layer serves a specific purpose and has a distinct structure \cite{8751934}. We propose a pipeline for creating and updating a four-layer 3D map from crowdsourced dashcam videos - as illustrated in Figure \ref{fig:system}. The 4 map layers consist of 1) a geometric map layer which stores the raw point cloud generated by the SfM, 2) a semantic map layer which stores the metadata with object semantics, 3) a real-time layer for recurrent change detection, and 4) a temporary layer that processes temporary changes. The layers are described in more detail in the following subsections.

The pipeline consists of two stages, the generation of initial map and the update of the map using newly collected data. The initial stage creates static information for the geometric and semantic map layers, while the update stage extracts dynamic map information from visual data in real time in order to detect changes in the environment by comparing the its current state to the state stored in the semantic map layer. 


\subsection{Geometric Map Layer} 

The pipeline starts from Step A.1 which reconstructs a 3D point cloud from 2D images using SfM. The output of the SfM pipeline, as described in Section \ref{background}, includes the 3D points, the camera pose of each image registered into the point cloud, as well as its camera extrinsic matrix $\mathbf{T}$ and camera intrinsic matrix (also called calibration matrix) $\mathbf{K}$. Assuming an ideal pinhole camera model, the camera projection matrix $\mathbf{P}$ is given as $\mathbf{P} = \mathbf{K}\mathbf{T}$. All this information is stored in the Geometric Map Layer as simple text files. The geometric map data remains unchanged unless there is a significant change in the road infrastructure. 

\subsection{Semantic Map Layer} 



The images used for creating the point cloud are also utilized for detecting objects in the environment by the use of a deep learning-based semantic segmentation (at Step A.2.1) and an object detection (at Step A.2.2) neural networks. The output of Step A.2.1 includes pixel-wise semantic segmentation predictions, while that of Step A.2.2 includes bounding boxes of detected objects (e.g. traffic signs) along with their corresponding classes. These outputs are used as input of Step A.3 by segmenting the point cloud, where each segment of the point cloud represents a class of object. The semantic information of static objects is stored as metadata of the point cloud at the Semantic Map Layer in the following text-based format: (latitude, longitude, class name, object color, date detected). The GPS coordinates of each object is obtaining by geo-registering the point cloud. In our case, since we are only focusing on traffic signs, the class name of the static object includes the information concerning the type of the traffic sign. A copy of the 3D metadata generated at Step A.3 is saved also at the Temporary Layer. 


\subsection{Temporary Layer} 

Initially the Temporary Layer stores an exact copy of the metadata generated at Step A.3. When changes in the environment are detected, they are stored in the Temporary Layer by modifying the metadata to represent the current state of the environment. The metadata of the Semantic Map Layer is only updated (refer to Step C.2) when a change becomes permanent. We deem that a change is permanent if it has been observed by a certain number of vehicles of the crowdsourcing application throughout a certain number of days. These values are still to be defined in future work.

\subsection{Real-time Layer} 

The real-time layer takes care of change detection in three steps. After the data have been collected in real-time, they are passed for the latter processing which - depending on the available computing resources - may not be performed in real-time. The first step is to run three different algorithms on input images: 1) the pixel-wise 3D localization (Step B.1.1), 2) the object detection (Step B.1.2), and 3) the camera pose estimation (Step B.1.3). These algorithms can be run in parallel since they do not depend on each other. Step B.1.1 calculates the relative 3D position to the camera for each pixel of the image. Step B.1.2 outputs the bounding boxes and classes of objects detected from each image. Step B.1.3 estimates the camera pose using the SfM model created in Step A.1. Since the point cloud is geo-referenced, the camera poses are converted automatically into coordinates in the world coordinate system (WCS). The second step (i.e. Step B.2) is to calculate the 3D locations of the detected objects in the WCS and to generate new 3D metadata for the point cloud accordingly. The third step (i.e. Step B.3) is to compare the newly generated metadata with the latest version stored at the temporary layer. If there exists any difference, a change is detected. The copy of metadata at the temporary layer will be updated accordingly (refer to Step C.1).

 In our system, deep learning has been applied for implementing the semantic segmentation algorithm of Step A.2.1, as well as the object detection algorithm present in Step A.2.2 and Step B.1.2. These deep learning models are trained independently before the initial stage. They do not need to be retrained unless the application or domain has changed. For example, if a model has been trained to classify traffic signs in one country, to work in another country - i.e. a new domain - it may need to be fine-tuned on a dataset including specific traffic signs of that country. 

\begin{figure}
\begin{subfigure}{.49\textwidth} 
  \centering
    \includegraphics[width=.99\textwidth]{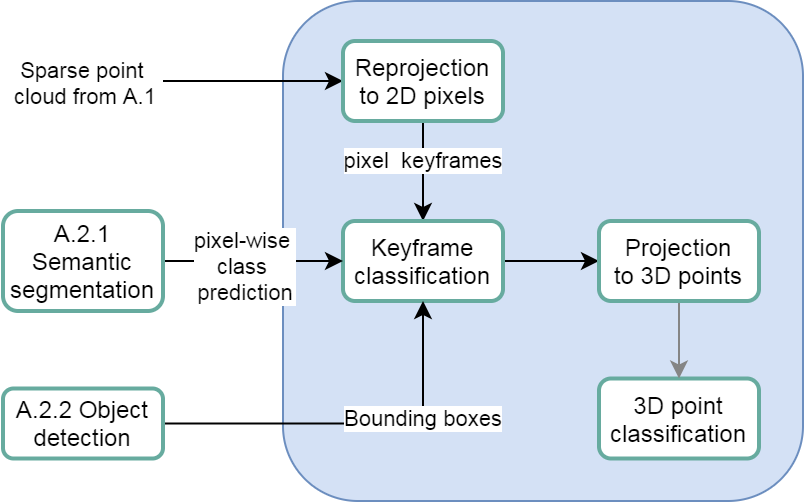}
    \caption{}
\label{fig:meta}
\end{subfigure}\hspace{10mm}
\begin{subfigure}{.45\textwidth}
  \centering
    \includegraphics[width=.99\textwidth]{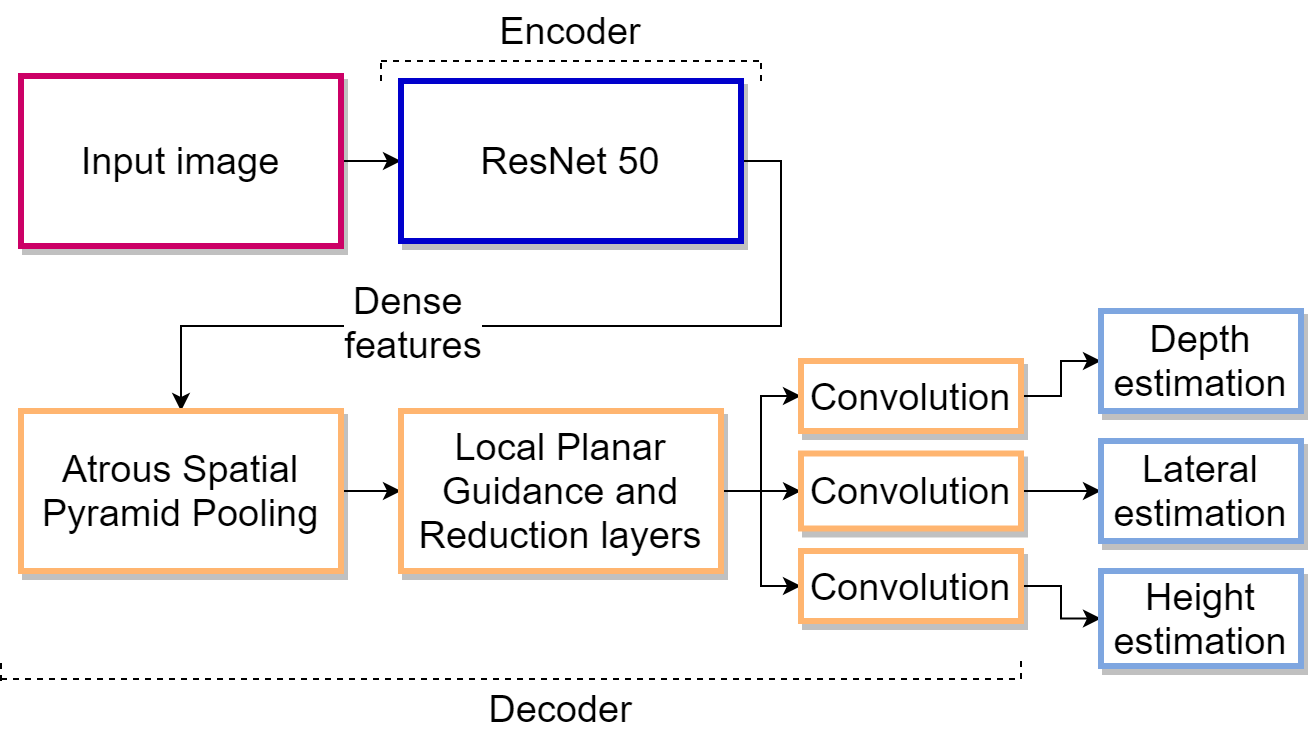}
    \caption{}
\label{fig:3DRGBarch}
\end{subfigure}
\caption{(a) Workflow of the initial 3D metadata generation (Step A.3 in Figure \ref{fig:system}); (b) Network architecture of pixel-wise 3D localization based on BTS \cite{lee2019big}. }
\end{figure}

\begin{figure*}[bt!]
\centering
\begin{subfigure}{.33\textwidth}
  \centering
  \includegraphics[trim=0 20 0 25,clip,width=.99\textwidth]{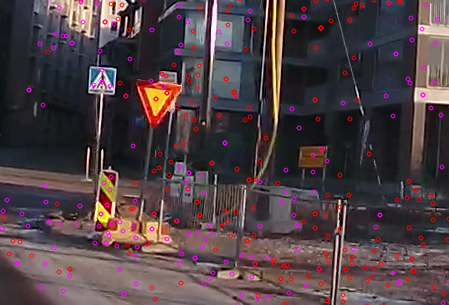}
  \caption{2D image keypoints}
  \label{fig:semseg1}
\end{subfigure}%
\begin{subfigure}{.33\textwidth}
  \centering
  \includegraphics[trim=0 39 0 0,clip,width=.99\textwidth]{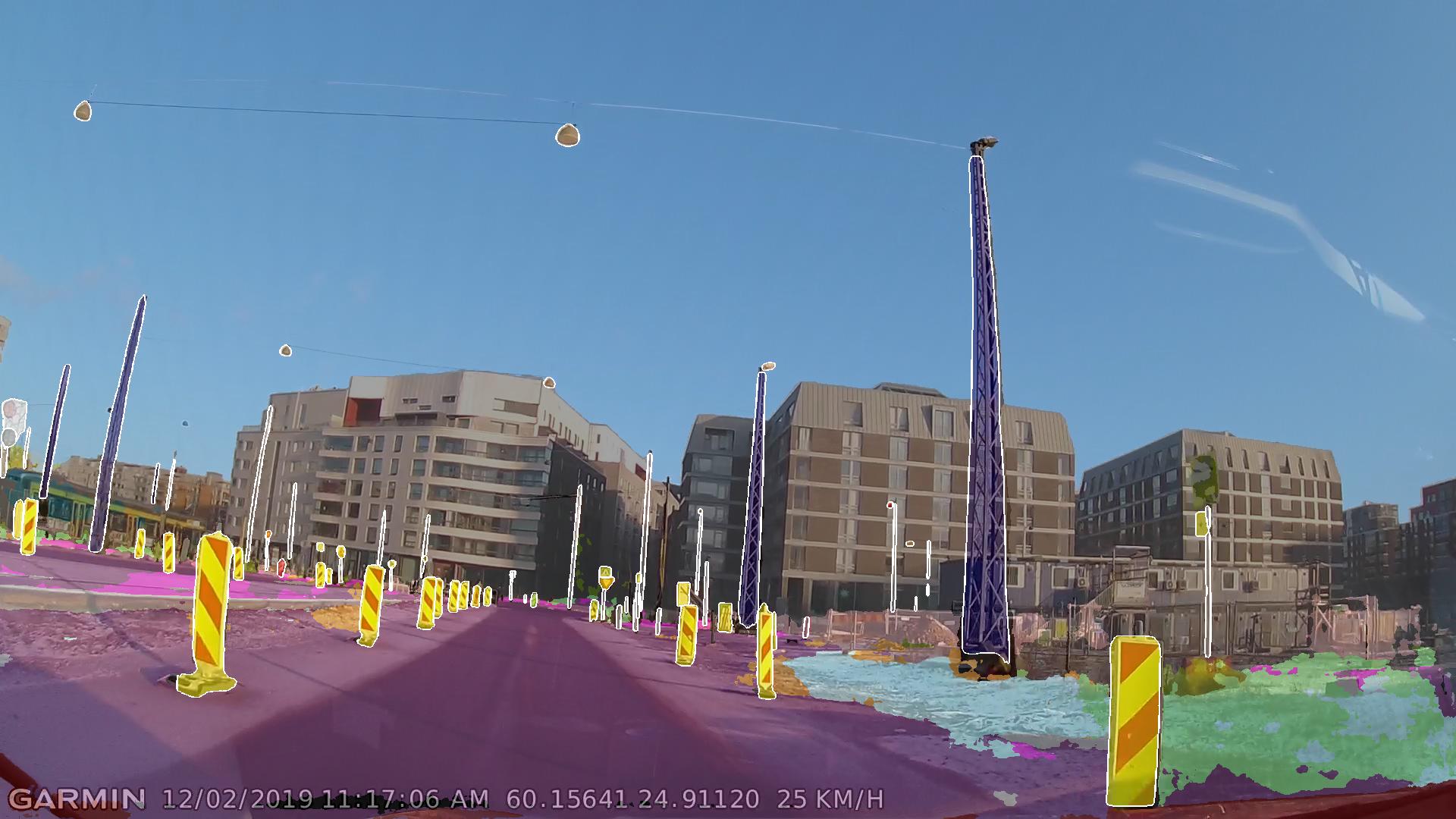}
  \caption{Semantic segmentation}
  \label{fig:semseg2}
\end{subfigure}
\begin{subfigure}{.33\textwidth}
  \centering
  \includegraphics[width=.99\textwidth]{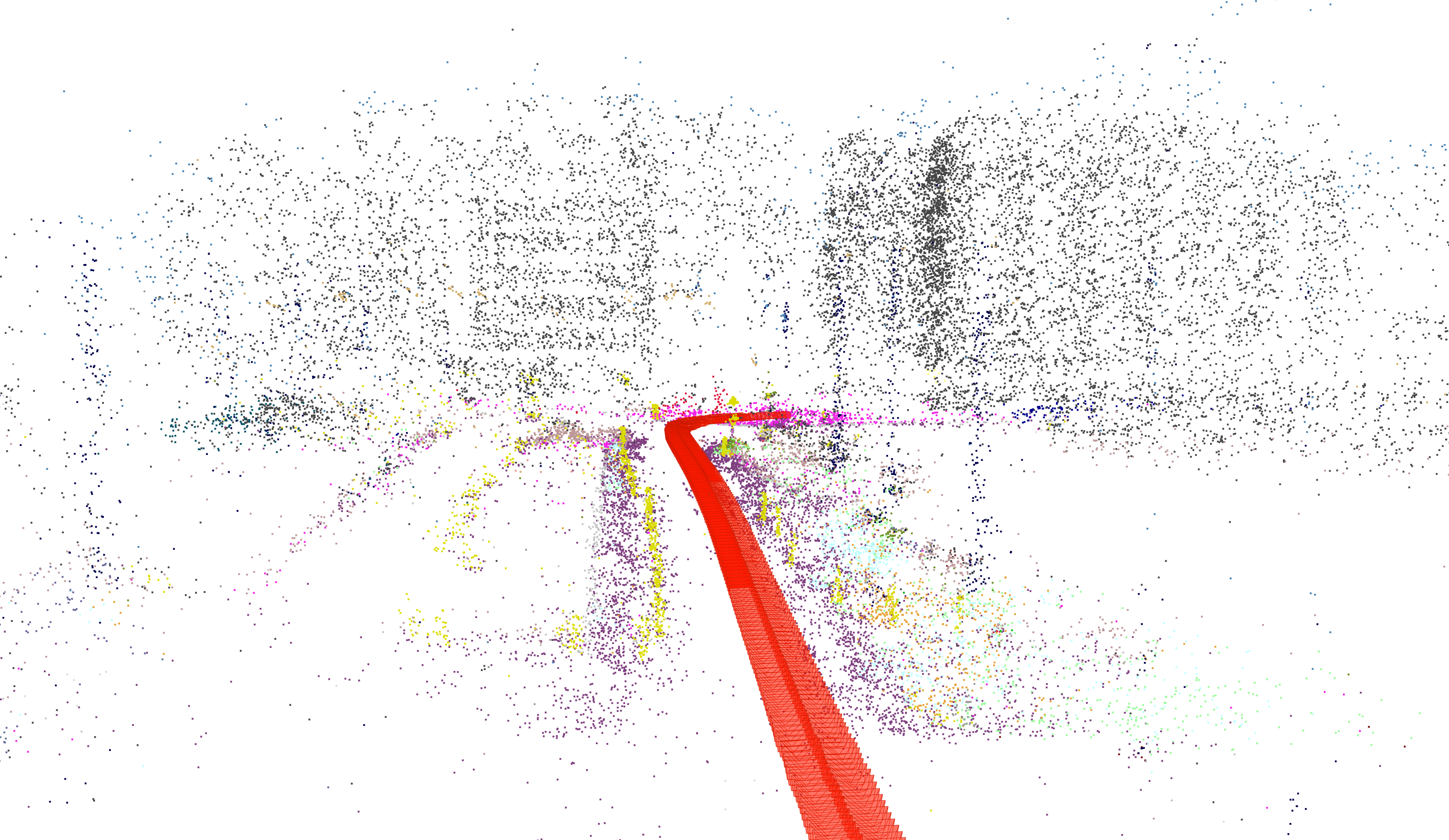}
  \caption{Point cloud segmentation}
  \label{fig:semseg3}
\end{subfigure}
\caption{(a) An example image and keypoints extracted from the image. The pink keypoints represent the ones utilized for 3D point generation, while the red ones have not been registered into the point cloud. Note that the image has been cropped from the original one to highlight the keypoints. (b) Visualization of the results of the semantic segmentation by generating an image where each pixel takes the color that represents its class of maximum probability. The input image is also overlaid to facilitate the visualization. (c) Results of the point cloud segmentation. The camera poses predicted by the SfM pipeline are shown in red color in Fig. \ref{fig:semseg3}.}
\label{fig:semseg}
\end{figure*}

\section{Initial 3D Metadata Generation}\label{metadatagen}

\begin{algorithm}[t]
Define distance threshold for clustering traffic signs $T_D$ \\
Create sparse point cloud from the images of the region (Step A.1) \\
Geo-register the point cloud (Step A.1) \\
\textbf{for} each image of the region \textbf{do}\\
   \hspace*{0.45cm}Obtain the semantic segmentation of $I$ (Step A.2.1) \\
   \hspace*{0.45cm}Run the object detection neural network on $I$ (Step A.2.2)\\
\textbf{for} each point $P$ in the sparse point cloud \textbf{do}\\
   \hspace*{0.45cm}Project $P$ on the set of images that observe it to gather a set of pixels \\
   \hspace*{0.45cm}Average the semantic segmentation predictions on the set of pixels \\
   \hspace*{0.45cm}Colorize $P$ according to the average semantic prediction \\
   \hspace*{0.45cm}\textbf{if} the average semantic prediction of $P$ is a traffic sign\\
   \hspace*{0.9cm}Obtain the traffic sign type of $P$ from the object detection neural network \\
   \hspace*{0.9cm}Find the GPS location of $P$ based on the geo-registration of the point cloud \\
   \hspace*{0.9cm}Save $P$'s GPS location and its traffic sign type\\
Run K-means clustering algorithm with number of clusters set to the minimum integer value that satisfies Eq. \ref{eq:meta_constraint} \\
Save in a text-based file format the GPS location of each cluster as well as its traffic sign type 
  \caption{Our algorithm for initial metadata generation.}
   \label{alg:meta_gen}
\end{algorithm}



This section describes the process of the initial 3D metadata generation (Step A.2) as illustrated in Figure \ref{fig:meta}. The images from the initial state of the environment used for the sparse point cloud generation (Step A.1) are segmented using the semantic segmentation (Step A.2.1) neural network proposed by Porzi \textit{et al.} \cite{8954334} - whose architecture consists of 50 layers built on the ResNet convolutional neural network \cite{resnet}. The output of the neural network represents a pixel-wise semantic prediction, visualized in Figure \ref{fig:semseg2}, with 65 urban street output classes. It consists of a text-based file that stores the probability of the 65 classes for each pixel in the input image, thus being similar to the input image itself with the exception that there are 65 channels instead of the 3 RGB channels. 



For each 3D point in the sparse point cloud, COLMAP provides a list of $k$ images that observe it as well as the pixel coordinates - named image keypoints and shown as pink dots in Fig. \ref{fig:semseg1} - where it is observed. Hence, given a 3D point, we obtain $k$ semantic segmentation predictions corresponding to the images that observe the point. The predictions consist of a probability - or confidence level - for each class considered in the semantic segmentation network. The 3D point is then assigned the class of highest average probability value. The files that store the point cloud are then modified to include the segmentation by colors.  Figure \ref{fig:semseg3} shows the semantically segmented point cloud with different colors for points of distinct classes. 

The results of the semantic segmentation do not include the type of traffic sign. Therefore, we utilize the object detection neural network (Step A.2.2) to obtain this information and include it to the point cloud.  The reasons for combining semantic segmentation with object detection are two-folded. Firstly, combining the class predictions of the two methods improves the generalization - this is known as ensemble learning \cite{ensemblelearning}. Secondly, the bounding boxes surrounding the detected objects often contain lots of space and may cover some objects that belong to different classes. Since semantic segmentation provides pixel-wise prediction, by calculating the intersection of each bounding box and the corresponding segment, more precise boundaries of objects can be obtained. Section \ref{obj_det_eval} discusses the choice of the objection detection algorithm as well as its training. Additional details on the semantic segmentation is given in \ref{sec:sson2dimages}.

After the information on the type of each traffic sign is stored into the point cloud, all points that represent traffic signs are extracted from the point cloud into a text-based file that, for each point - stores its GPS location and its traffic sign type. Since a traffic sign can consist of a multitude of points, a clustering algorithm is executed to group all points that pertain to the same traffic sign into a single point located in the center of the cluster. The clustering is performed with the use of K-means clustering. We denote as $c$ the number of clusters given as input to the K-means clustering algorithm, $D_i$ as the distance from the center of the cluster $i$ to the furthest point pertaining to the cluster (note that $D_i$ is a function of $c$), and $T_D$ as the maximum allowed distance $D_i$. We propose to set the number of clusters equal to the minimum integer value above zero that does not violate the following constraint \ref{eq:meta_constraint}. The output of the clustering algorithm is the metadata - i.e. a set of points representing each of the traffic signs including their GPS location and type. The choice of $T_D$ is discussed in Section \ref{eval}. Algorithm \ref{alg:meta_gen} summarizes the process of generating the metadata.
\begin{equation}
    \text{max}_i D_i < T_D
    \label{eq:meta_constraint}
\end{equation}

\section{Change Detection} \label{changedetectionmethod}

\begin{algorithm}[t]
Define range of distance from camera to traffic signs $U$ \\
Define radius to search for matching traffic signs in the metadata $R$ \\
\textbf{for} each new image $I$ \textbf{do}\\
   \hspace*{0.45cm}Obtain $I$'s camera pose estimation (Step B.1.1) \\
   \hspace*{0.45cm}Run the object detection neural network $I$ (Step B.1.2) \\
   \hspace*{0.45cm}Run the modified BTS neural network $I$ (Step B.1.3) \\
   \hspace*{0.45cm}\textbf{for} each detected traffic sign $T_s$ \textbf{do}\\
   \hspace*{0.9cm}\textbf{if} distance from the camera to $T_s$ is within a range $U$\\
   \hspace*{1.35cm}Find and store the GPS location of $T_s$ (Step B.2) \\
   \hspace*{1.35cm}Average the GPS location of $T_s$ over previous stored measurements \\
   \hspace*{1.35cm}Find closest traffic sign $T_s'$ of same type as $T_s$ in the metadata (Step B.3) \\
   \hspace*{1.35cm}\textbf{if} $T_s'$ does not exist or distance from $T_s$ to $T_s'$ is greater than $R$ (Step B.3) \\
    \hspace*{1.75cm}Report change on the Temporary Map Layer (Step B.3) \\
  \caption{Our change detection algorithm based on the utilization of the camera pose estimation, object detection, and modified BTS neural network.}
   \label{alg:cd}
\end{algorithm}

The newly collected video/images are processed in 3 steps (i.e. Step B.1.x, Step B.2, and Step B.3) online to detect potential changes in the environment based on a comparison with the metadata. Step B.1.1 consists of obtaining the camera pose with respect to the WCS. The same method for object detection as utilized in Step A.2.2 is also used here in Step B.1.2 to produce bounding boxes of traffic signs and to identify their type. In Step B.1.3 applies monocular depth, lateral and height distance estimation to gather the relative positions with respect to the camera of all pixels in the image. Since these steps are independent of each other, they can be executed in parallel.

Step B.2 utilizes the estimated camera poses, the bounding boxes with the types of the detected traffic signs, and the relative pixel positions to the camera. These three inputs are processed to obtain the 3D locations of the detected traffic signs in the WCS. With this, for each traffic sign, we search for matching traffic signs of same type within a specified radius in the copy of metadata stored in the temporary layer. If there is a mismatch (e.g. there did not exist such a traffic sign earlier), a change is reported and the copy of metadata at the temporary layer can be updated accordingly. Algorithm \ref{alg:cd} summarizes the change detection procedure.

In this section, we explain in details how to estimate camera poses (Step B.1.1), how to calculate pixel-wise relative location with respect to camera (Step B.1.1), and how to convert it into 3D object locations (Step B.2) in WCS (e.g. GPS locations of traffic signs).





\subsection{Camera Pose Estimation} 
\label{sec:camera_pose_estimation}


In our pipeline, the camera pose estimation (i.e. estimation of camera position and orientation in the WCS) is performed in two distinct manners depending on the presence of traffic signs. In the first method, given an image $I_t$ at a time instant $t$, if a traffic sign is detected in it, the position and orientation of its camera (with respect to the WCS) is obtained by registering it into the point cloud using SfM with the custom feature matching \cite{7780814}. In custom feature matching, the image pairs to be matched can be defined in a custom manner. In our case, we opt to match the image $I_t$ with the nearest image (in terms of Euclidean distance calculated with GPS coordinates) that was utilized to build the point cloud - $I'_t$. Using the described custom feature matching significantly reduces the possibility of matching failure in places where the amount of visual features is insignificant.



In the second method, the camera position of $I_t$ is obtained directly from its GPS coordinates, whereas the orientation of $I_t$ is calculated by assuming that its orientation with respect to $I_t'$ equals that of $I_{t-1}$ with respect to $I'_{t-1}$. In practice, this assumption means that the camera pose is always fixed with respect to the car - i.e. there is no relative movement between the car and the camera - and the car follows the exact orientation of the road reconstructed in the point cloud. Since in general this assumption holds, this represents a good approximation for finding the camera pose. Mathematically, this can be expressed with the following equation:

\begin{equation}
\label{eq:cpe}
    R_{t} = R_{t-1} \cdot R'^T_{t-1} \cdot R'_{t}
\end{equation}

\noindent
where $R_t$ is the rotation matrix of the image $I_t$ with respect to the absolute reference frame - which represents its camera orientation - and similarly for $ R_{t-1}$, $R'^T_{t-1}$ and $R'_{t}$. The subscript $T$ denotes the transpose operation.

The first method is utilized when the orientation of $I_{t-1}$ is not known or periodically to avoid the accumulation of errors of the second method. 
The second method is a significantly computationally cheaper alternative to the first method with lower but still good accuracy. Compared with deep learning based camera pose estimation, such as PoseNet \cite{kendall2015posenet}, obtaining camera poses from the SfM pipeline typically requires smaller computational costs and excludes the need for an extensive training dataset. The evaluation of the camera pose estimation is given in Section \ref{sec:camera_pose_estimation}. 

\subsection{3D Object Localization}


For each image, we first apply object detection to create a bounding box that covers the traffic sign in question (step B.1.2). After that, we select the pixel at the center of the bounding box to represent the location of the object, and calculate its 3D coordinates relative to the camera following a process called pixel-wise 3D localization (step B.1.1). The process is implemented using convolutional neural network (CNN), as described in Fig. \ref{fig:3DRGBarch}. We use the state-of-the-art network, BTS, for monocular depth estimation \cite{lee2019big}. Our method can also work with other CNN-based monocular depth estimation methods. 

Originally the BTS network produces a single channel, which is the pixel-wise depth prediction. We have modified the output layer to produce pixel-wise output with three channels representing x, y, and z coordinates in a 3D space. To train the network, we create a labelled image set including all the images used for creating the point cloud at Step A.1. The three channels of the labels in this case represent the x, y, and z coordinates of a 3D point (in the SfM point cloud) with respect to the camera. Note that since the point cloud is sparse, the images are also sparsely labeled - i.e. some pixels may not have annotations. During the training, these non-labeled pixels are masked to not influence in the minimization of the loss function. For each input image, the SfM pipeline outputs an estimated camera pose, the coordinates of the 3D points, and the 2D keypoints that have been used to generate those 3D points. Also due to the nature of the sparsity in the dataset, we opted to fine-tune the model instead of training it from scratch. In the fine-tuning, the encoder and the early layers of the decoder had their weights frozen.  

\begin{figure*}[th]
\centering
\begin{subfigure}{.25\textwidth}
  \centering
  \includegraphics[width=.99\textwidth]{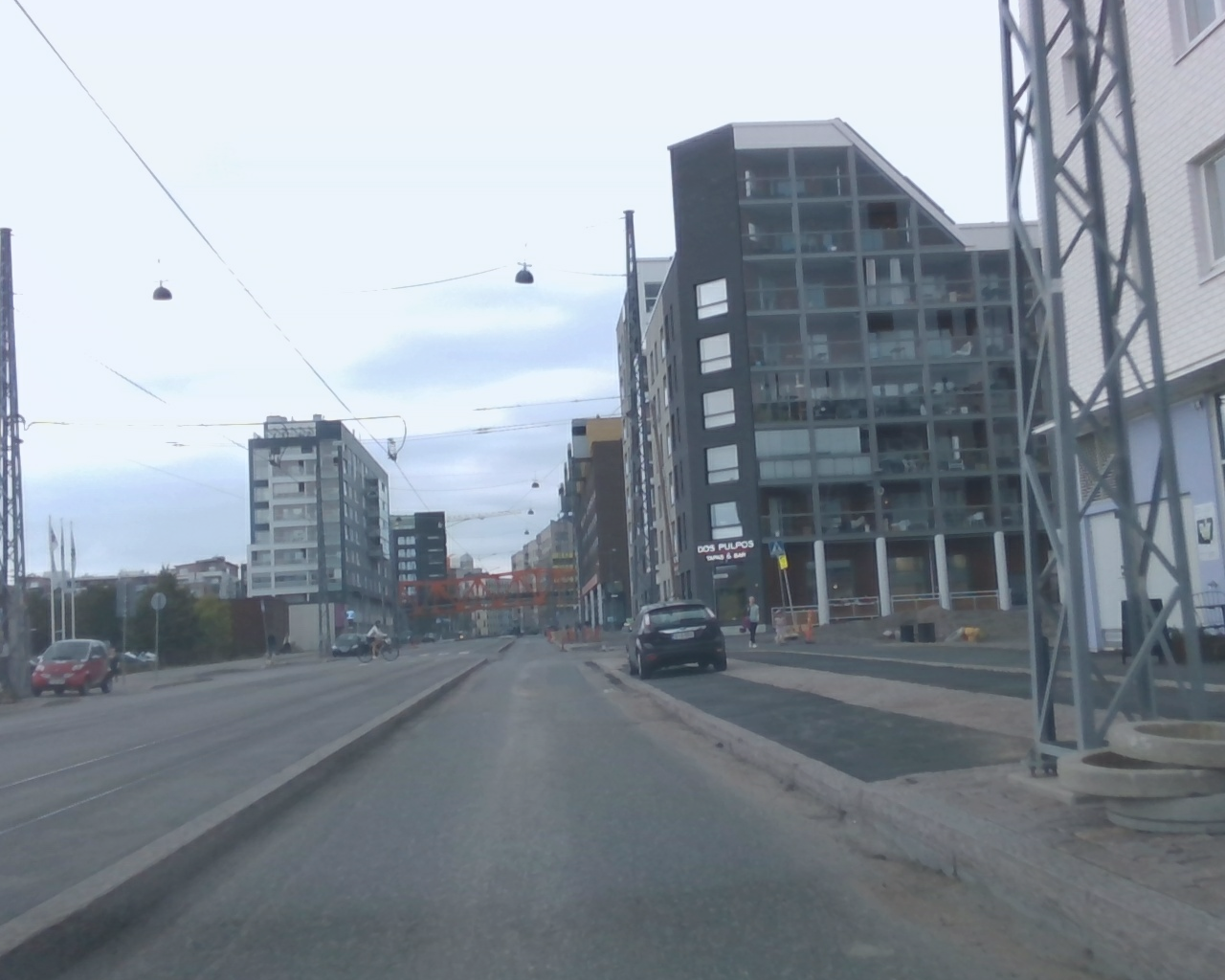}
  \caption{Image}
  \label{fig:sub1}
\end{subfigure}%
\begin{subfigure}{.25\textwidth}
  \centering
  \includegraphics[width=.99\textwidth]{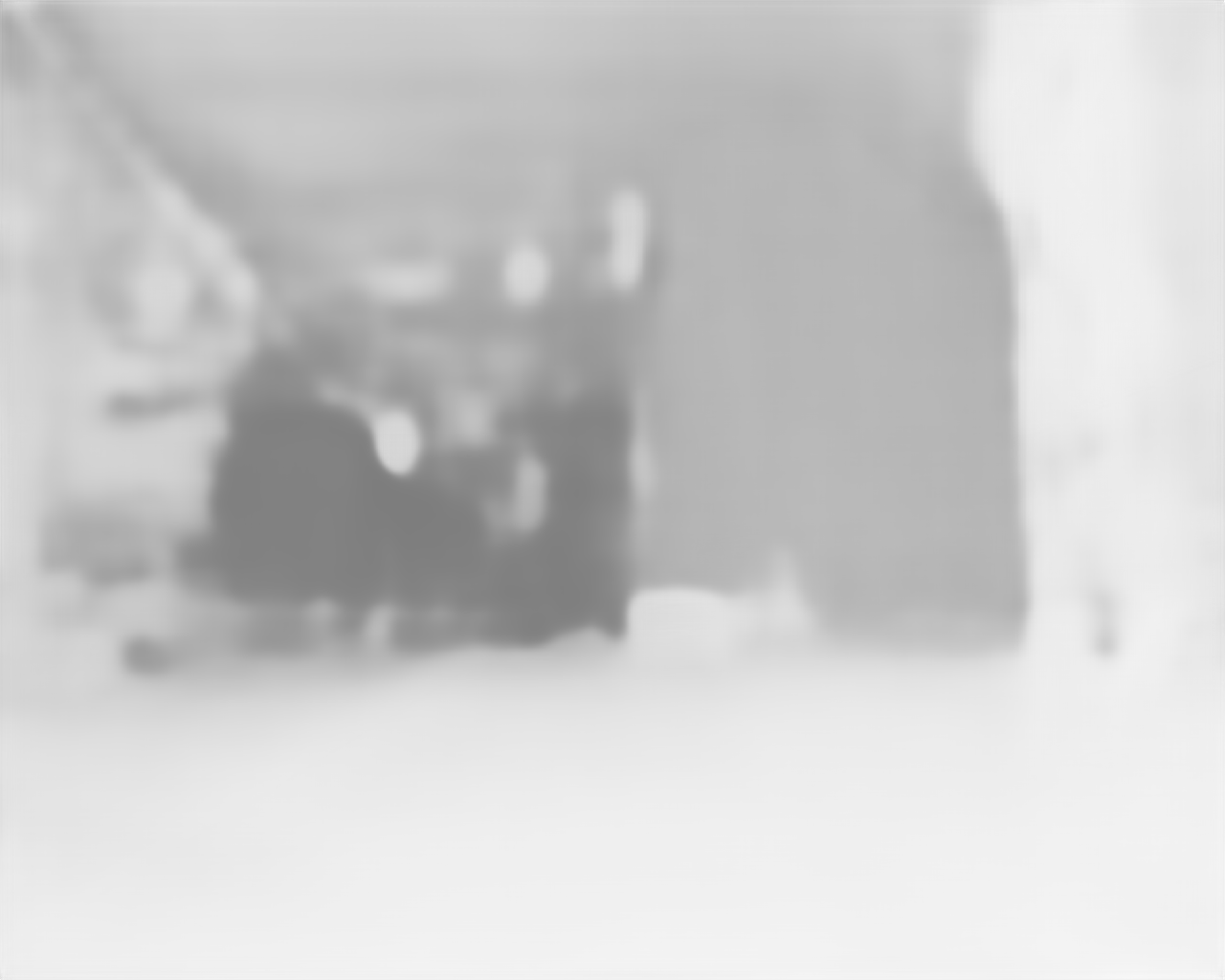}
  \caption{Depth}
  \label{fig:sub2}
\end{subfigure}%
\begin{subfigure}{.25\textwidth}
  \centering
  \includegraphics[width=.99\textwidth]{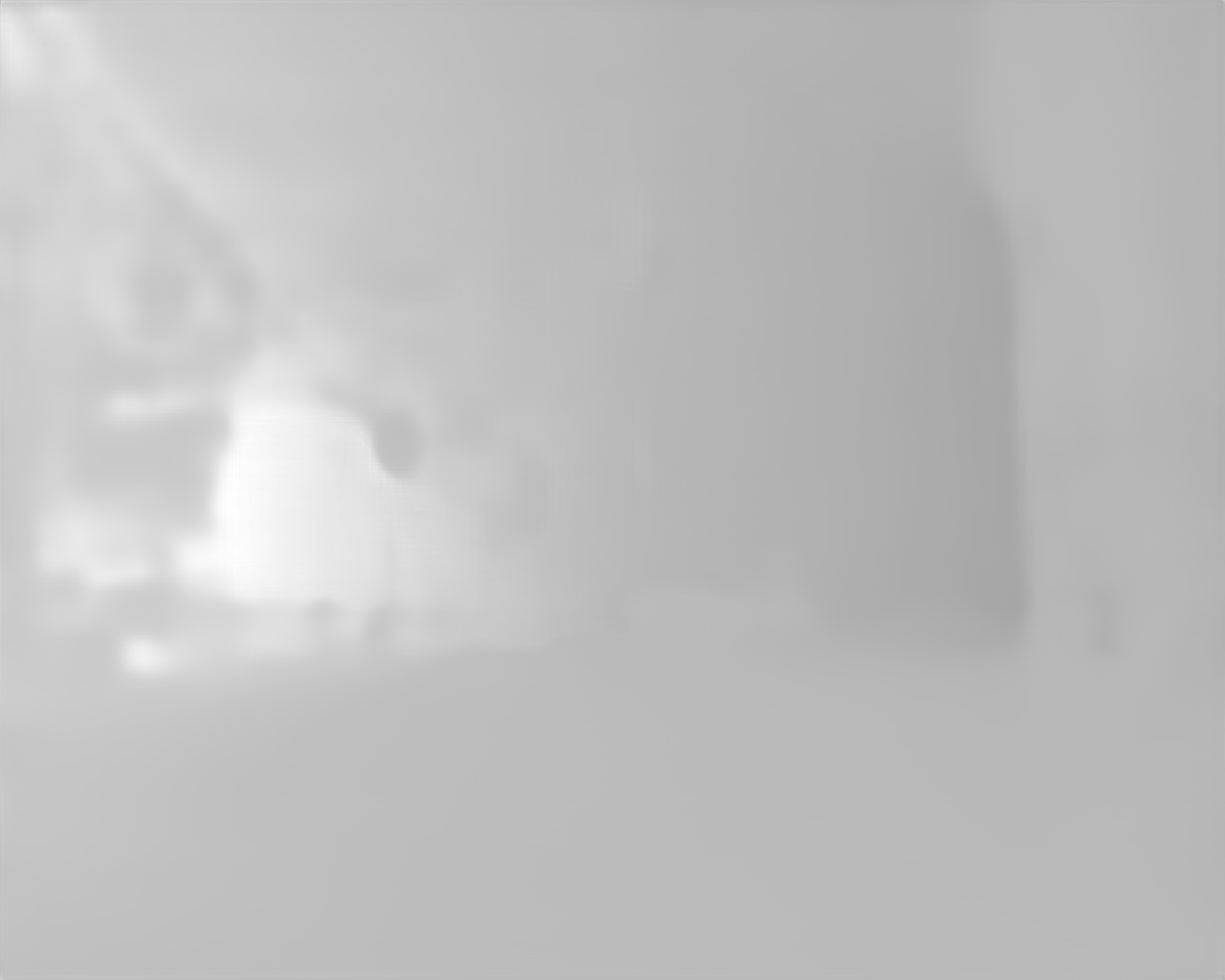}
  \caption{Lateral}
  \label{fig:sub3}
\end{subfigure}%
\begin{subfigure}{.25\textwidth}
  \centering
  \includegraphics[width=.99\textwidth]{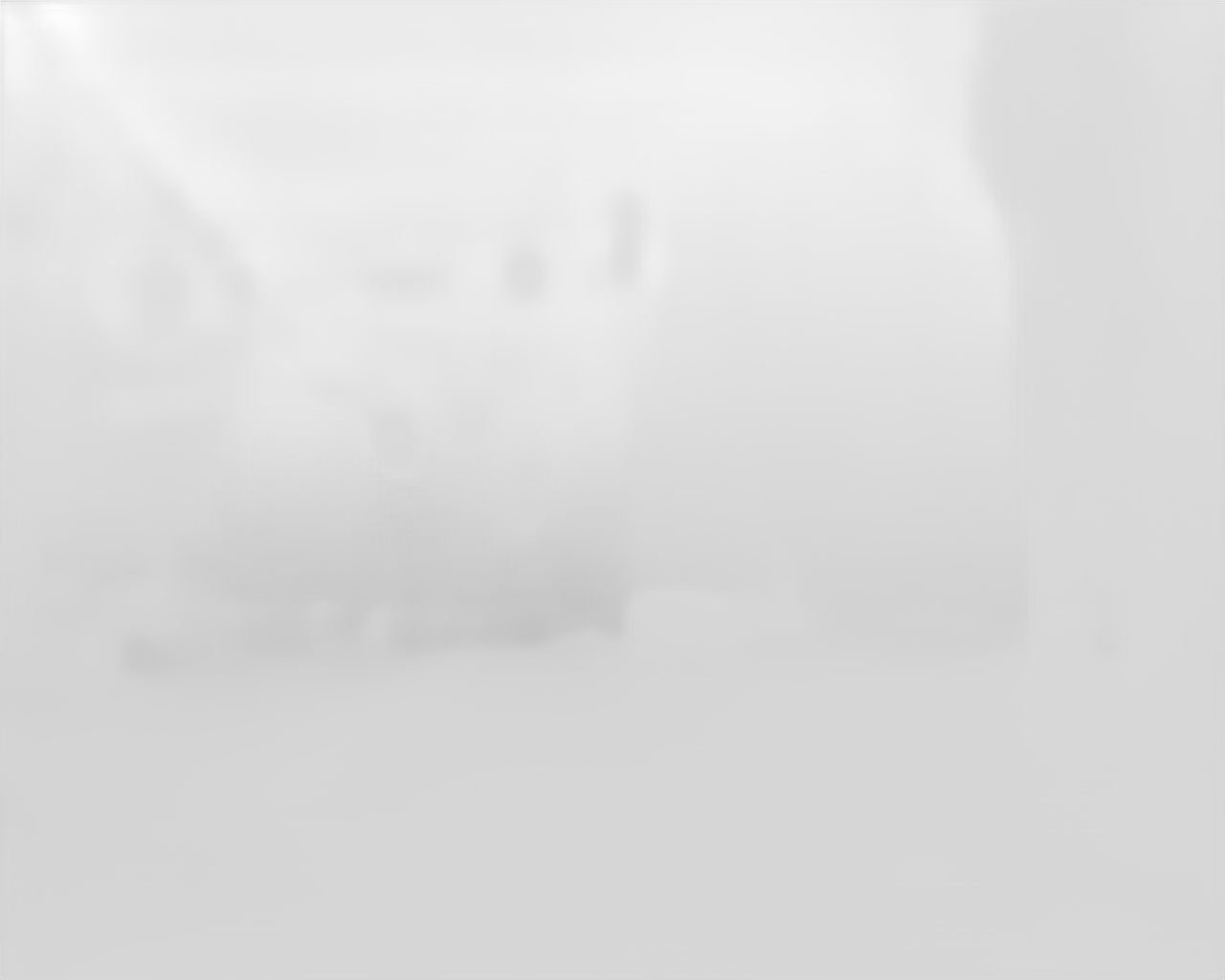}
  \caption{Height}
  \label{fig:sub4}
\end{subfigure}
\caption{Illustrative examples of the output of the BTS network}
\label{fig:3dRGB}
\end{figure*}

An example is given in Figure \ref{fig:3dRGB} to illustrate the outputs of the network. In Fig. \ref{fig:sub2} - Fig. \ref{fig:sub4}, the color of each pixel represents the distance (either depth, lateral or height) between the point depicted by the pixel and the camera. The whiter pixels indicate closeness to the camera. Note that the pixels which do not belong to any keypoint are ignored in the loss during the training. The output of the BTS network is projected into the WCS according to Eq. \ref{eq:bts_proj}. Since the radius of the Earth in meters was utilized during geo-registration, all the calculated distances are also presented in meters. 

\begin{equation}
    P = R^{T} \cdot B + C
    \label{eq:bts_proj}
\end{equation}

\noindent
where $P$ is the 3D position of a certain pixel in the image in the WCS, $R$ is the rotation matrix from the image reference frame to the WCS, $C$ is the position of the image in the WCS, and $B$ is the vector representing the lateral, height and depth distances of the point with respect to the image.


We may detect the same traffic sign from several images, which means we may get multiple predicted locations for a single traffic sign. To obtain a more accurate location of the traffic sign, we first filter out some noisy predictions by limiting the minimum and maximum distances from the camera to the detected traffic sign. The reason for setting the range of distance comes from the fact that the pixel-wise 3D localization algorithm tends to perform better when the traffic sign is within a certain range of distance to the camera. Therefore, the minimum and maximum distance thresholds are decided based on the performance of the modified BTS neural network on the test set (Section \ref{sec:3dobjectlocalization}). After that, we calculate the centre of the predicted locations within a specified radius, and set it as the location of the detected traffic sign.



\section{Datasets} \label{implementation}

\begin{table}[bt]
\caption{Dataset description including trajectories for data collection, camera setup, and the number of images collected from each trajectory}
\centering
\begin{tabular}{ c|c|c|c|c|c } 
\hline
Index & Date & Trajectory & \# of views & \# of images of each trajectory & Camera model\\
\hline
 I & Day 1 & \ref{fig:otaniemi} (E, F) - campus & single & 460, 576 &  Garmin 55\\
 II & Day 1 & \ref{fig:jatkasaari} (B) - residential & single & 725  &  Garmin 55\\ 
 III & Day 1 & \ref{fig:jatkasaari} (A) - residential & double & 1525 &  Garmin 55\\
 IV & Day 2 & \ref{fig:jatkasaari} (A) - residential & single & 706 &  RealSense D435\\
V & Day 1 & \ref{fig:jatkasaari} (C, D) - residential & triple & 945, 2874 &  Garmin 55\\
VI & Day 3 & \ref{fig:g_reg} (G, H) - campus & single (central) & 656, 2000 & iPhone 12 Pro Max \\
VII & Day 3 & \ref{fig:g_reg} (G, H) - campus & single (left)  & 656, 2000 & Garmin 55 \\
\hline
\end{tabular}
\label{table:datasets}
\end{table}

The data for the training and evaluation of the performance of the different system building blocks in Section \ref{eval} and Section \ref{eval_our} were collected from two different sites: in a residential area and around a university campus. All the collected datasets have extremely different appearances due to differences in weather or lighting conditions, camera models utilized, and camera placements. The distinction in weather and lighting conditions is a result of a data collection that took place on different days and even different seasons of the year. Notice that the purpose of having this variability across the datasets is to simulate a potential crowd-sourcing use. The length of roads present in the residential area summed up to 2.4 km, whereas on the campus, this number was 4 km. In all the recordings, the vehicles were driving mostly at a speed within the range from 20 to 30 km/h.

\begin{figure*}[tb]
\centering
\begin{subfigure}{.4\textwidth}
  \centering
  \includegraphics[width=.99\textwidth]{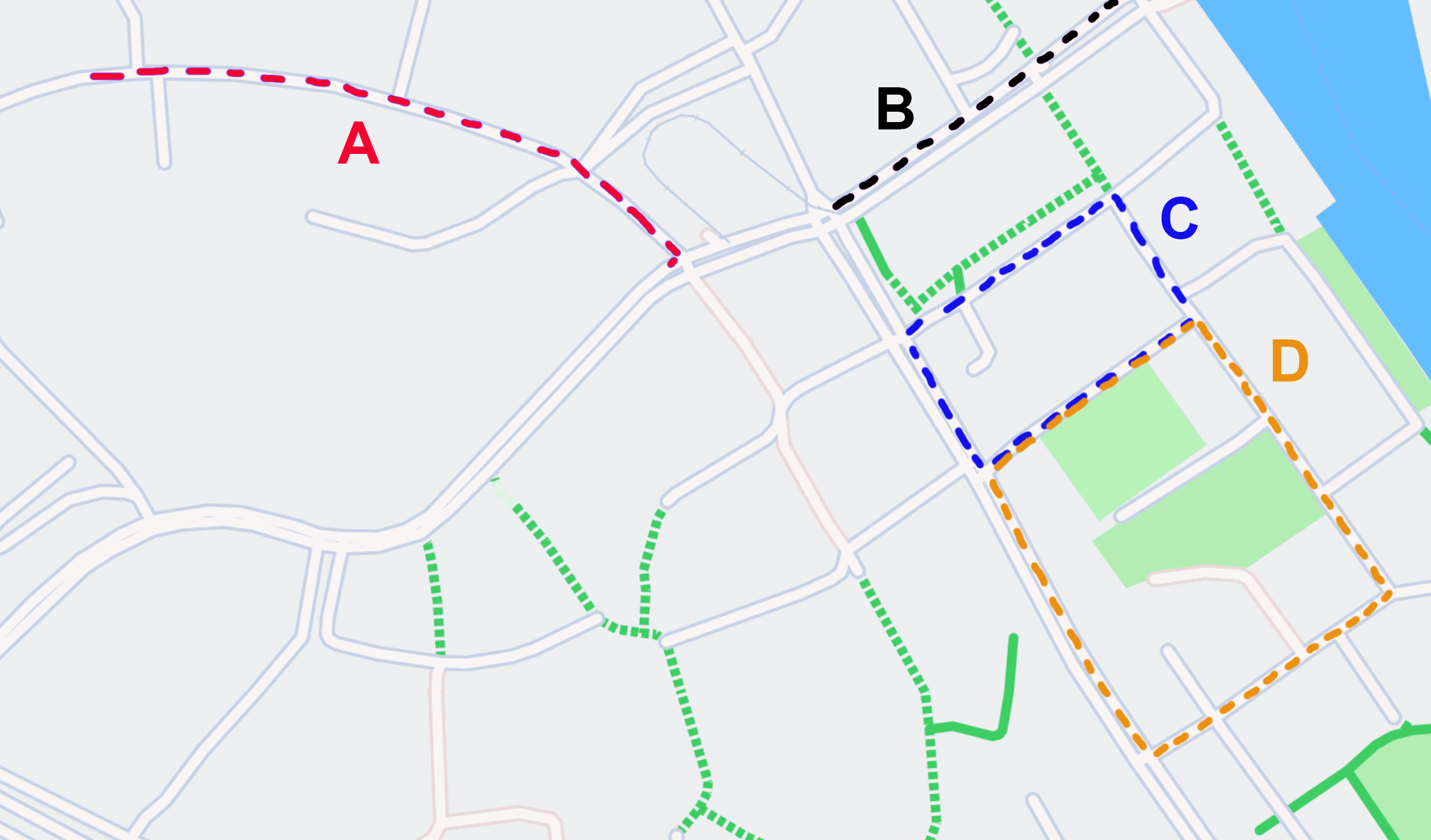}
  \caption{}
  \label{fig:jatkasaari}
\end{subfigure}%
\begin{subfigure}{.52\textwidth}
  \centering
  \includegraphics[trim=90 0 0 0,clip,width=.99\textwidth]{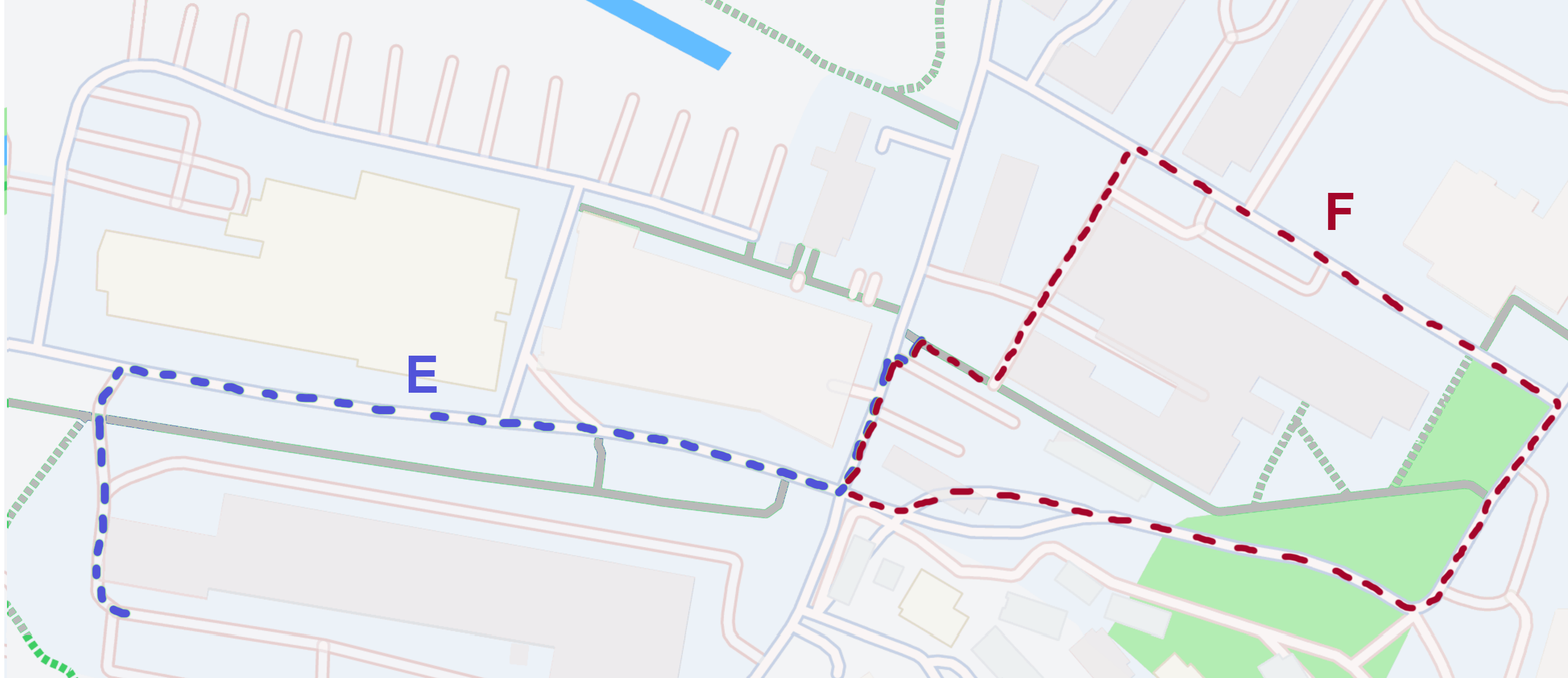}
  \caption{}
  \label{fig:otaniemi}
\end{subfigure}%
\\
\begin{subfigure}{.4\textwidth}
  \centering
  \includegraphics[trim=40 20 40 20,clip,width=.99\textwidth]{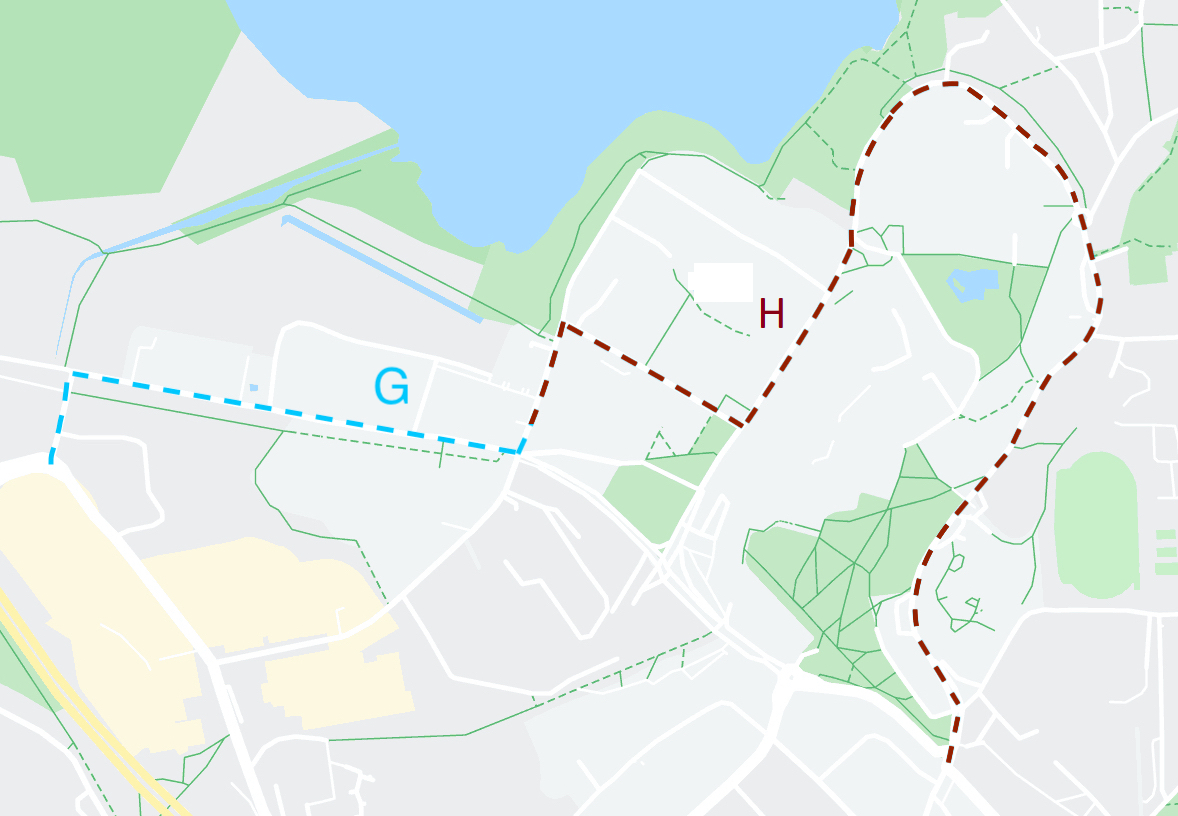}
  \caption{}
  \label{fig:g_reg}
\end{subfigure}%
\begin{subfigure}{.52\textwidth}
  \centering
  \includegraphics[trim=350 0  350 0,clip,width=.31\textwidth]{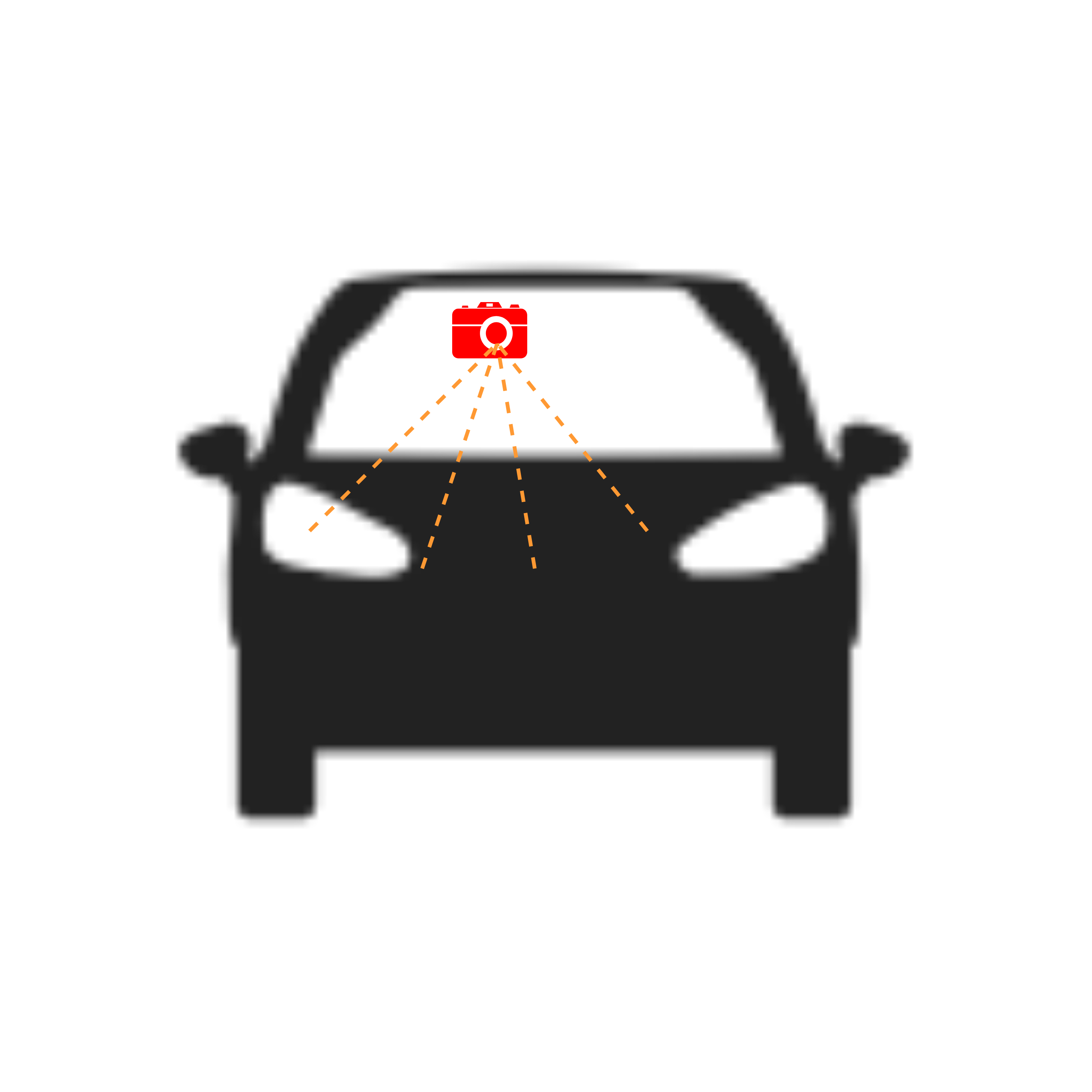}
  \includegraphics[trim=350 0 350 0,clip,width=.31\textwidth]{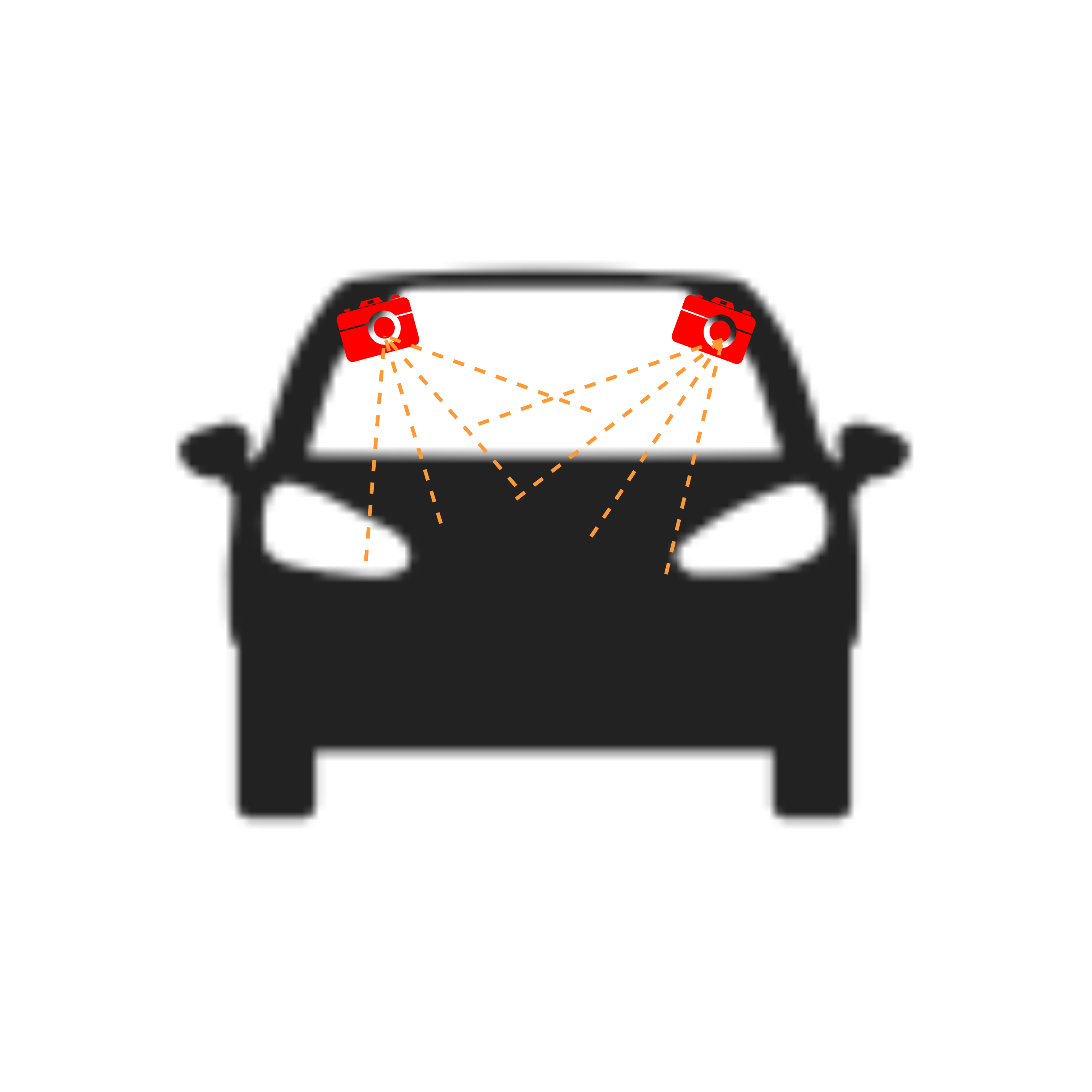}
  \includegraphics[trim=350 0 350 0,clip,width=.31\textwidth]{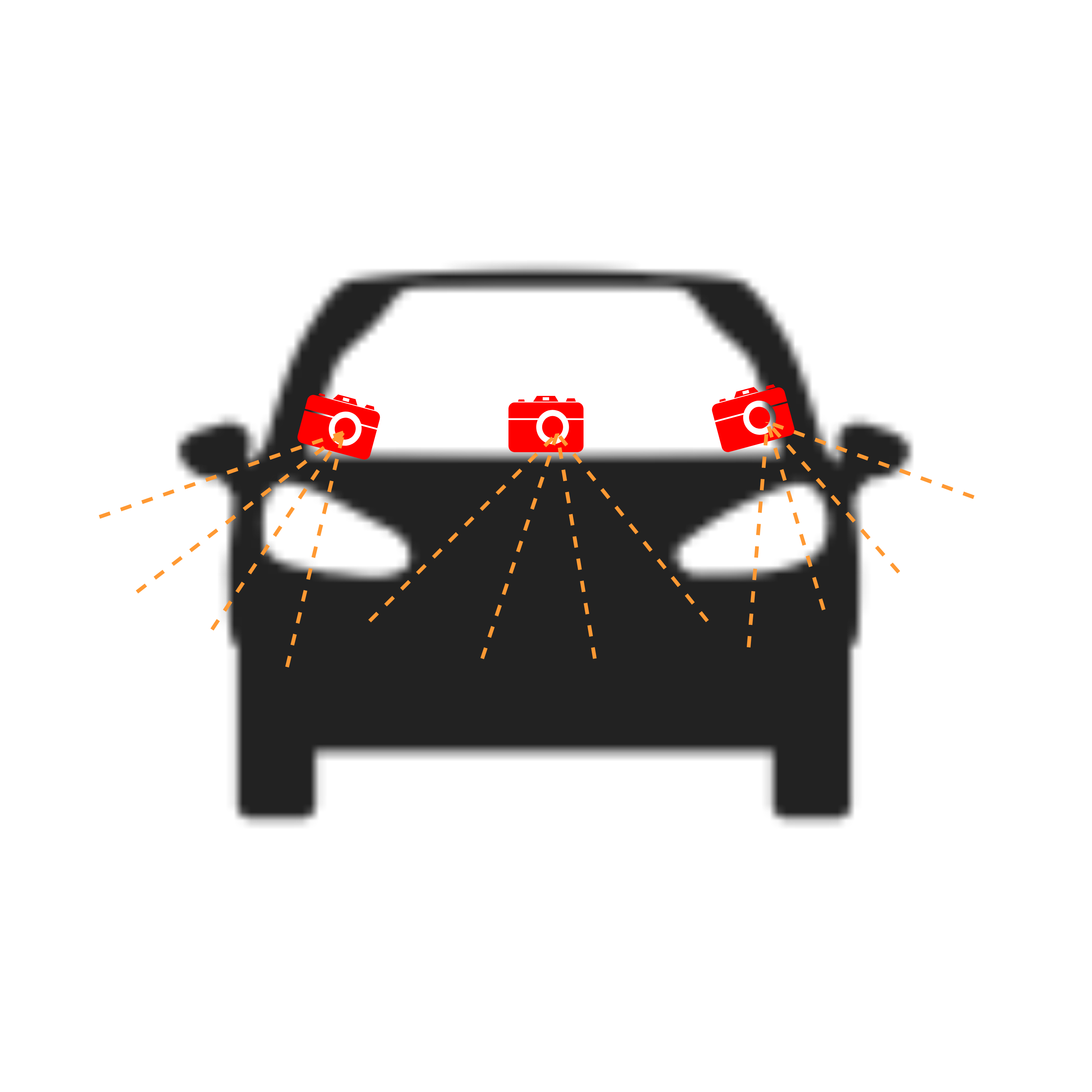}
  \caption{}
  \label{fig:cams3}
\end{subfigure}%
\caption{Data collection trajectories and camera placements. Fig. \ref{fig:jatkasaari} illustrates the trajectories A, B, C and D in a residential area. Fig. \ref{fig:otaniemi} and Fig. \ref{fig:g_reg} depict trajectories E, F and G
on a university campus. Fig. \ref{fig:cams3} different configurations of camera placements: single view vs. double view vs. triple view.}
\label{fig:cams}
\end{figure*}

\begin{figure*}[bt!] 
  \centering
    \includegraphics[width=\textwidth]{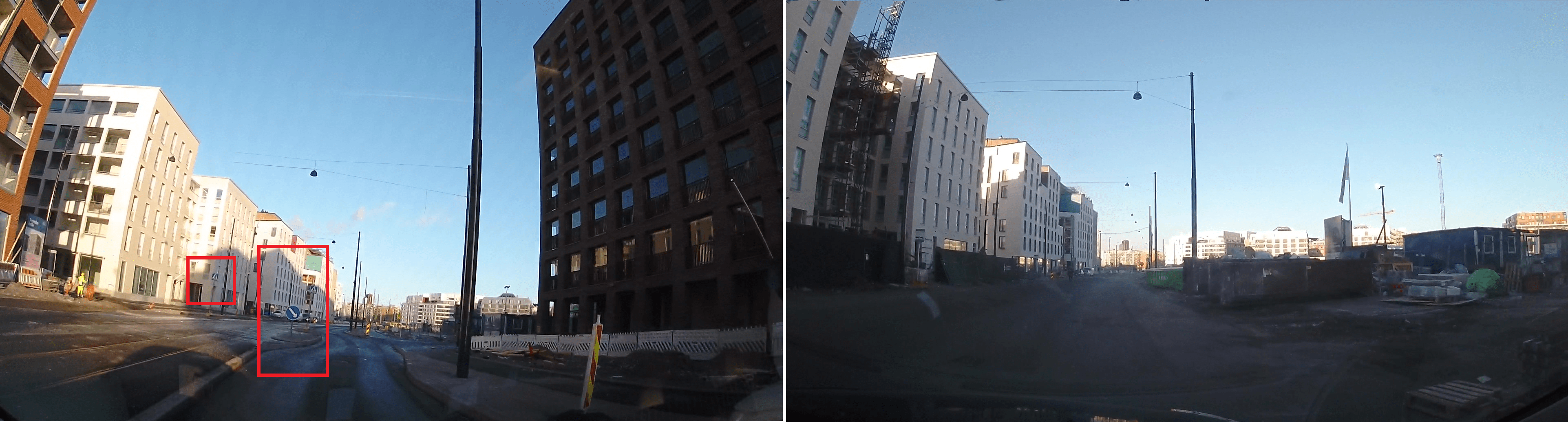}
    \caption{An example of scenario change in the residential area. The image on the left is taken from Dataset IV, whereas the one on the right from Dataset III. Notice that the traffic signs inside the red rectangles are only present in one state of the environment.}
\label{fig:example_change}
\end{figure*}

Table \ref{table:datasets} summarizes the seven datasets, which in total included 23057 images. 
As summarized in Table \ref{table:datasets}, seven datasets, including in total 23057 images, were generated. A Garmin 55 dashcam was utilized on Day 1, an Intel RealSense D435 on Day 2, and an iPhone 12 Pro Max on Day 3. The camera setups are visualized in Figure \ref{fig:cams3}. All the cameras were placed to face roughly the same direction, which results in an overlap in the views for datasets with more than one view. The video resolution was set to 1920x1080, while the frame rate to 30 FPS, which was later downsampled by decimation to 10 FPS as a higher frame rate is not required given that the vehicle speed is relatively low.



The camera recordings took place on February 22, 2019 (Day 1), December 2, 2019 (Day 2), and February 2021 (Day 3). During data collection, the vehicle went through 4 trajectories in the residential area (Fig. \ref{fig:jatkasaari}) and 4 trajectories on the campus (Fig. \ref{fig:otaniemi} and Fig. \ref{fig:g_reg}). For the residential area, only trajectory A (Datasets III and IV) suffered changes from February 2019 to December 2019, hence they will be utilized for evaluating the change detection. In the campus area, changes have not been captured. However, our change detection algorithm will still be utilized in trajectory G (Datasets VI and VII) to confirm the absence of changes. Figure \ref{fig:example_change} illustrates an example of a change in the environment. The remaining data are assigned to the training of the pixel-wise 3D localization system component (Step B.1.1). 





\section{Performance of the Initial 3D Metadata Generation} \label{eval}

Our system generates the initial 3D metadata of SfM point clouds at Step A.3. The accuracy of the generated metadata depends on the accuracy of the SfM point cloud built at Step A.1 as well as the accuracy of semantic segmentation (Step A.2.1) and object detection (Step A.2.2). In this section, we evaluate the accuracy of each building block of the Geometric Map and Semantic Map layers and analyze how it affects the overall accuracy of the generated 3D metadata. We conducted all the experiments on a system running Ubuntu 18.04 and powered by four NVIDIA GTX 1080 Ti with 11GB of RAM each and two Intel Xeon Gold 6134 CPUs.

\begin{figure}[t] 
\begin{subfigure}{.31\textwidth}
  \centering
    \includegraphics[angle=270,trim=0 115 0 75,clip,width=0.99\textwidth]{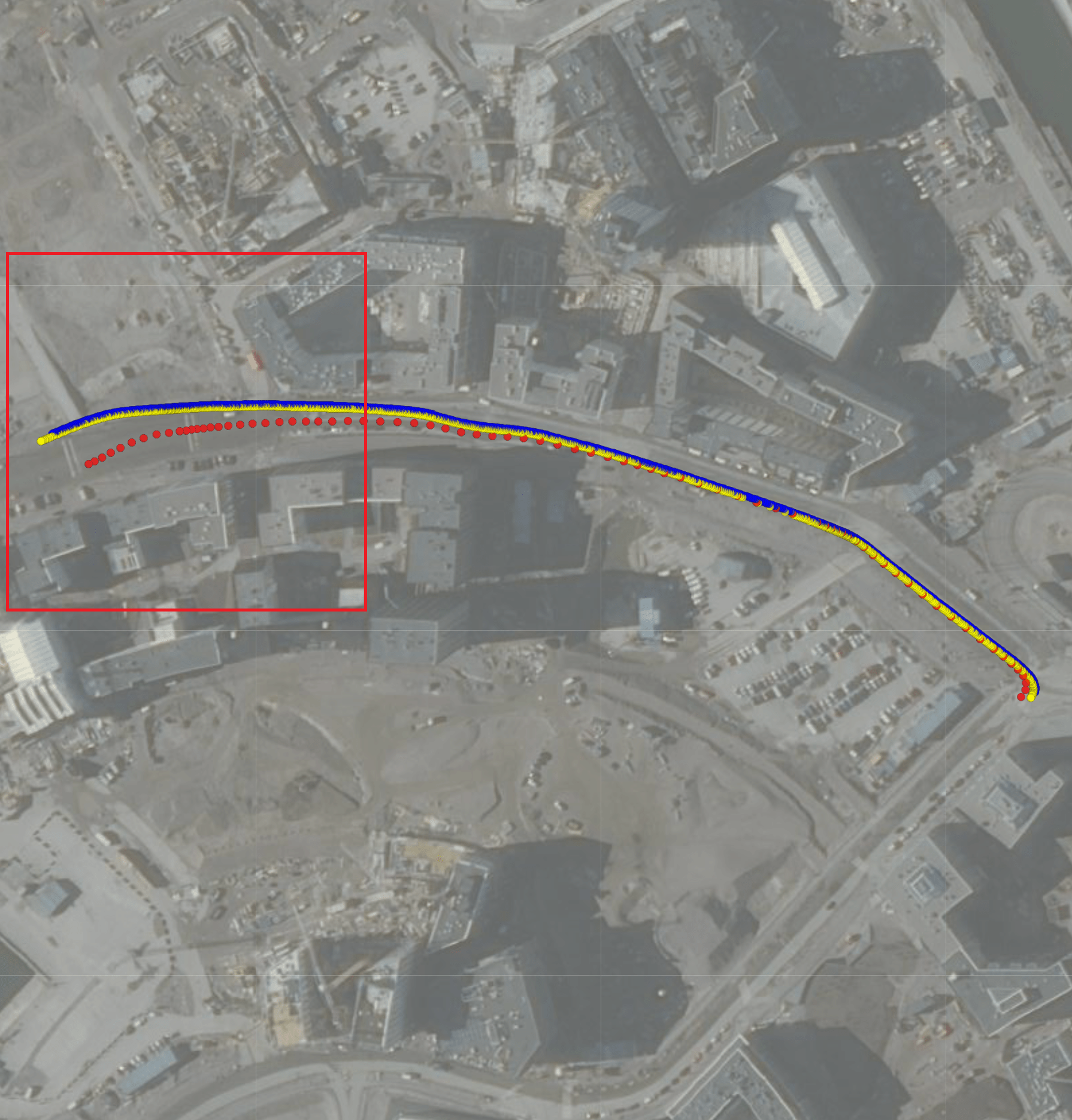}
    \caption{}
\label{fig:map}
\end{subfigure}\hspace{4mm}
\begin{subfigure}{.665\textwidth}
\centering
    \includegraphics[trim=0 0 0 0,clip, width=.99\textwidth]{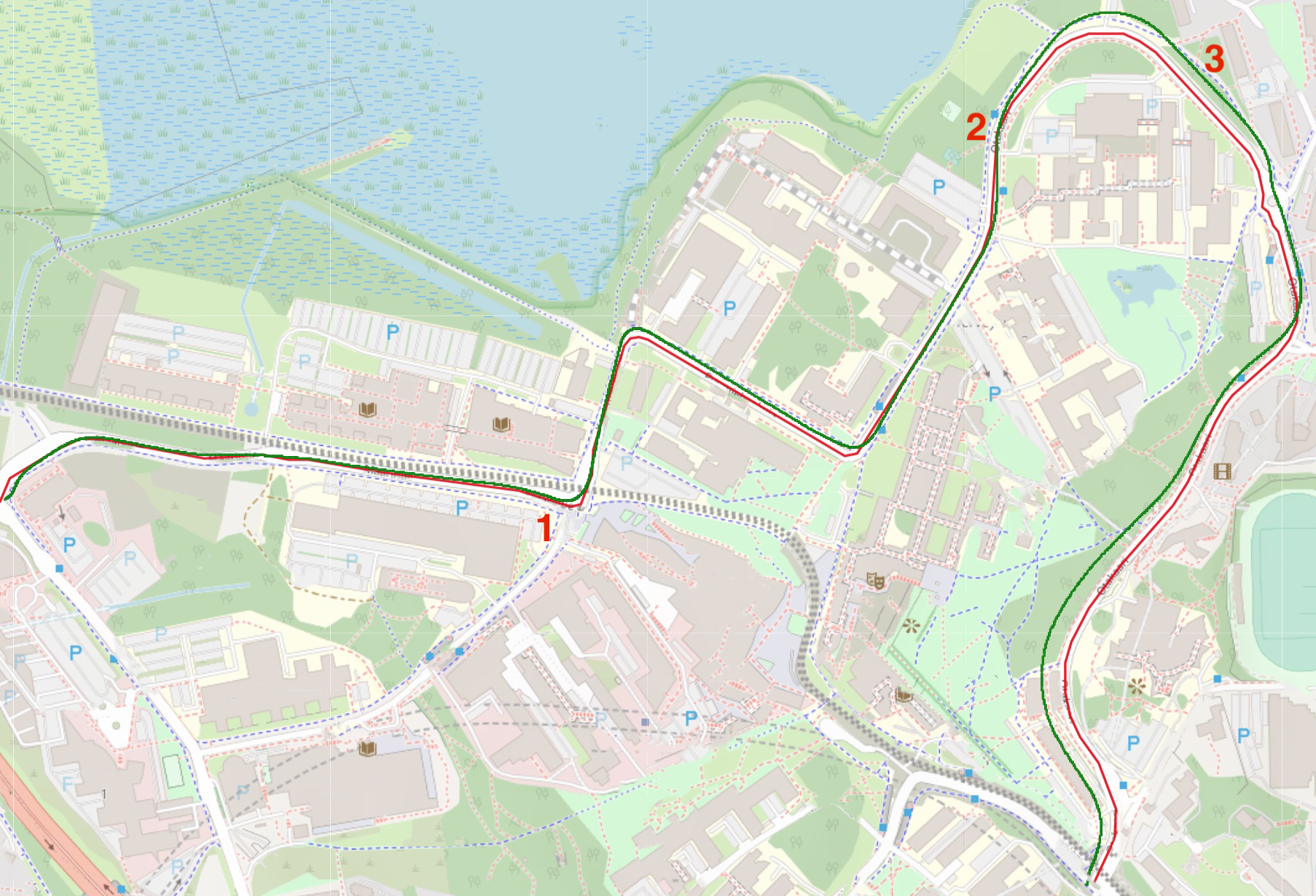}
\caption{}
\label{fig:map_campus}
\end{subfigure}
\caption{(a). Estimated by SfM camera positions of each image in Dataset III (camera 1 in blue and camera 2 in yellow) vs. the ground truth obtained with RTK technology (in red); (b) Camera poses of Dataset VI estimated by SfM (green) vs the path extracted from Google Maps for comparison (red solid line), the red numbers specify the locations of example images in Fig. \ref{fig:feat_example}}
\end{figure}

\subsection{SfM-based 3D Reconstruction}
\label{sec:sfmrecons}
As discussed in Section \ref{background}, we implement 3D reconstruction based on COLMAP \cite{7780814}, and employ the vocabulary-tree-based method and sequential method for feature matching \cite{schonberger2016vote}. For each trajectory described in Section \ref{implementation}, we create a sparse point cloud and perform geo-registration on it. The quality of the generated point cloud can be reflected in the accuracy of the estimated camera positions (Step B.1.3) and pixel-wise localization (Step B.1.1), which in general affects the accuracy of 3D object localization (Step B.2) in the process of change detection. In addition, point cloud quality affects the localization accuracy of the metadata at Step A.3. 

We take Dataset III as an example to evaluate the accuracy of the camera pose estimations produced at Step A.1. Dataset III includes video collected from two dashboard cameras at 30fps and RTK-based positioning data with a 1Hz sampling rate. Since RTK provides centimeter-level positioning accuracy, the positions derived from RTK samples are considered as ground truth in this case. The estimated camera positions are compared with RTK measurements in Figure \ref{fig:map}. Since the sampling rate of RTK positioning data is lower than the frame rate of the video, we calculate the distance error of each estimated camera pose as the distance from its closest RTK position. The median distance error is 7.09 meters, with a standard deviation of 4.96 meters. Concerning the bias caused by inconsistent sampling rates, the actual errors may be lower than the ones we calculated. 
\begin{figure}[t] 
\begin{subfigure}{.31\textwidth}
  \centering
    \includegraphics[angle=0,trim=0 0 0 0,clip,width=0.99\textwidth]{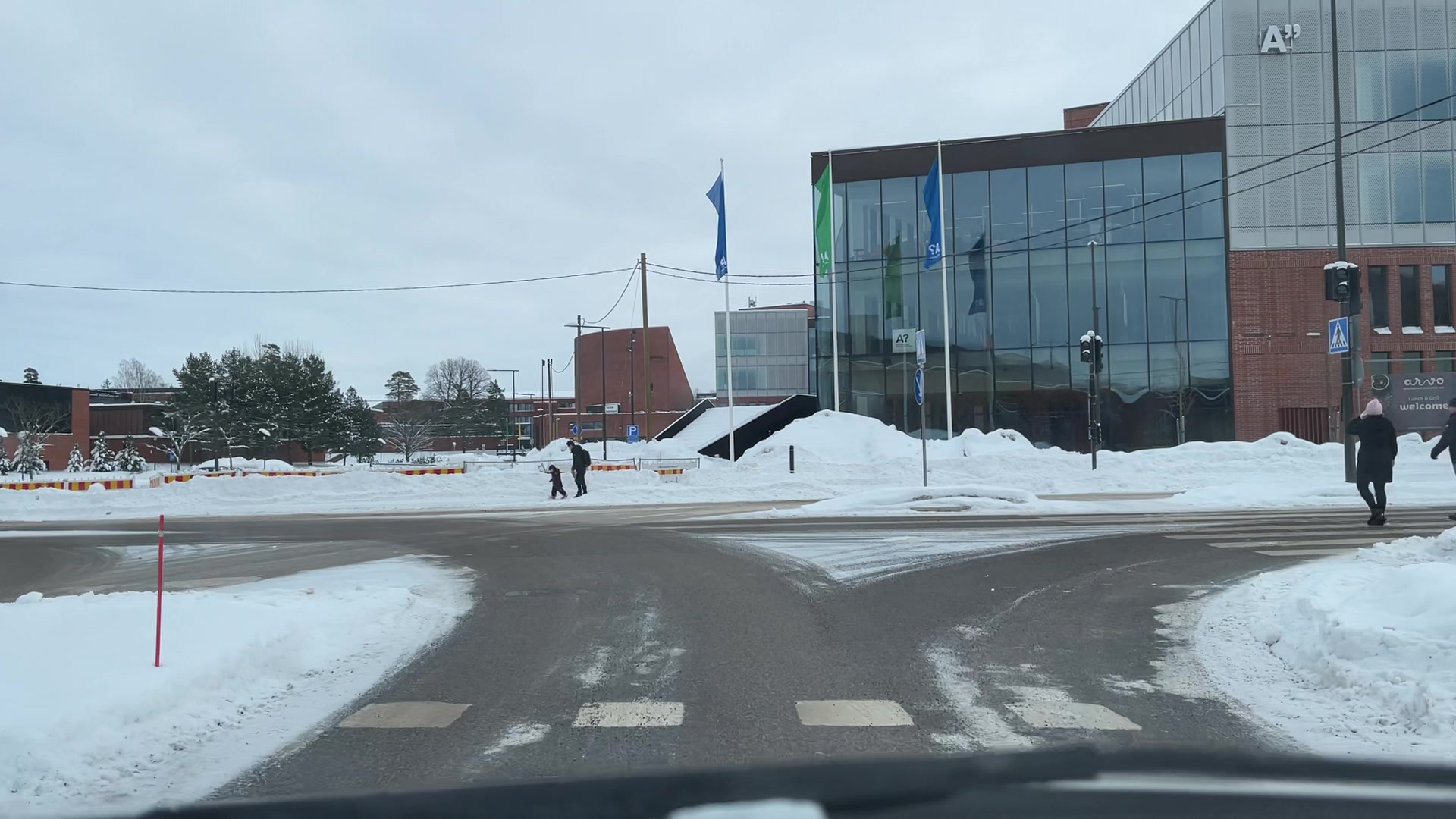}
    \caption{}
\label{fig:sufficient}
\end{subfigure}\hspace{4mm}
\begin{subfigure}{.31\textwidth}
\centering
    \includegraphics[trim=0 0 0 0,clip, width=.99\textwidth]{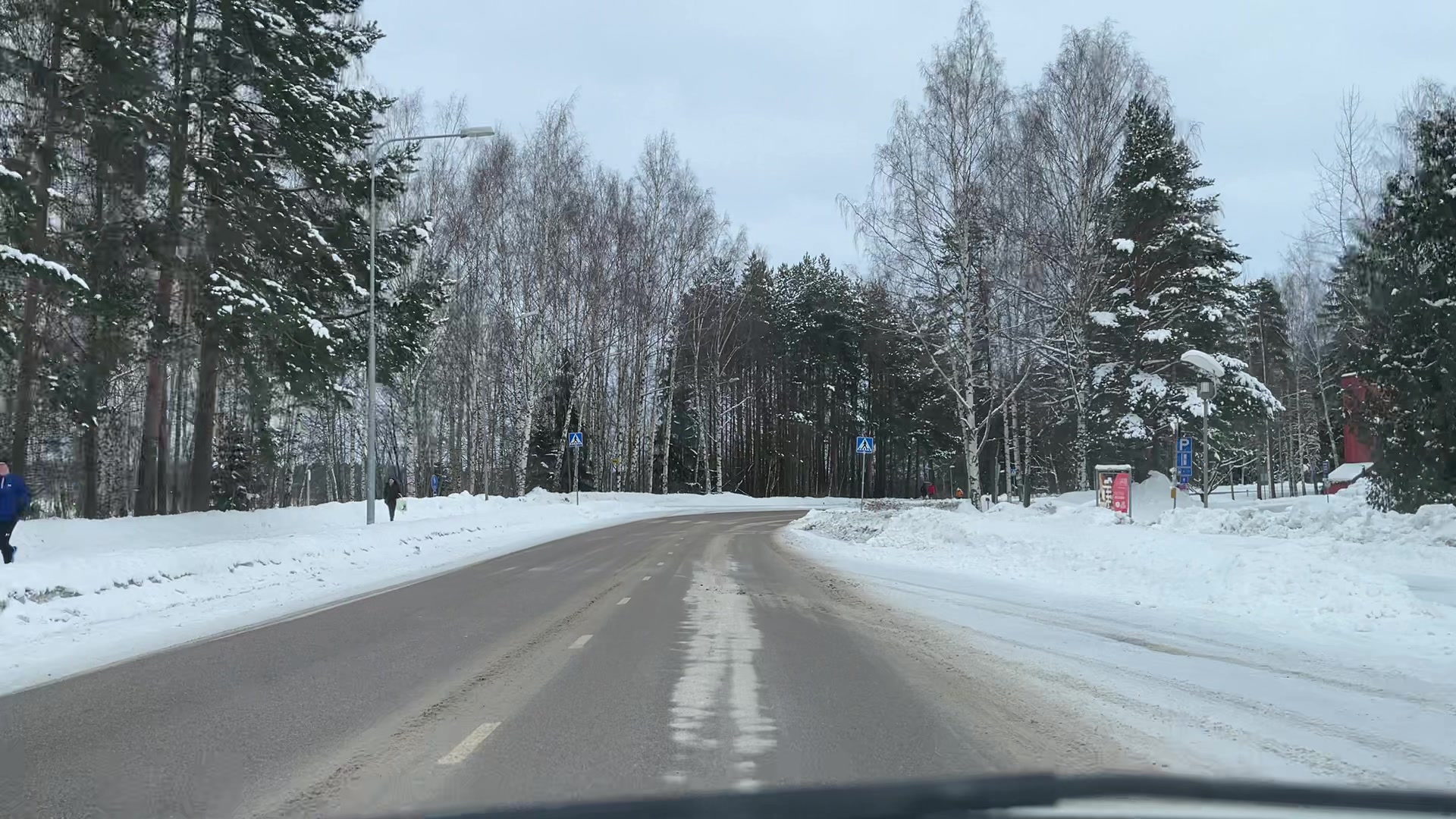}
\caption{}
\label{fig:insuff1}
\end{subfigure}
\begin{subfigure}{.31\textwidth}
\centering
    \includegraphics[trim=0 0 0 0,clip, width=.99\textwidth]{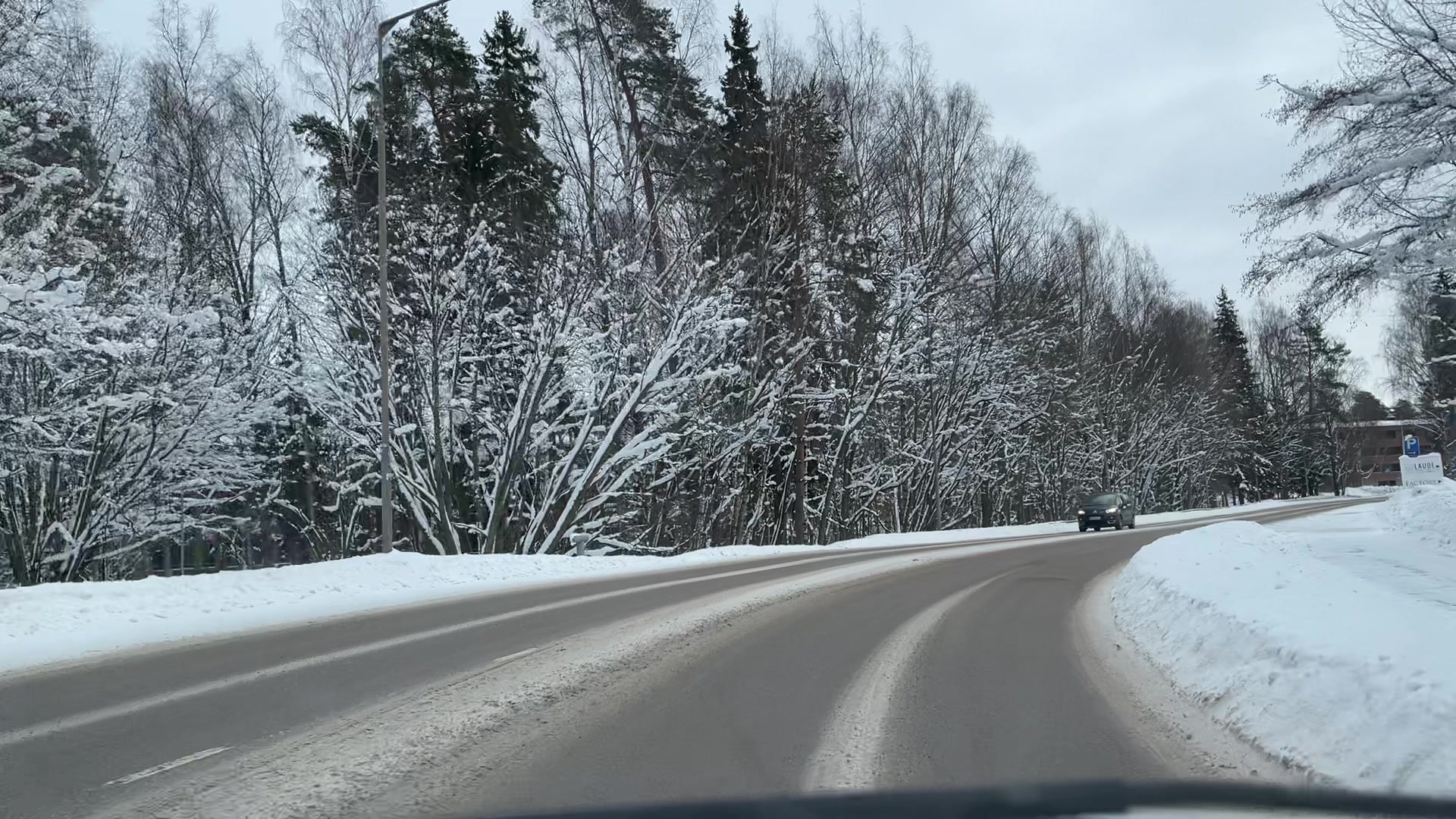}
\caption{}
\label{fig:insuff2}
\end{subfigure}
\caption{Examples of images with sufficient and insufficient number of recognizable features. While a) has enough recognizable features due to the presence of buildings, b) and c) suffer from the lack of features since they are mostly covered with trees.}
\label{fig:feat_example}
\end{figure}

\begin{figure}[tb]
\centering
\includegraphics[width=1.0\textwidth]{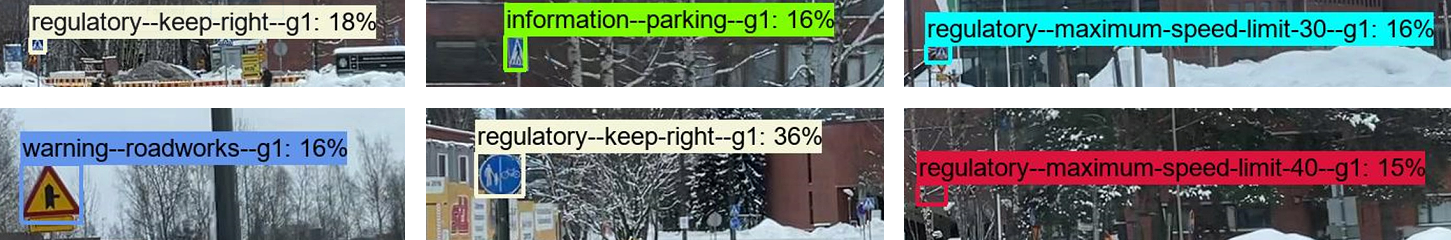}
\caption{Examples of mistakes in the objection detection algorithm. On the first row, informative traffic signs of pedestrians crossing are mistaken as being of another type. On the second row, a few more examples of misclassification including an instance where tree branches are mistaken as a traffic sign.}
\label{fig:mistakes_od}
\end{figure}

As highlighted using a red box in Figure \ref{fig:map}, an accumulated drift appears at the upper end of the trajectory. Such a drift can be significantly reduced if 1) the SfM reconstruction is performed with the data that has a loop closure, 2) the area of reconstruction is not too large, for example, one building block with a loop closure making up around 600 meters (see region C and D in Fig. \ref{fig:jatkasaari}), and 3) the general recommendations for the SfM reconstruction are met - such as good visibility, scenery rich with features, sufficient overlap between the camera views, etc. As shown in Fig. \ref{fig:feat_example}, the image \ref{fig:sufficient} has distinguishable visual features and helps in more accurate SfM reconstruction (the location of the camera pose is identified by the number 1 in Fig. \ref{fig:map_campus}. While images \ref{fig:insuff1} and \ref{fig:insuff2} have similar appearances of the snowy roads and the forest, though being in different locations (see their locations identified by the number 2 and 3 in Fig. \ref{fig:map_campus}). This example is most cases fails to match unless the surrounding images are specified using for example sequential matching. Except for that, the estimated camera positions align well with the ground truth. In addition, the alignment depends on the selected GPS coordinates for the geo-registration. In this case, those were selected in order to align properly the side of the road outside the red box in Figure \ref{fig:map}. While it is impossible to select coordinates in a way to align precisely with every part of the reconstruction, it can be selected in a way to distribute the error equally along the reconstructed region.


Figure \ref{fig:map_campus} compares the camera poses estimated by the SfM reconstruction with the GPS path extracted from Google maps. In a similar way to reconstruction in Figure \ref{fig:map} the geo-registration of the region has been intentionally selected to align more precisely the upper side of the region. That side has a greater number of visual features, while starting from the center of the path the environment is mostly in the forest and covered by the snow. For the reason of the featureless region, the selected feature matching was sequential. This is a more appropriate method since it can track images to be matched in sequential order and performs better when the environment shows high level of similarity along the way, such as a forest. The drawback is the accumulated drift, which can be seen at the top and the right-hand side of the path.

\subsection{Traffic Sign Detection and Localization} 
\label{sec:offline_localization}
\subsubsection{Semantic Segmentation on 2D Images} 
\label{sec:sson2dimages}

We implement the semantic segmentation component (Step A.2.1), described in Section \ref{sem_objdet}, based on the Seamseg architecture proposed by Mapillary \cite{8954334}. We utilize the model trained and tested by the authors on the Mapillary dataset \cite{8237796}, which achieves 50.4\% IoU (Intersection over Union) \cite{8954334}. A visual example of the model performance can be seen in Figure \ref{fig:semseg2}.

\subsubsection{Object Detection} \label{obj_det_eval}


TensorFlow Object Detection API is utilized for the traffic sign detection component, as was mentioned in Section \ref{sem_objdet}. As for the neural network architecture, the SSD Resnet-50 FPN pre-trained on COCO dataset was selected \cite{ssdresnet} due to a good trade-off between accuracy and speed. The training set is composed of approximately 16000 annotated images of traffic signs from the Mapillary dataset \cite{8237796}. However, since certain traffic signs present in our test regions differ significantly from those available in Mapillary (see Table \ref{table:tfs}), we expanded the dataset by including additional 4000 annotated images focused on the traffic signs exclusive to our test regions. Half of these additional annotated images consist of real images collected in different regions, but in the same country. The other half is composed of images where the traffic signs of interest were artificially overlaid on generic background images. The 24 traffic sign classes present in the test regions are illustrated in Table \ref{table:tfs}. Overall, 20000 annotated images formed the dataset, where 500 of them were utilized for validation with the rest being assigned to training. The training consisted of 30 epochs.

As test set, we manually labeled the traffic signs in 97 images of region G on Day 3 (i.e. Dataset VI). Note that we purposely test the object detection on images recorded from a different camera than the images that formed the training set. This allows for a more realistic measurement since it is expected that crowdsourced data are taken from distinct cameras. The object detection algorithm demonstrated an mAP (mean average precision) of 0.518 at IoU threshold of 0.4, which is a fairly good result given the circumstances of different weather, lighting, camera conditions. Figure \ref{fig:mistakes_od} illustrates examples of mistakes by the object detection algorithm. It is observed that when the pedestrians crossing traffic sign is located at a far distance to the car or in case it is at a tilted angle with respect to the direction of movement of the car, the algorithm mistakes it for another traffic sign. Other examples included in Figure \ref{fig:mistakes_od} indicate that the algorithm can confuse traffic signs of similar appearances - which is the case of the roadworks and perpendicular road junction traffic signs. Since the confidence score of these misclassifications happens to be below 0.4, we opted to discard any detection whose score is below this threshold value.



\begin{table}[b]
\centering
\caption{Traffic signs installed along Trajectory A (left) and Trajectory G (right). The traffic sign names presented with an asterisk \text{*} are those included in our expansion of the Mapillary dataset.}
\begin{tabular}{ c|c} 
 Class & Image  \\
\hline
 information--pedestrians-crossing--g1 & \includegraphics[height=0.4cm]{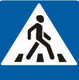} \\ 
 shared-path-pedestrians-and-bicycles--g1* & \includegraphics[height=0.4cm]{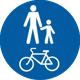} \\ 
information--dead-end-except-bicycles--g1* & \includegraphics[height=0.4cm]{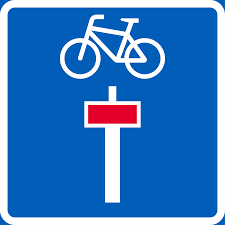} \\ 
complementary--obstacle-delineator--g2 & \includegraphics[height=0.4cm]{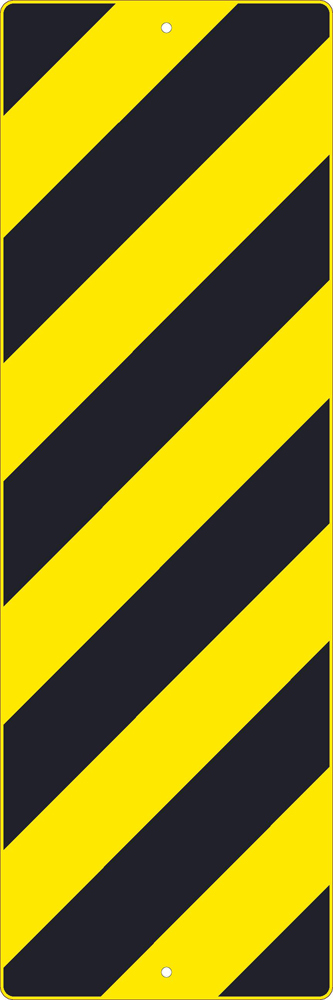}  \\
 complementary--one-direction-right--g1 & \includegraphics[height=0.4cm]{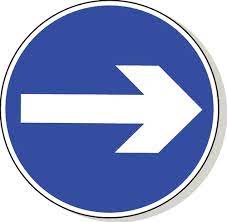} \\ 
 regulatory--pass-on-either-side--g1 & \includegraphics[height=0.4cm]{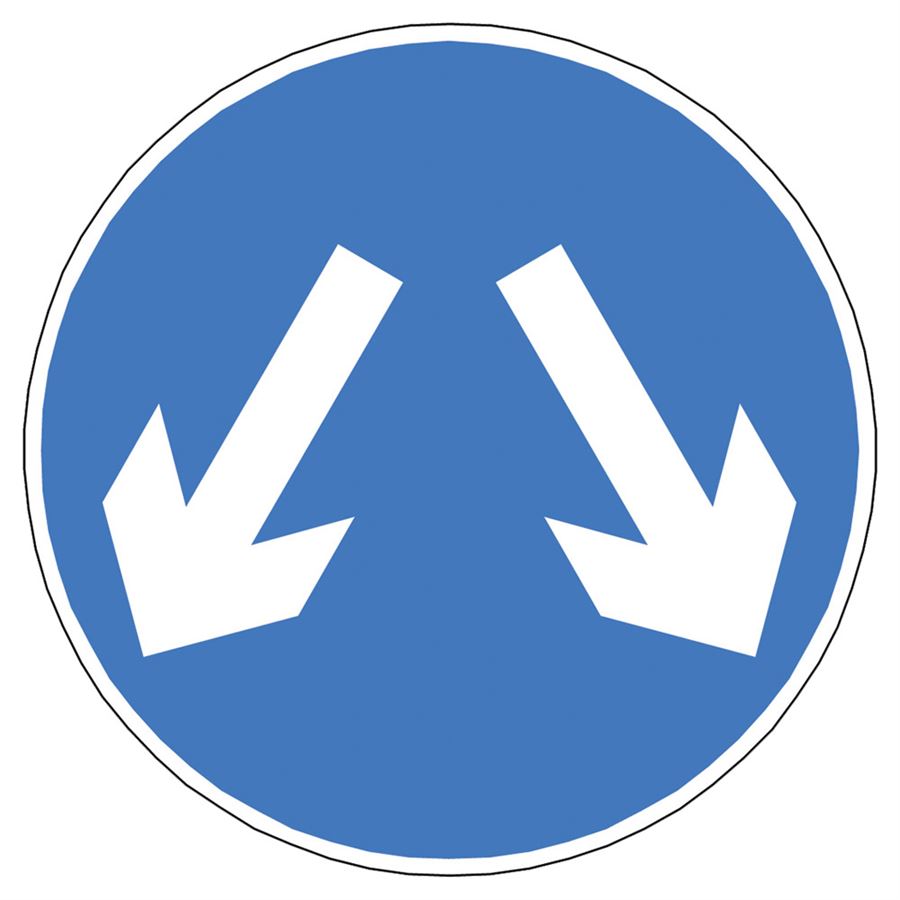}  \\ 
regulatory--maximum-speed-limit-40--g1* & \includegraphics[height=0.4cm]{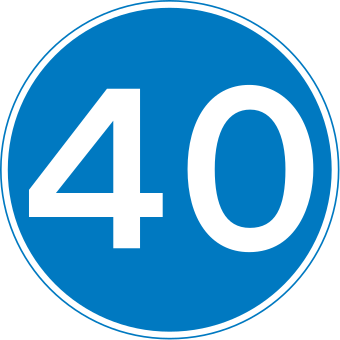}  \\
regulatory--maximum-speed-limit-30--g1* & \includegraphics[height=0.4cm]{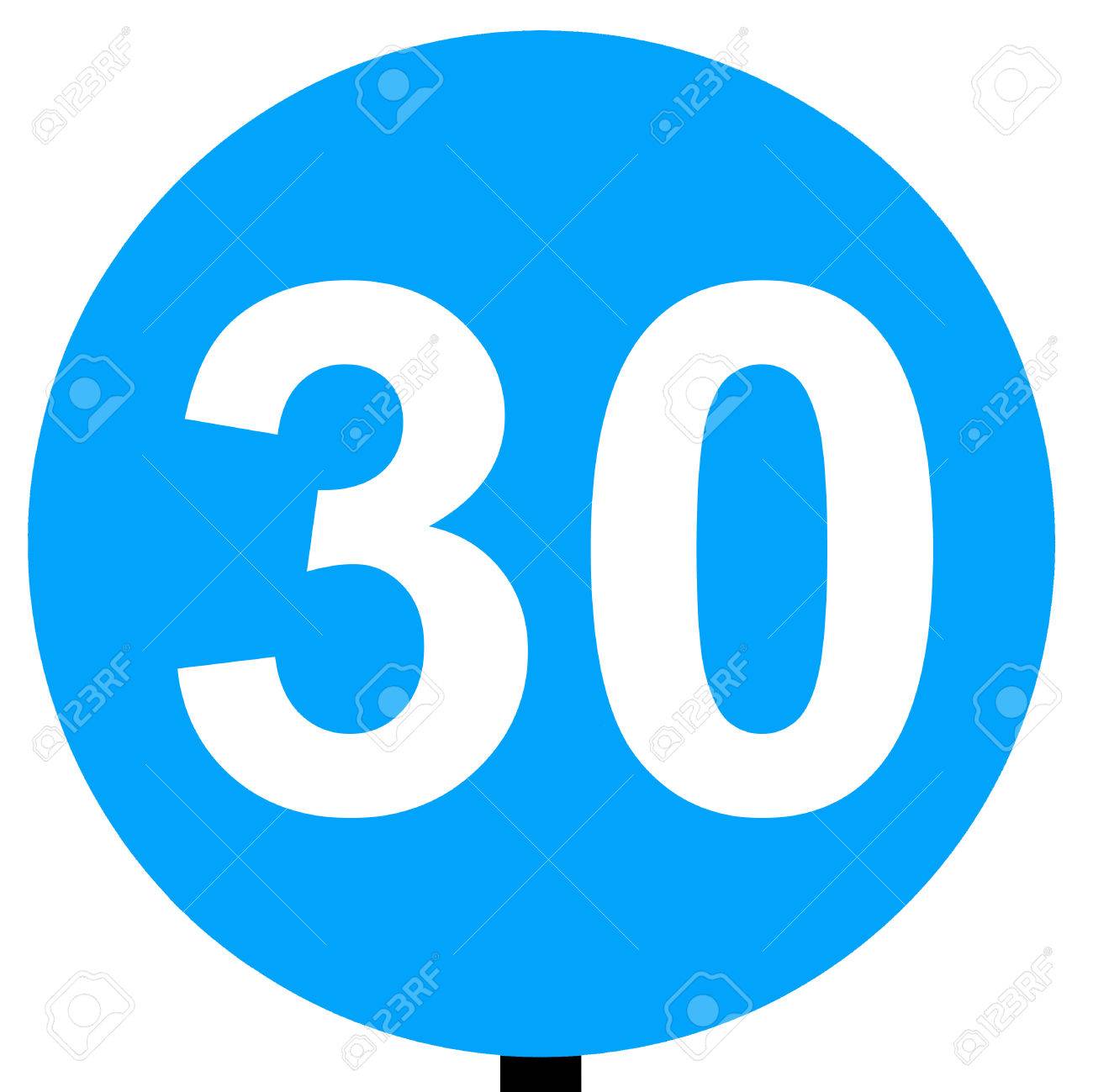}  \\
regulatory--end-of-maximum-speed-limit-40--g1* & \includegraphics[height=0.4cm]{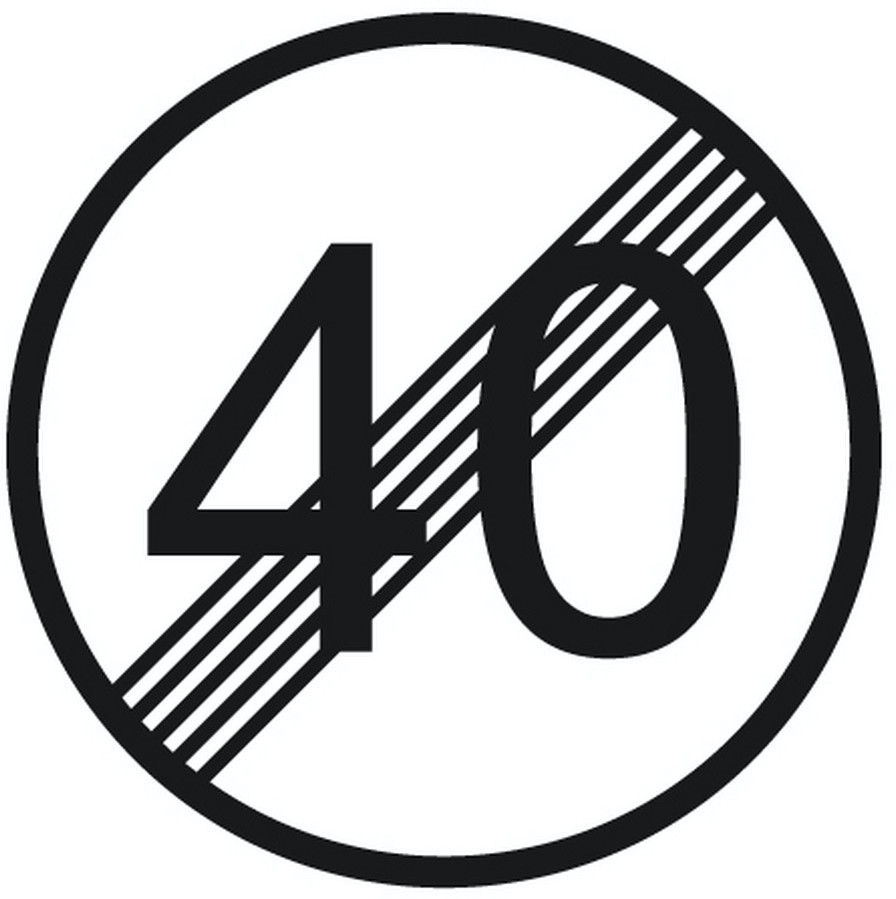}  \\
regulatory--no-pedestrians-or-bicycles--g2* & \includegraphics[height=0.4cm]{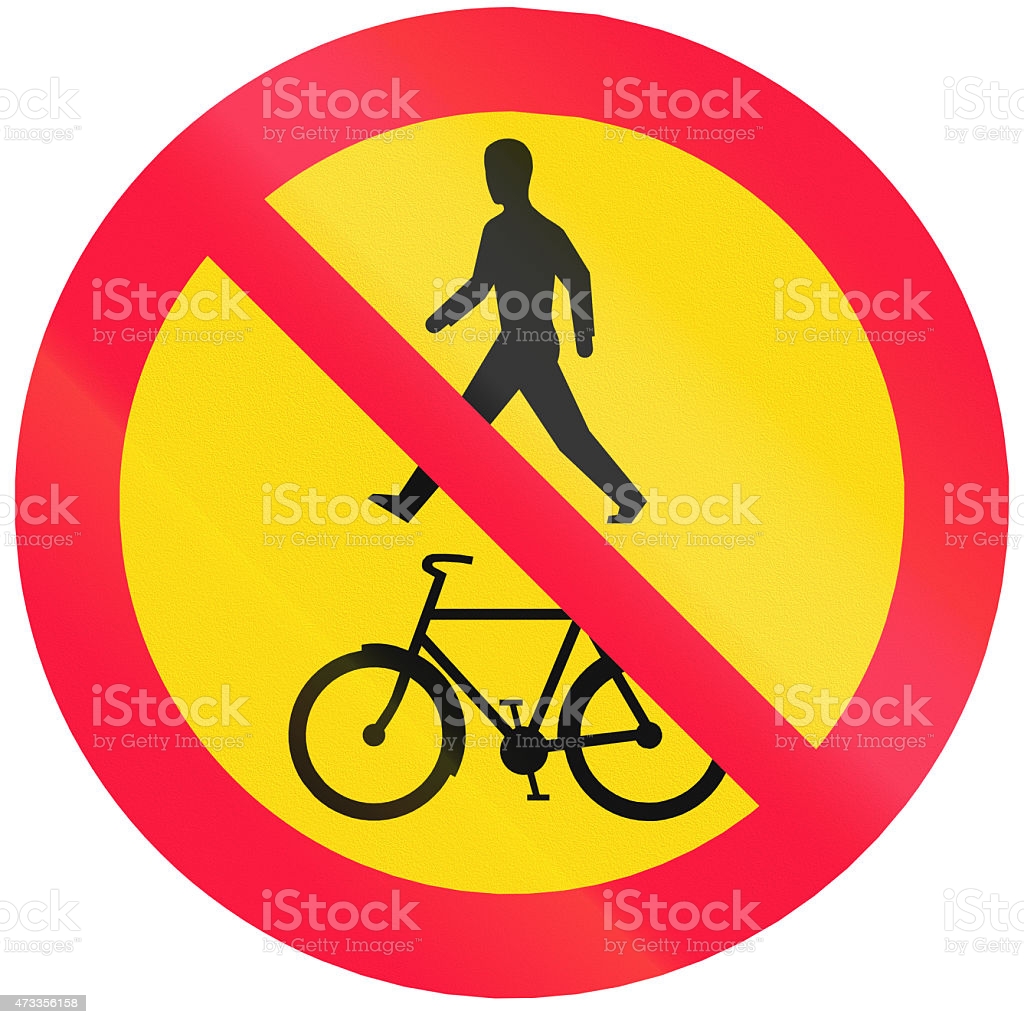} \\
junction-with-a-side-road-perpendicular-right--g1* & \includegraphics[height=0.4cm]{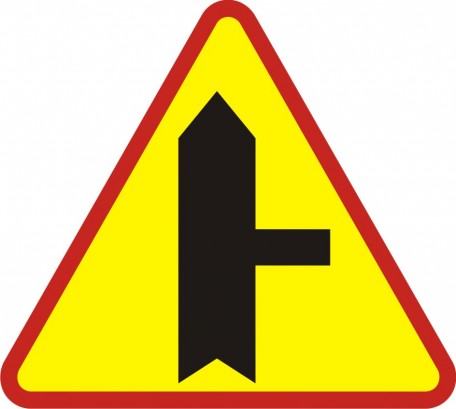}  \\
regulatory--no-motor-vehicles--g6* & \includegraphics[height=0.4cm]{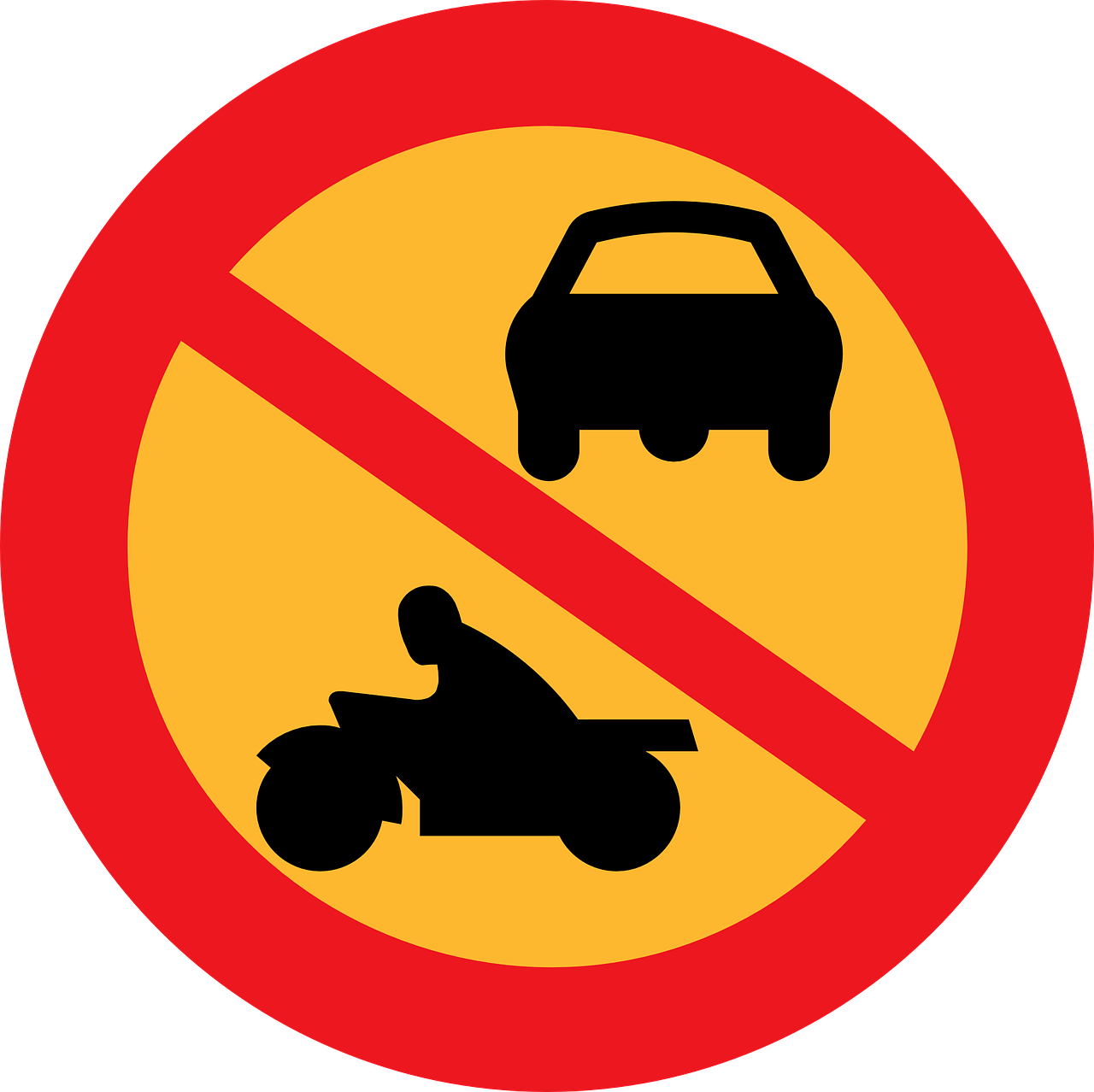}  \\
\hline
\end{tabular}
  \hspace{2.5em}
\begin{tabular}{ c|c} 
 Class & Image  \\ \hline
    regulatory--no-parking--g1 & \includegraphics[height=0.4cm]{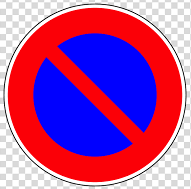} \\
    regulatory--roundabout--g1 & \includegraphics[height=0.4cm]{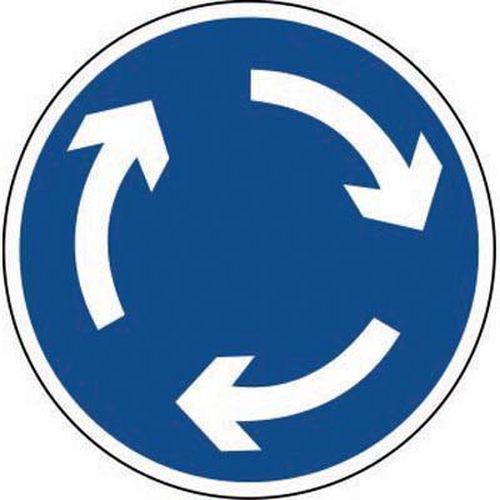}  \\
    regulatory--turn-right--g1 & \includegraphics[height=0.4cm]{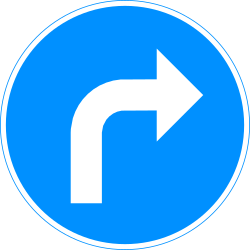} \\
    regulatory--no-stopping--g1 &\includegraphics[height=0.4cm]{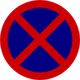} \\ 
 information--parking--g1 & \includegraphics[height=0.4cm]{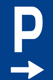}  \\ 
 warning--roadworks--g1* & \includegraphics[height=0.4cm]{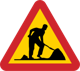}  \\
 regulatory--stop--g1 & \includegraphics[height=0.4cm]{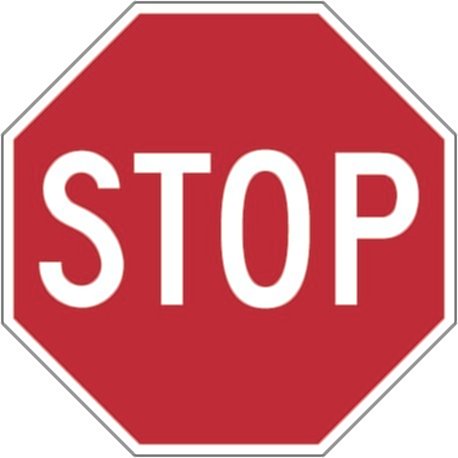}  \\ 
 regulatory--keep-left--g1 & \includegraphics[height=0.4cm]{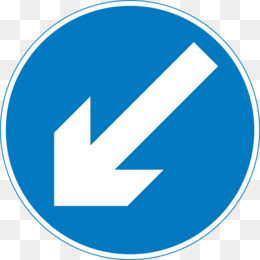} \\ 
 warning--t-roads--g1* & \includegraphics[height=0.4cm]{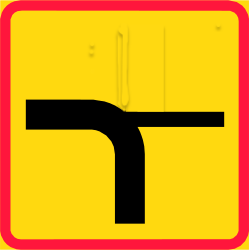}  \\
    regulatory--keep-right--g1 & \includegraphics[height=0.4cm]{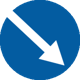}  \\ 
     regulatory--yield--g1 & \includegraphics[height=0.4cm]{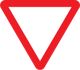} \\ complementary--buses--g1 & \includegraphics[height=0.4cm]{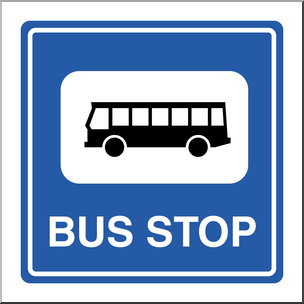} \\
    \hline
  \end{tabular}
\label{table:tfs}
\end{table}

\begin{figure*}[t!]
\centering
\begin{subfigure}{.33\textwidth}
  \centering
    \includegraphics[trim=0 0 0 0,clip, width=.99\textwidth]{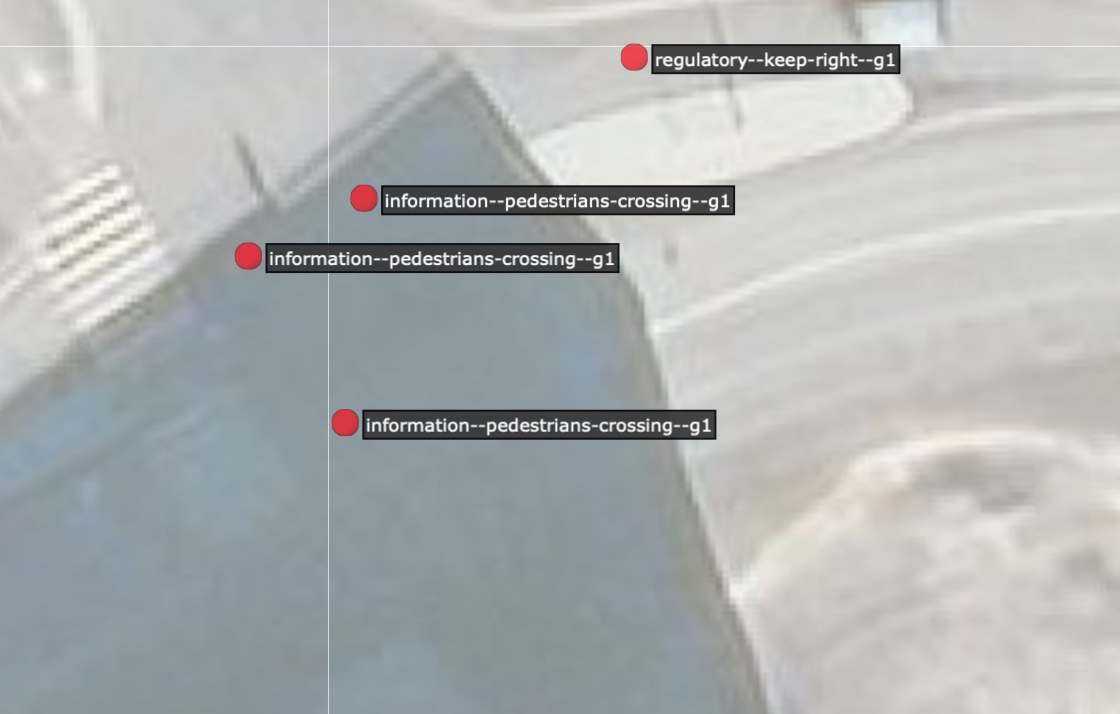}
    \caption{Prediction}
\label{fig:metaresult1}
\end{subfigure}%
\begin{subfigure}{.33\textwidth}
  \centering
    \includegraphics[trim=0 0 0 0,clip,width=.99\textwidth]{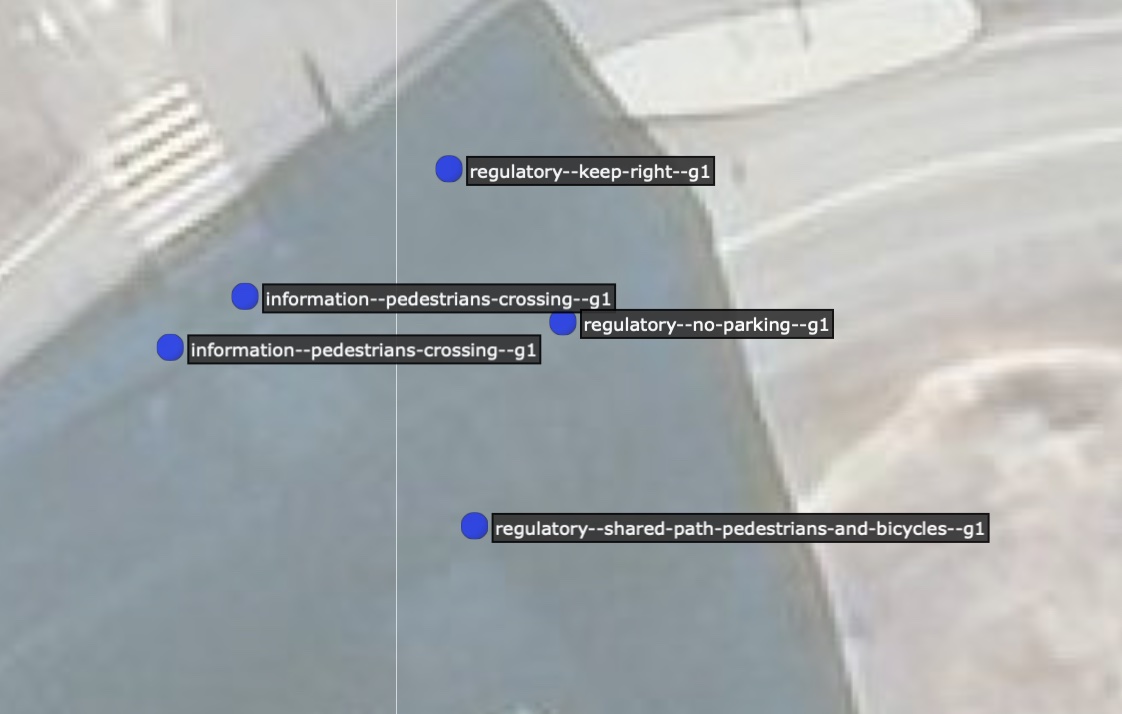}
    \caption{Ground truth}
\label{fig:metagt1}
\end{subfigure}
\begin{subfigure}{.33\textwidth}
  \centering
    \includegraphics[trim=0 0 280 0,clip,width=.99\textwidth]{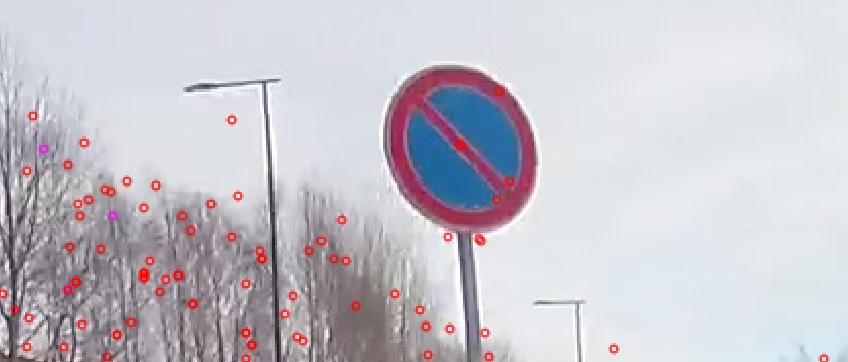}
    \caption{Features unmatched}
\label{fig:features_unmatched}
\end{subfigure}
\caption{In region G, the "no parking" traffic sign was not able to be localized. Notice its absence in a) as compared with the ground truth in b). c) shows the traffic sign in question with unmatched features across images in red circles on it. Due to these unmatched features, the triangulation of the traffic sign was not able to be performed, thus resulting in its absence in the metadata. }
\label{fig:metaresultgt1}
\end{figure*}

\begin{figure}[t!]
\centering
\begin{subfigure}{.99\textwidth}
\centering
    \includegraphics[trim=0 0 0 0,clip, width=.99\textwidth]{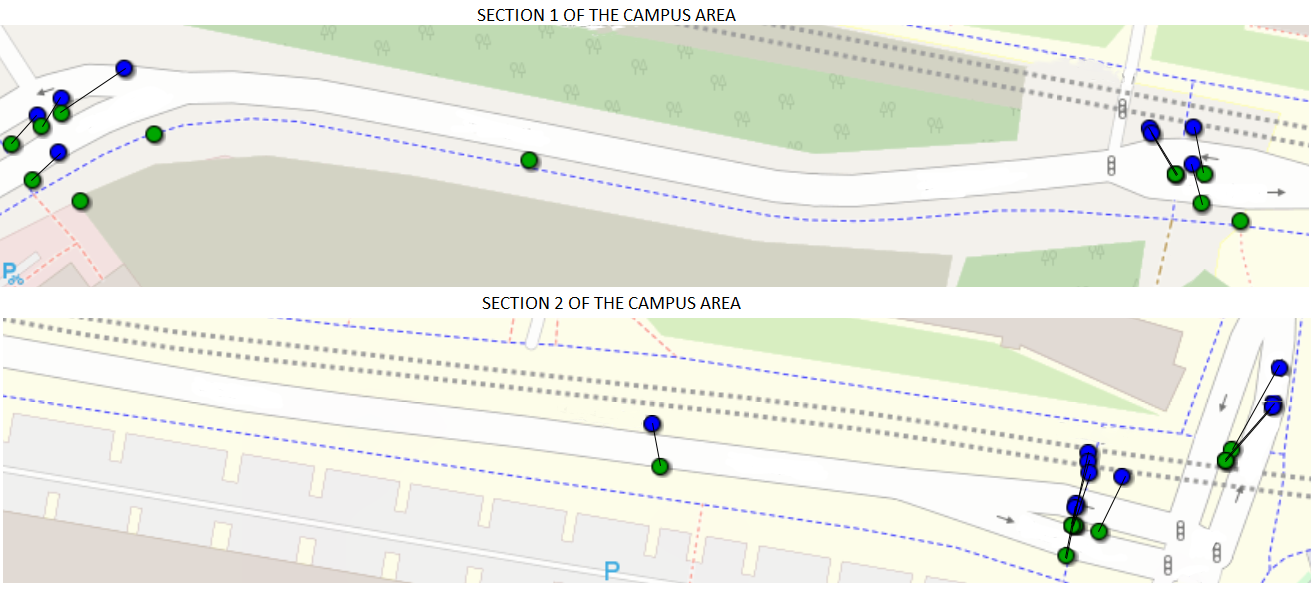}
\caption{Estimated locations of detected traffic signs (in blue) vs. the ground truth (in green) for the campus area.}
\label{fig:meta_gt}
\end{subfigure}
\\
\begin{subfigure}{.99\textwidth}
\centering
    \includegraphics[trim=0 0 0 0,clip, width=.99\textwidth]{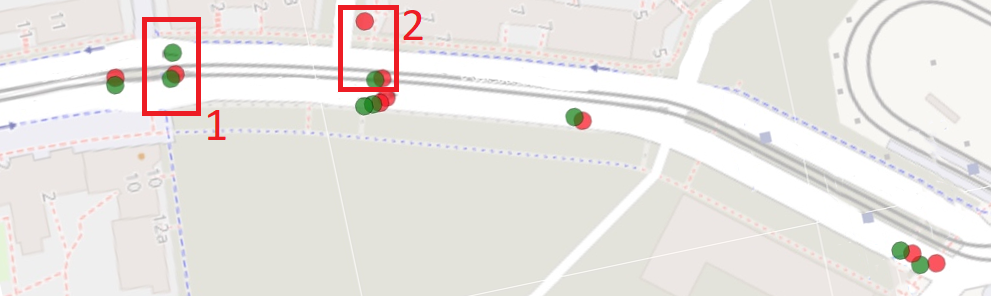}
\caption{Estimated locations of detected traffic signs (in red color) vs. the ground truth (in green color) for the residential area.}
\label{fig:metatrafficsigns}
\end{subfigure}
\caption{Traffic sign localization using the initial metadata generation method. Due to its length, the campus area is divided into two sections. Note that the road on section 2 is exactly the continuation of that on section 1.}
\end{figure}

\subsubsection{3D Object Localization} \label{sec:3dobjectlocalization}

To evaluate the accuracy of the object localization of our initial metadata generation method (Section \ref{metadatagen}, step A.3), we utilize 25 traffic signs (see Table \ref{table:tfs}) installed along the trajectories A and G as examples. The locations of these traffic signs provided by the geo-referenced and semantically segmented SfM point cloud are compared to their ground truth.

At first, it has to be noted that the system is unable to locate the traffic signs when there are insufficient 2D image keypoints in the intersection between a semantic segment and the corresponding bounding box of the object in question. In the example shown in Fig. \ref{fig:metaresultgt1} of region G, the "no parking" traffic sign has not been localized due to the quality of the SfM reconstruction (see the prediction in Fig  \ref{fig:metaresult1} vs. the ground truth in Fig. \ref{fig:metagt1}). In Fig. \ref{fig:features_unmatched}, unmatched features are shown in red color, whereas the matched ones are illustrated in purple. Even though features for the traffic sign in question have been extracted, these are not matched with the features seen in other images. Thus being unable to triangulate the position of the traffic sign and, as a consequence, resulting in the absence of the "no parking" traffic sign in the metadata. An explanation for this is that the images utilized in the SfM reconstruction have been collected in a sequential manner in only one driving direction which might not provide a sufficient overlap between the views. Still, this demonstrates a realistic scenario. 

\begin{figure*}[t]
\centering
\begin{subfigure}{.49\textwidth}
  \centering
    \includegraphics[trim=0 40 0 0,clip, width=.99\textwidth]{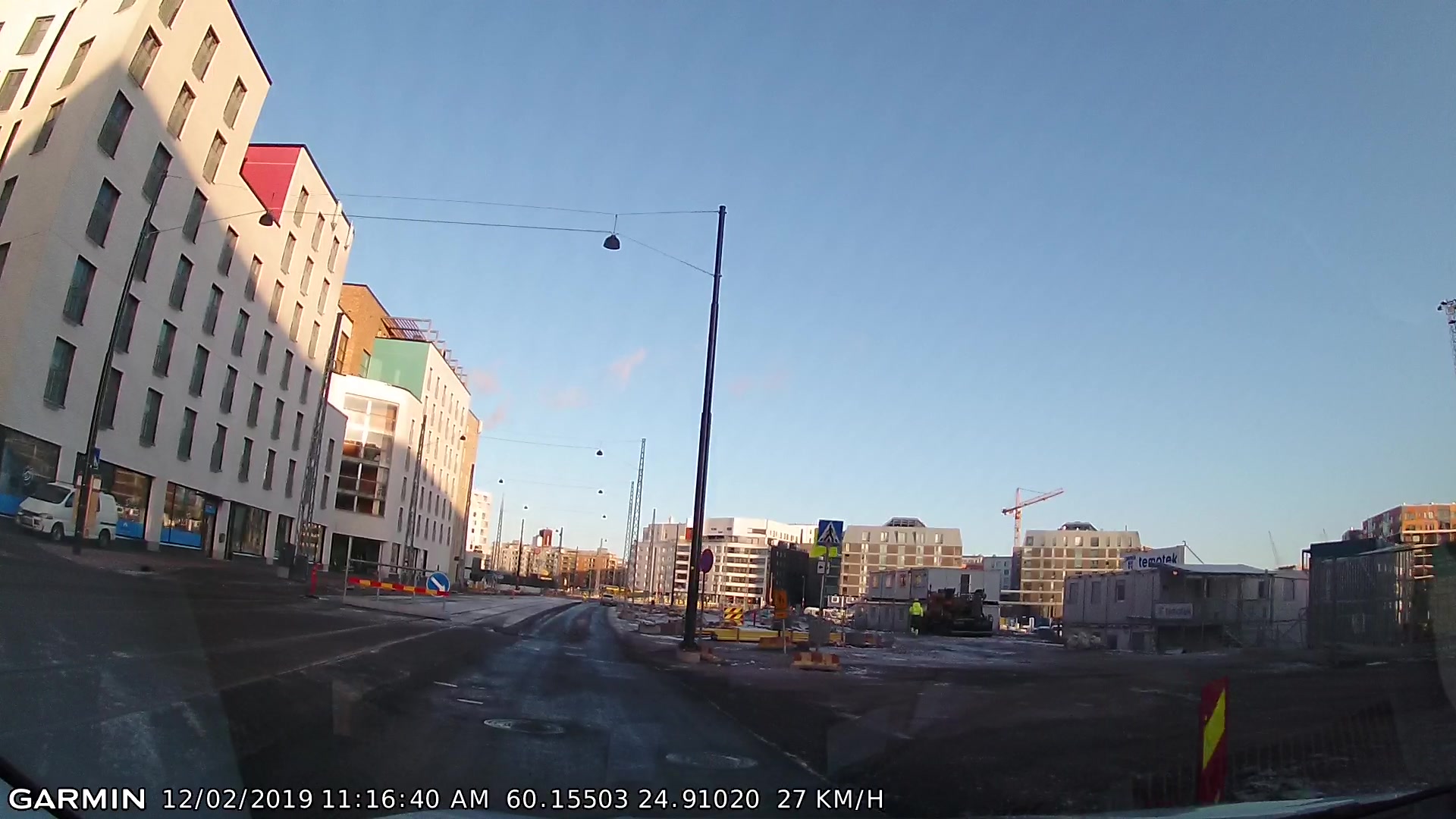}
    \caption{Original image of Fig. \ref{fig:occluded2} (Dataset III in Table \ref{table:datasets})}
\label{fig:occluded1}
\end{subfigure}%
\begin{subfigure}{.49\textwidth}
  \centering
    \includegraphics[trim=0 51 0 35,clip,width=.99\textwidth]{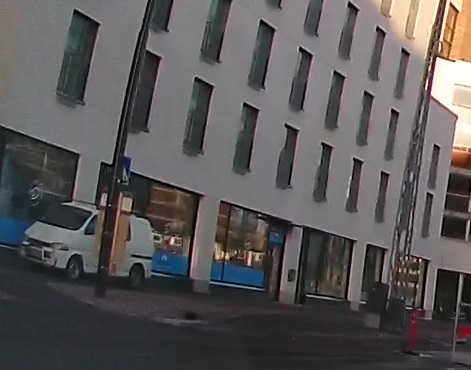}
    \caption{Zoomed in cropped image of Fig. \ref{fig:occluded1}}
\label{fig:occluded2}
\end{subfigure}
\\
\begin{subfigure}{.49\textwidth}
  \centering
    \includegraphics[trim=0 40 0 0,clip, width=.99\textwidth]{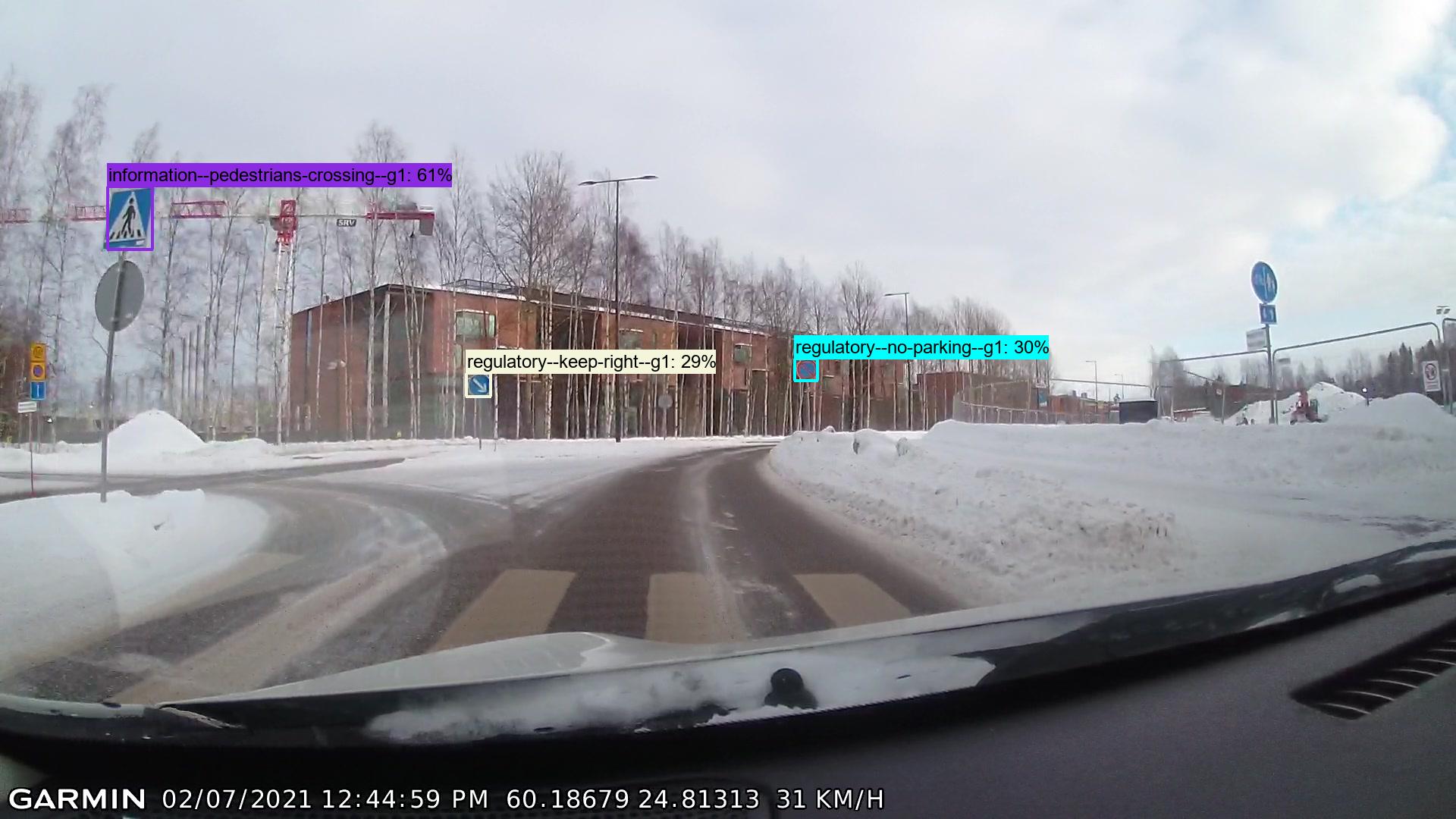}
    \caption{An example with Pedestrians and bicycles traffic sign (Dataset IV in Table \ref{table:datasets})}
\label{fig:objdetex}
\end{subfigure}%
\begin{subfigure}{.49\textwidth}
\centering
    \includegraphics[trim=0 40 0 0,clip, width=.99\textwidth]{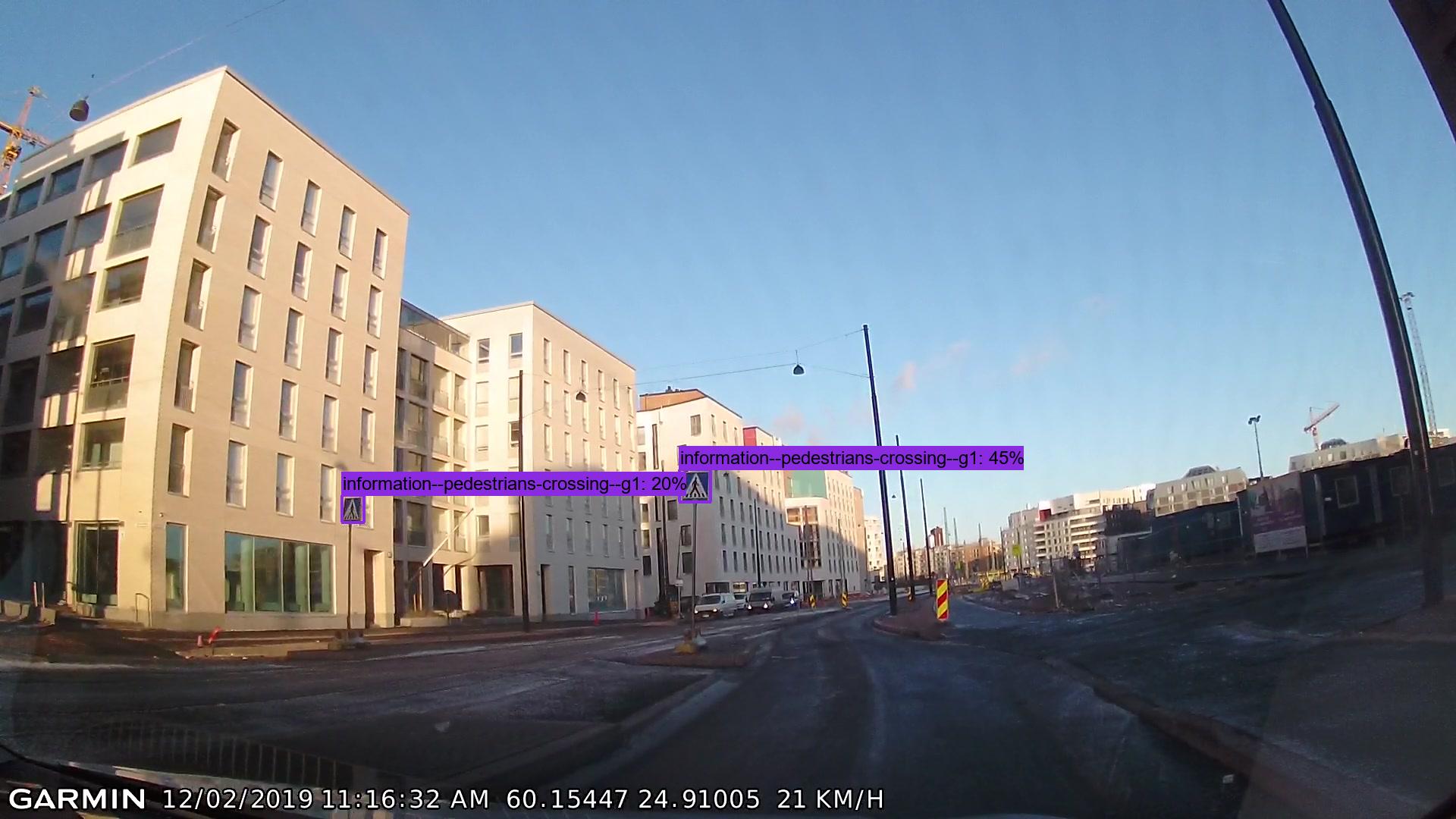}
\caption{An example of two traffic signs of the same type located closely to each other (Dataset III in Table \ref{table:datasets})}
\label{fig:pedestrianwalk}
\end{subfigure}
\caption{An example of traffic signs which are not clearly visible due to a long distance from camera or occlusion.}
\label{fig:occluded}
\end{figure*}

In region A (inside rectangle 1 of Fig. \ref{fig:metatrafficsigns}), it can be observed the absence of one traffic sign (note the presence of two green circles compared to one sole red circle). The traffic sign that was unable to be localized is of type pedestrian walk (Fig. \ref{fig:pedestrianwalk}). Its absence is explained by the fact that there are two traffic signs of the same type located very near each other. Our system mistakes these two instances as only one. It is an effect of the distance threshold for clustering $T_D$ (Algorithm \ref{alg:meta_gen}). To compensate for the localization errors of the SfM point cloud, this threshold imposes that instances of the same type are clustered as one sole instance. Therefore, a more accurate localization by the SfM method is one solution for this issue. However, a more ingenious approach would consist of tracking each traffic sign across multiple images to identify which points in the point cloud belong to the same traffic sign and thus differentiate between such closely located objects of same class.

In rectangle 2 of Fig. \ref{fig:metatrafficsigns}, one traffic sign was localized twice (notice the presence of two red circles compared to only one green circle). This traffic sign was located once in an accurate location, but also far from the ground truth due to the inaccuracy of the SfM reconstruction. One possible solution is to increase the number of viewing angles and the image quality of the dataset used for the SfM reconstruction. Some traffic signs have not been localized due to object detection confidence for specific classes being lower than the threshold. An example of this is the "junction with a side road" traffic sign as shown in Fig. \ref{fig:mistakes_od}. Reducing the threshold is not a solution as it would result in a multitude of erroneous predictions. In future work, training the objection detection model with a larger dataset would help to reduce this error.



Even though the metadata may miss traffic signs, this can be corrected in the change detection stage (described in Section \ref{changedetectionmethod}). The detection of an unseen traffic sign in the initial stage will trigger the correction of the metadata by the change detection method. In the future, the work can be extended to include pedestrians or cyclists carrying smartphones and filming the environment to improve the localization of traffic signs along the pedestrian path, since sometimes these signs might be quite far from the main road or occluded by trees, street poles or other objects (see examples in Fig. \ref{fig:occluded}). This specific case has been the cause of 2 errors out of a total 5. 

In quantitative measures, our system locates 8 out of 9 traffic signs along the driving direction in the residential area, and 16 out of 20 traffic signs in the university area. Compared with the ground truth, in the campus area, the median distance error is 10.4 meters with a standard deviation of 2.9 meters. As for the residential area, the median distance error and standard deviation were measured to be 3.6 meters and 1.4 meters, respectively. We posit that this disparity between the campus and residential area is due to the fact that the former consists of a more featureless region (i.e. devoid of objects such as a building), which affects negatively the feature extraction and feature matching processes in the reconstruction of the model. We also look into the causes of the localization errors in the successful cases. These errors have been mainly caused by the errors in the SfM-based 3D reconstruction and geo-registration. In our experiments, the data was collected from vehicles driving through the trajectories following one direction rather than two. We used up to 3 cameras facing the front for data collection. More cameras facing different directions would help reduce the error in the point cloud segmentation by improving the accuracy of depth prediction for the 3D points. In addition, the GPS coordinates used for the geo-registration were selected manually by visually analyzing the image and approximating its location using Google Maps and Google Street Maps. This always introduces a certain degree of human error. 


\section{Evaluation of Change Detection} \label{eval_our}

A number of changes has been observed in terms of traffic sign deployment in the test areas between Day 1 and Day 2, and between Day 3 and Day 4. In this section, we evaluate the performance of the change detection method using 4 datasets. Dataset III (Day 1) and Dataset IV (Day 2) are used to compare the changes in the residential area, while Dataset VI (Day 3) and Dataset VII (Day 4) for evaluation of the system in the campus area. Particularly, we will measure the accuracy of Step B.1.1, Step B.2, and Step B.3, respectively. 

\subsection{Camera Pose Estimation}
\label{sec:camera_pose_estimation}

Our camera pose estimation (Step B.1.1.) is comprised of two distinct methods. The first method - corresponding to the registration of the image into the point cloud - presents an accuracy of 7.79 meters with respect to the camera pose estimation. This result is the same as the one presented in Section \ref{sec:sfmrecons} since the point cloud itself is also used here to estimate the camera position. As the ground truth for the camera orientation is not available for any image, it is not possible to measure the accuracy of COLMAP when estimating the camera orientation of the images. 


The second method of the camera pose estimation presented an average error of 6.22 meters with respect to the camera position. This result was obtained by comparing directly the position of the car given by the GPS device with that provided by the RTK system. The data utilized for the measurement of the camera position error comprised of the entire residential area. Again, as mentioned in Section \ref{sec:sfmrecons}, due to the inconsistent sampling rates of the GPS (30Hz) and the RTK system (1Hz), the actual error may be lower. Regarding the camera orientation estimation - given by the assumption given in Eq. \ref{eq:cpe} - we have measured its validity by comparing the results given by it with those provided by COLMAP during the image registration. The results showed that the assumption - on which the second method is based - is able to estimate the camera orientation with approximately 6.18 degrees of error - which represents a fairly good result. The measurement of the camera orientation error utilized the data of the campus area.

\begin{figure*}[bt!] 
  \centering
    \includegraphics[width=\textwidth]{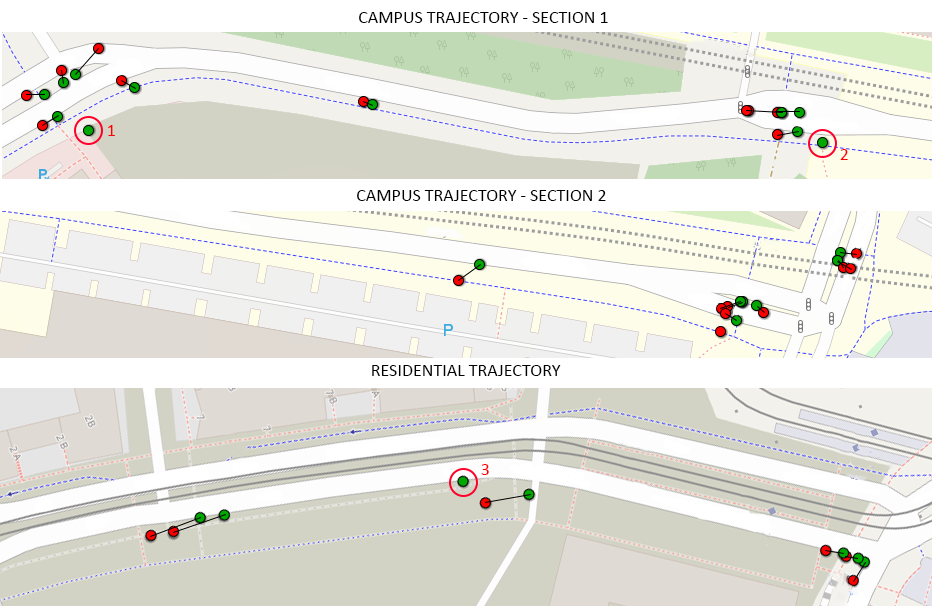}
    \caption{Estimated traffic sign locations of the real-time layer (in red) vs. the ground truth (in green). The lines linking a green circle to a red one indicate a match between the ground truth and the real-time layer. The encircled traffic signs represent unmatched cases, each of which with a number as the identifier.}
\label{fig:all_tfloc}
\end{figure*}

\begin{figure*}[bt!] 
  \centering
    \includegraphics[width=\textwidth]{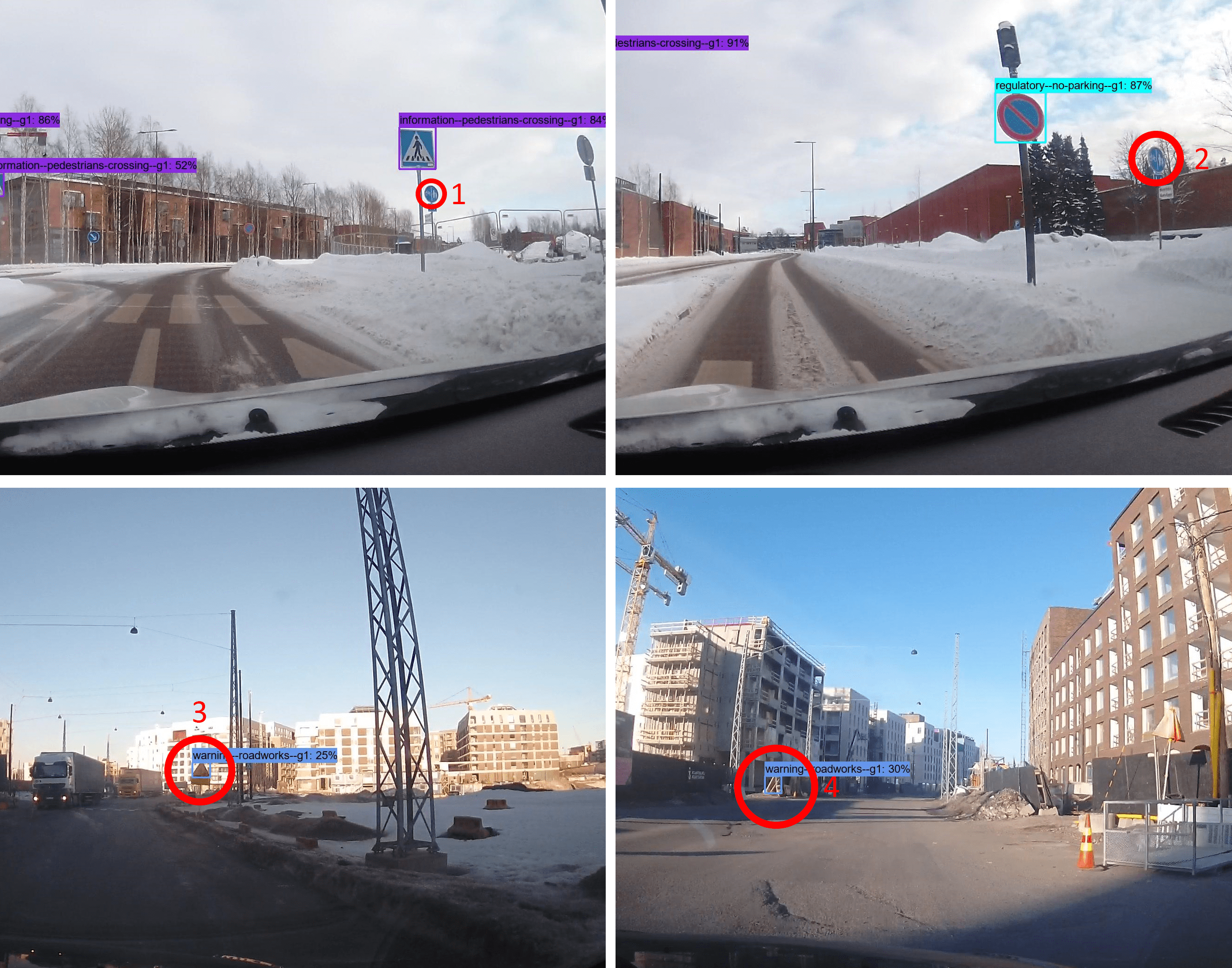}
    \caption{Analysis of the traffic sign detection in the campus and residential areas. \textbf{Top left and top right} (Dataset IV in Table \ref{table:datasets}): traffic signs of shared path between pedestrians and bicycles that were not detected due to their distance to the camera as well as the insufficient training data for this traffic sign in the object detection neural network - campus area. \textbf{Bottom left} (Dataset VII): the sole undetected traffic sign classified with a 25\% confidence score - below the confidence score threshold. \textbf{Bottom right} (Dataset VII): a case where barricades are wrongly detected and classified as a roadworks traffic sign with 30\% confidence score. This explains why there should exist a threshold of confidence scores to consider.}
\label{fig:all_analysis}
\end{figure*}

\begin{table}[b]
\begin{tabular}{|l|c|c|c|c|c|c|c|c|c|c|c|c|}
\hline
\multicolumn{1}{|c|}{\multirow{2}{*}{\textbf{Range}}} & \multicolumn{3}{c|}{\textbf{0 - 10m}}   & \multicolumn{3}{c|}{\textbf{10 - 25m}}  & \multicolumn{3}{c|}{\textbf{25 - 50m}}  & \multicolumn{3}{c|}{\textbf{50 - 100m}} \\ \cline{2-13} 
\multicolumn{1}{|c|}{}                                & \textbf{D.} & \textbf{H.} & \textbf{L.} & \textbf{D.} & \textbf{H.} & \textbf{L.} & \textbf{D.} & \textbf{H.} & \textbf{L.} & \textbf{D.} & \textbf{H.} & \textbf{L.} \\ \hline
Abs. error                                            & 3.33m       & 1.01m       & 1.73m       & 3.55m       & 0.99m       & 1.54m       & 6.42m       & 1.22m       & 2.55m       & 12.58m      & 1.66m       & 4.42m       \\ \hline
Rel. error                                            & 0.62        & 3.12        & 3.21        & 0.24        & 1.53        & 2.12        & 0.19        & 2.18        & 2.60        & 0.19        & 2.50        & 1.45        \\ \hline
\end{tabular}
\caption{Depth, lateral and height errors of the pixel-wise 3D localization method for different ranges of ground truth distances. Abbreviations: D for depth, H for height, and L for lateral.}
\label{table:bts_results}
\end{table}

\subsection{Pixel-wise 3D Localization} 

As described in Section \ref{changedetectionmethod}, pixel-wise 3D localization (Step B.1.3.) is built on top of the BTS \cite{lee2019big} network. The network is initialized with the weights trained with Kitti depth dataset \cite{geiger2012we}. After that, we have fine-tuned the network with 13032 samples collected from different regions of the environment (A - D and F - G in Fig. \ref{fig:jatkasaari} and Fig. \ref{fig:otaniemi}) for 30 epochs. The validation set was comprised of 706 samples from region A collected on a different day than those samples from the same region present in the training set. Finally, the test set consisted of 576 samples collected in region E. 


 Table \ref{table:bts_results} shows the error results on the test set on two distinct metrics: the absolute error and the relative error. The absolute error is defined as the mean (calculated on all pixels of all images) of the absolute error (Eq. \ref{eq:abserror}), whereas the relative error is the mean (calculated on all pixels) of the ratio between absolute error and the ground truth (Eq. \ref{eq:relerror}). 
 
  \begin{equation}
     \text{Absolute error} = \sum_{image}^{n_i}  \sum_{pixel}^{n_p} \dfrac{|PRED - GT|}{n_i \cdot n_p}
     \label{eq:abserror}
 \end{equation}
 
  \begin{equation}
     \text{Relative error} = \sum_{image}^{n_i}   \sum_{pixel}^{n_p} \dfrac{|PRED/GT - 1|}{n_i \cdot n_p}
     \label{eq:relerror}
 \end{equation}
 
  \noindent
 where PRED and GT stand for the prediction of the neural network and its ground truth, respectively. Also, $n_p$ and $n_i$ refer to the number of pixels in an image and the number of images in the test set, respectively.
 
 It is observed that the absolute errors grow as the ground truth distance increases, whereas the relative errors decrease. Compared to the original BTS \cite{lee2019big}, our relative error in depth is approximately 2.2 - 4.5 times larger. This is due to the following reasons. Note that we have applied data augmentation methods such as color, gamma, and brightness changes to reduce over-fitting. However, they were not effective in significantly reducing the error. The potential solutions to improve the accuracy of the SfM point cloud will be discussed in Section \ref{discussion}.
 
\begin{itemize}
    \item Most of the weights in the neural network are shared between three tasks of distance estimation: depth, height, and lateral. This worsens the performance for the depth estimation since the learned features need to be more generic for a better estimation of the three distances together.
    \item Our dataset is comprised of samples captured with three different cameras, whereas in the original work of BTS \cite{lee2019big} the authors trained separate neural networks for the samples of each camera available. It is known that differences in the camera directly affect the performance of computer vision algorithms.
    \item Our dataset is sparse, i.e. most of the pixels in the images are unlabeled. This is due to the fact that the point clouds are also sparse. 
    \item The ground truth is formed by estimations provided by the SfM reconstruction, thus containing errors that affect both the training of the neural network and the accurate evaluation of the trained model on the test set.
\end{itemize}
 

\subsection{Traffic Sign Localization} 

We perform the online traffic sign localization (Step B.2.) on both campus and residential trajectories. Dataset IV and VII were utilized for the former and the latter trajectories, respectively. Since the accuracy of the 3D object localization is dependent on the distance between the camera and the object (Table \ref{table:bts_results}), we propose to set a lower and a higher distance threshold. This signifies discarding traffic signs localized at a distance to the camera smaller than the lower threshold and bigger than the higher threshold. By carefully selecting the values of the distance thresholds, it is possible to achieve better results.

\textbf{Residential trajectory.} Figure \ref{fig:all_tfloc} illustrates the estimated traffic sign locations and their corresponding ground truth for the residential trajectory. It is observed that our system can locate 6 out of the 7 traffic signs. The missed traffic sign was detected with low confidence - below the considered threshold in the object detection neural network - and thus discarded (Figure \ref{fig:all_analysis}). We hypothesize that the reason for this is due to the insufficient number of labeled examples of construction work in the training dataset of the object detection neural network. Reducing the confidence score threshold is not a solution, since it results in the appearance of multiple erroneous detections like the one shown in Figure \ref{fig:all_analysis} (bottom right) where construction barricades are wrongly detected as a traffic sign of construction-work type. With respect to distance metrics, the traffic signs were located with a median distance error of 9.1 meters and 5.1 meters. A detection accuracy of 83.3\% is seen for this trajectory.

\textbf{Campus trajectory.} Figure \ref{fig:all_tfloc} illustrates the estimated traffic sign locations and their corresponding ground truth for the campus trajectory. 2 of the traffic signs present in the ground truth and without matches from the real-time layer are of the same type: shared path between pedestrians and bicycles (cases 1 and 2). These mismatches are due to the failure of the object detection network in detecting this specific type of traffic sign (Figure \ref{fig:all_analysis}). Again, we posit that a larger training set for the objection detection neural network could solve this issue. Also, we consider these minor failures since these traffic signs are not addressed to the driver. Overall, a median error of 5.27m and a standard deviation of 2.08m are observed for the trajectory. The considerable difference between these values for the campus trajectory compared with the residential trajectory is caused by the fact that the images of the campus trajectory are more aligned with those that composed the training set of the 3D object localization neural network.

\begin{figure*}[bt!] 
  \centering
    \includegraphics[width=\textwidth]{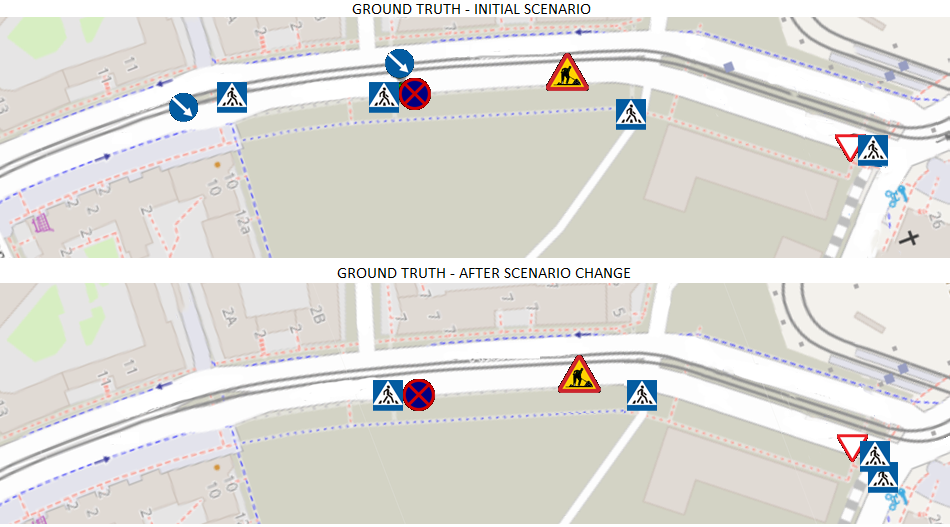}
    \caption{Ground truth of the before (top) and after (bottom) scenario change in the residential area.}
\label{fig:cd_jatkasaari_before_after}
\end{figure*}

\begin{figure*}[bt!] 
  \centering
    \includegraphics[width=\textwidth]{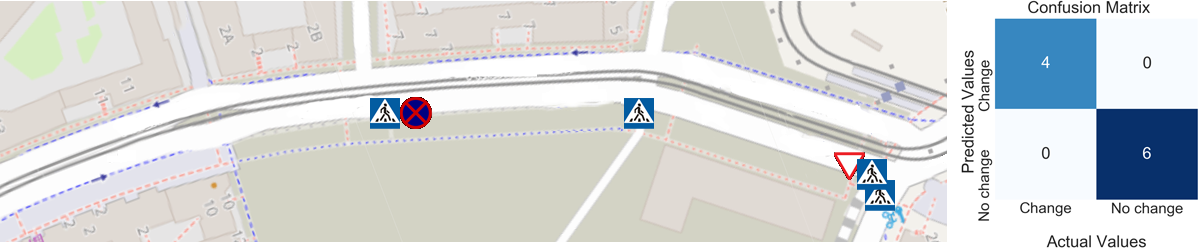}
    \caption{Detected and localized traffic signs of the real-time layer along the residential area with the confusion matrix detailing the results of the change detection.}
\label{fig:cd_jatkasaari_online}
\end{figure*}

\begin{figure*}[bt!] 
  \centering
    \includegraphics[width=\textwidth]{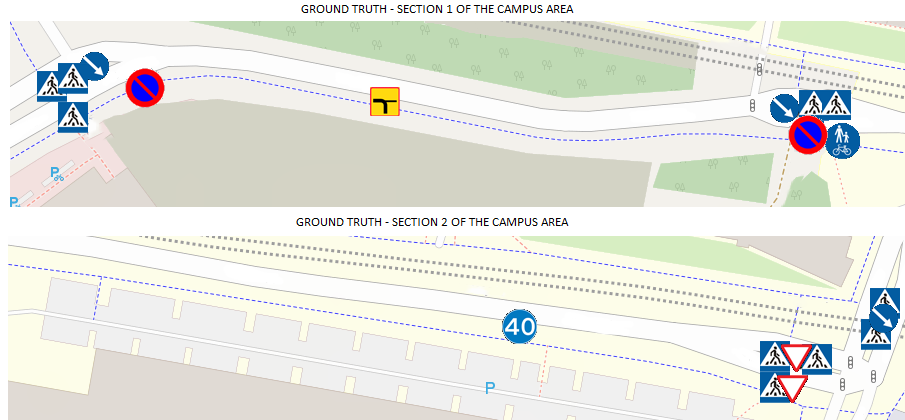}
    \caption{Ground truth of the campus area. Note that this area does not suffer any change.}
\label{fig:cd_otaniemi_gt}
\end{figure*}

\begin{figure*}[bt!] 
  \centering
    \includegraphics[width=\textwidth]{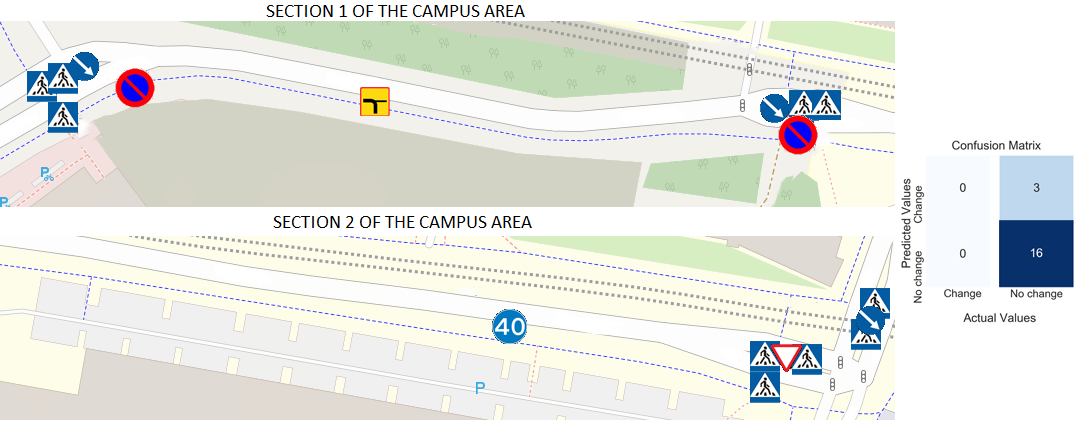}
    \caption{Detected and localized traffic signs of the real-time layer along the campus area with the confusion matrix detailing the results of the change detection.}
\label{fig:cd_otaniemi_online}
\end{figure*}

These are solid results considering that only monocular images were used to predict the location of the changes and the labels of the training dataset utilized to train the prediction neural network were automatically generated. Although sub-meter level accuracy was shown to be possible in change localization\cite{slamcu}, it requires the utilization of a set of additional sensors such as LiDAR, IMU, vehicle speed sensors, and highly accurate positioning solutions. We envision that, whenever a change is detected and localized, a cautionary area of a predetermined radius encompassing it is created. This signifies that, in real life, changes can be indicated as an area - instead of a point - to incorporate the inaccuracy of the change localization algorithm. In the circumstance of an autonomous vehicle entering this area, an immediate switch from automatic to manual operation mode is required. Considering this concept, highly accurate change localization results are not required.

\subsection{Change Detection and Localization} 

When a traffic sign is detected and localized in the second stage (real-time layer, step B.2), the system searches around its location for matching traffic signs in the current copy of the metadata at the temporary layer (step B.3). Since there exist localization errors in the metadata and in the online traffic sign location, we define another distance threshold for reducing the false-positive errors in change detection. The newly detected traffic sign is considered to match an existing one in the current copy of metadata, if they are of the same type and the distance between them is below the threshold. When a mismatch happens, a change is detected and reported. To determine this threshold, we consider that there can be an error of up to 10 meters in the localization of the traffic signs in the metadata as well as in the real-time layer. Also, taking into consideration a worst-case scenario where the total localization error of a traffic sign in the metadata compared to the real-time layer is doubled up, we determine the threshold value to be 20 meters.

\textbf{Residential trajectory.} 
Figure \ref{fig:cd_jatkasaari_before_after} illustrates the arrangement of the traffic signs (ground-truth) in the residential region before and after the environment has suffered changes. Figure \ref{fig:cd_jatkasaari_online} illustrates the environment as seen by the real-time layer as well as the confusion matrix related to the detection of changes in the environment from the comparison with the metadata. It can be observed that the removal of 4 traffic signs was correctly detected by our algorithm. Moreover, the permanence of 6 traffic signs after the scene change was also accurately identified. Note that the road-works traffic sign was not detected by the real-time layer. This is also true for the metadata, hence no change with respect to this specific traffic sign is detected - which is also the case observed in the ground-truth before and after the scene change. The confusion matrix shown in Figure \ref{fig:cd_jatkasaari_online} exhibits that our change detection algorithm reached the maximum possible performance.

\textbf{Campus trajectory.} Figure \ref{fig:cd_otaniemi_gt} illustrates the arrangement of the traffic signs (ground-truth) in the campus region. Even though this area does not suffer any change, our change detection algorithm is required to identify the absence of changes. Similar to the residential area, Figure \ref{fig:cd_otaniemi_online} illustrates the environment as seen by the real-time layer as well as the confusion matrix. The comparison with the metadata reports the appearance of two traffic signs - no-parking and T-junction - and the removal of a yield traffic sign. This represents three erroneous cases of change in the environment. However, notice that the appearance in the real-time layer of traffic signs that were erroneously not included in the metadata - which is the case of the no-parking and T-junction traffic signs - signifies that the metadata can be corrected. It is also observed that 16 out of the 20 traffic signs were correctly reported as objects of no change. Overall, the change detection method exhibits an accuracy of 85\%.

\subsection{Change Detection and Localization}

To measure its latency, we executed the real-time layer on an Intel(R) i7-11700F processor clocked at 2.50GHz and an NVIDIA RTX 3070 8GB GPU. Among the processes in the real-time layer, the pixel-wise 3D localization (Step B.1.1) and the object detection (B.1.2) were the most computationally expensive ones by lasting approximately 0.10s and 0.06s per image, respectively. The camera pose estimation (B.1.3) performed by the first method (discussed in Section \ref{sec:camera_pose_estimation}) took approximately 0.05s per image. The second method of camera pose estimation, since it involves very few operations of multiplications, was shown to be of negligible cost as well as the online 3D object localization (B.2) and metadata comparison for change detection (B.3). Overall, the real-time layer was able to process on average 5 FPS. In our tests, steps B.1.1, B.1.2 and B.1.3 were executed in a sequential manner. Since they are independent on each other, in future work it is possible to implement them in parallel on the GPU for better resource utilization.

\section{Related Work} \label{related}

In this section we first review the latest works in the literature related to change detection while comparing their design choices with our pipeline - Table \ref{tab:cd_comparison} provides an overview of the comparisons. Since semantic mapping (metadata generation and maintenance) and monocular depth estimation are also part of our pipeline, we also describe the literature on these topics.

\begin{figure*}[t]
\centering
\begin{subfigure}{.195\textwidth}
  \centering
  \includegraphics[width=.99\textwidth]{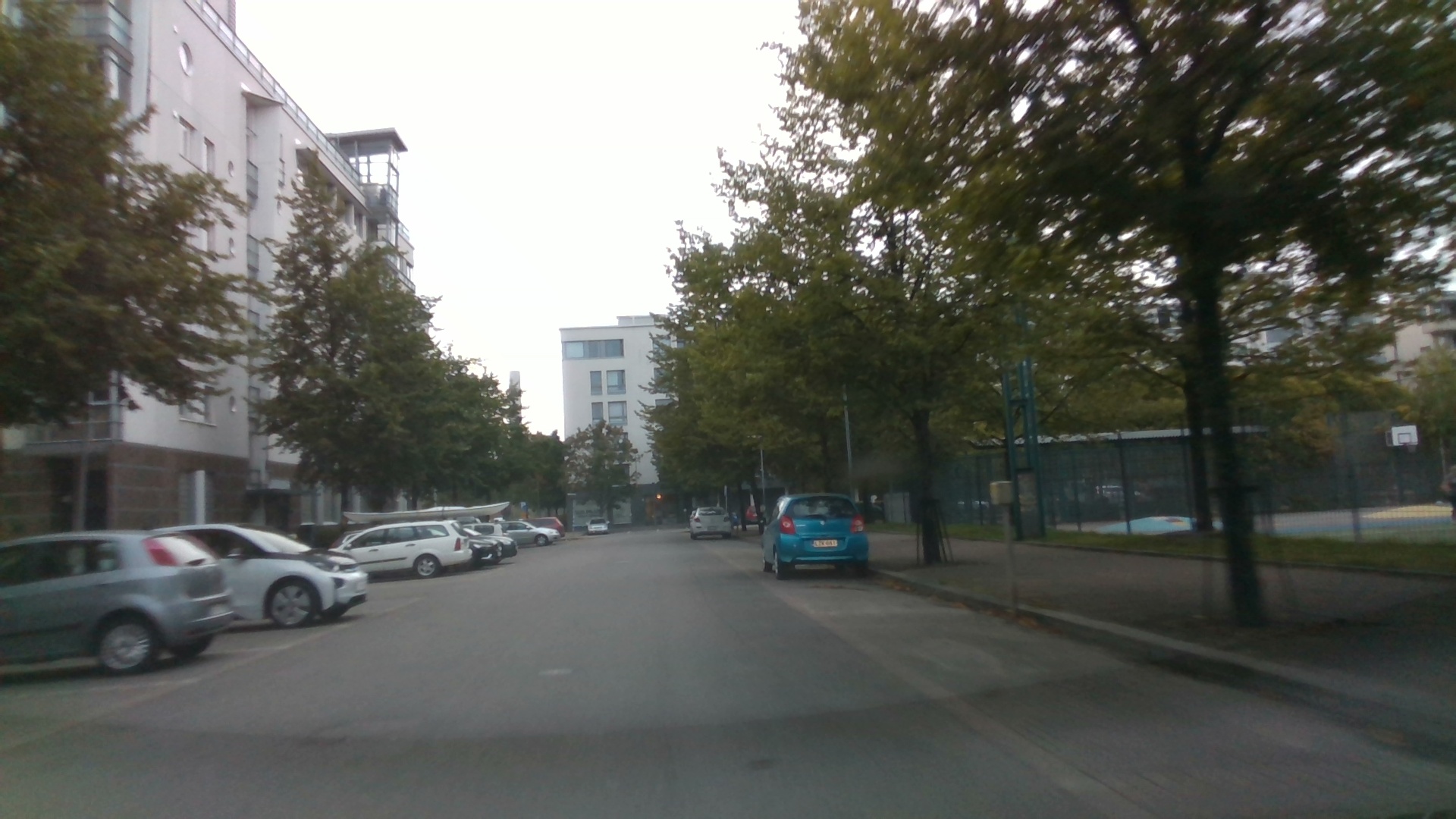}
  \caption{Before}
  \label{fig:sub1}
\end{subfigure}%
\begin{subfigure}{.195\textwidth}
  \centering
  \includegraphics[width=.99\textwidth]{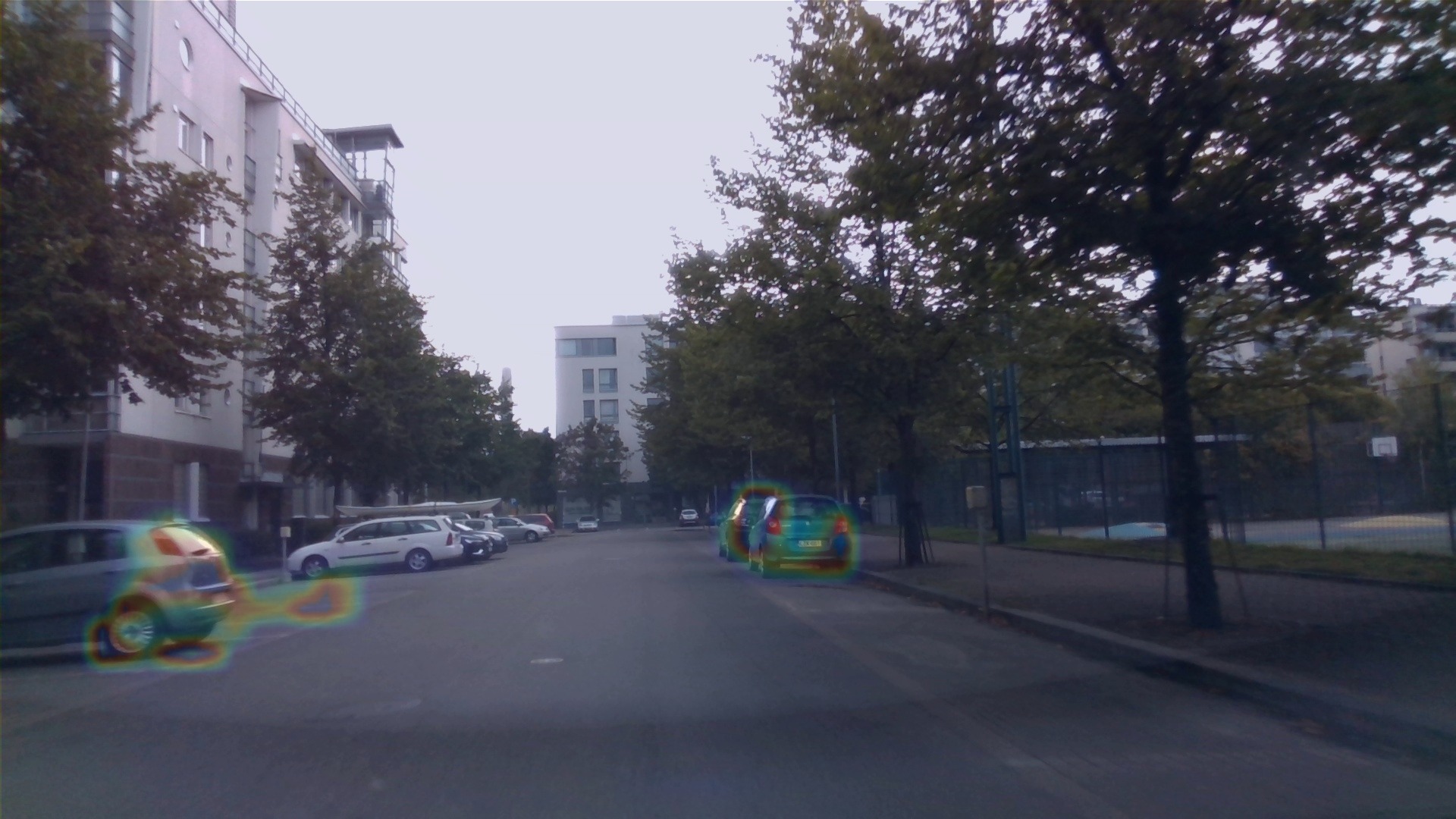}
  \caption{After}
  \label{fig:sub2}
\end{subfigure}
\begin{subfigure}{.195\textwidth}
  \centering
  \includegraphics[width=.99\textwidth]{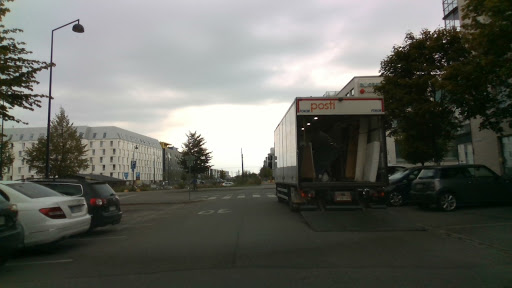}
  \caption{Original}
  \label{fig:monodepth1}
\end{subfigure}%
\begin{subfigure}{.195\textwidth}
  \centering
  \includegraphics[width=.99\textwidth]{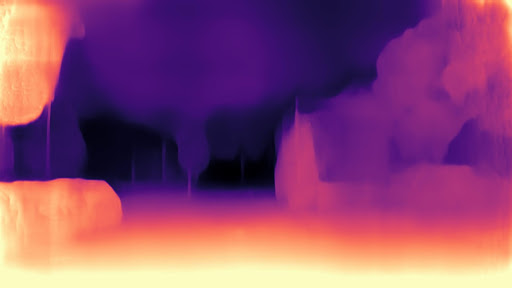}
  \caption{Before training}
  \label{fig:monodepth2}
\end{subfigure}%
\begin{subfigure}{.195\textwidth}
  \centering
  \includegraphics[width=.99\textwidth]{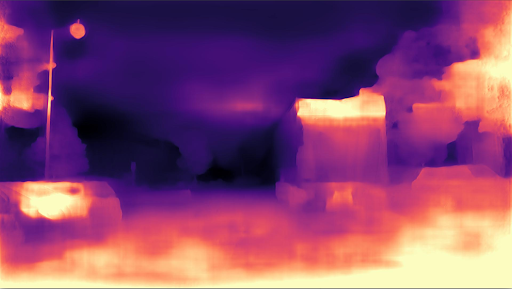}
  \caption{After training}
  \label{fig:monodepth3}
\end{subfigure}%
\caption{Example results of SceneChangeDet based change detection and Monodepth based depth estimation. Changes detected by SceneChangeDet in (b) compared with (a) are highlighted with colors. Given an original image (c), the results of the initial Monodepth model trained on the Kitti dataset (d) are compared with the model retrained on our datasets (e).}
\label{fig:monodepth}
\end{figure*}

\noindent
\subsection{Change Detection and Localization} 

\begin{table}[]
\begin{tabular}{|l|c|c|c|c|c|c|c|}
\hline
\multicolumn{1}{|c|}{\textbf{Method}} & \textbf{\begin{tabular}[c]{@{}c@{}}Requires \\ dense \\ point cloud\end{tabular}} & \textbf{\begin{tabular}[c]{@{}c@{}}Requires \\ manual \\ labeling\end{tabular}} & \textbf{\begin{tabular}[c]{@{}c@{}}Required \\ sensors\end{tabular}}                                                                  & \textbf{\begin{tabular}[c]{@{}c@{}}For \\ long-lasting\\ changes\end{tabular}} & \textbf{\begin{tabular}[c]{@{}c@{}}Performs \\ 3D change\\ localization\end{tabular}} & \textbf{\begin{tabular}[c]{@{}c@{}}Accuracy of \\ 3D change\\ localization\end{tabular}} & \textbf{\begin{tabular}[c]{@{}c@{}}Required \\ number of \\ image views\end{tabular}} \\ \hline
\begin{tabular}[c]{@{}l@{}}Palazzolo and \\  and Stachniss \cite{palazzolo2017change}    \end{tabular}                           & Yes                                                                               & No                                                                              & RGB images                                                                                                                            & Yes                                                                                & No                                                                                    & -                                                                                        & Several                                                                               \\ \hline
Alcantarilla  \textit{et al.} \cite{alcantarilla2018street}                                 & Yes                                                                               & Yes                                                                             & \begin{tabular}[c]{@{}c@{}}RGB images,\\ IMU, GPS\end{tabular}                                                                        & Yes                                                                                & No                                                                                    & -                                                                                        & Few                                                                                   \\ \hline
SceneChangeDet \cite{guo2018learning}                        & No                                                                                & Yes                                                                             & RGB images                                                                                                                            & No                                                                                 & No                                                                                    & -                                                                                        & Few                                                                                   \\ \hline
    Zhang \textit{et al.} \cite{zhangchangedetection}   & No                                                                                & No                                                                              & \begin{tabular}[c]{@{}c@{}}RGB images,\\ IMU, GPS, \\ steering \\ wheel angle, \\ vehicle speed\end{tabular}                          & Yes                                                                                & Yes                                                                                   & -                                                                                        & Few                                                                                   \\ \hline

\begin{tabular}[c]{@{}l@{}}Hu  \textit{et al.}  \cite{diffnet} \\ Wang  \textit{et al.}  \cite{transcd} \\ Santos  \textit{et al.}  \cite{scdmultiscale}  \end{tabular}
                                    & No                                                                                & Yes                                                                             & RGB images                                                                                                                            & Yes                                                                                & No                                                                                    & -                                                                                       & Few                                                                                   \\ \hline
Yew and Lee  \cite{cityscalecd}         & No                                                                                & No                                                                              & RGB images                                                                                                                            & Yes                                                                                & Yes                                                                                   & -                                                                                        & Few                                                                                   \\ \hline
 Jo \textit{et al.} \cite{slamcu}                               & No                                                                                & No                                                                              & \multicolumn{1}{l|}{\begin{tabular}[c]{@{}l@{}}RGB images,\\ LiDAR, IMU,\\ RTK, steering\\ wheel angle,\\ vehicle speed\end{tabular}} & Yes                                                                                & Yes                                                                                   & Under 1m                                                                                & Few                                                                                   \\ \hline
Ours                                  & No                                                                                & No                                                                              & \begin{tabular}[c]{@{}c@{}}RGB images,\\ GPS\end{tabular}                                                                             & Yes                                                                                & Yes                                                                                   & Under 7m                                                                                     & Few                                                                                   \\ \hline
\end{tabular}
\caption{Comparison of our work with previous studies on change detection. Although the work of Yew and Lee \cite{cityscalecd} and Zhang \textit{et al.} \cite{zhangchangedetection} can localize changes in a three-dimensional space, the authors have not evaluated the performance of this metric in question.}
\label{tab:cd_comparison}
\end{table}

Palazzolo and Stachniss \cite{palazzolo2017change} proposed an approach for spotting differences between an existing 3D model of an environment and a small sequence of images recorded in the environment. Their approach requires that the locations of all images with respect to the 3D model are precisely known. In addition, it requires dense point clouds as input, which are challenging to generate from images captured from vehicle-carried cameras with limited viewing angles. The authors utilized images from Google Street Maps that were taken using a complex array of a multitude of cameras capturing the same scene from numerous perspectives. Such availability of different perspectives of the same scene is infeasible in crowdsourcing scenarios, where only one forward-pointing camera is used.

Alcantarilla \textit{et al.} \cite{alcantarilla2018street} developed a change detection system for urban scenarios. First, their approach consisted in obtaining two dense 3D reconstructions, each representing states of the environment at different times. After that, an accurate geo-registration on both point clouds was performed, which allowed the alignment of both models. This alignment was used to obtain pairs of images (each image from a different state of the environment) taken at the same location. For each pair of images, a dense convolution neural network - trained via supervision - was employed to detect changes. Their method proved effective in detecting changes under different lighting and seasonal conditions. Again, the limitation of the method is the requirement for 3D dense reconstructions of both states. Due to the high computational complexity, the execution of this method in real-time may be infeasible for certain hardware specifications. 

Rosen \textit{et al.} \cite{davidrosen} proposed a feature-based model of environmental change detection and incorporated it into graphical SLAM techniques. The method was evaluated with simulated data only. An improvement was proposed in \cite{hashemifar} where the authors modified ORB-SLAM - a state-of-the-art visual SLAM algorithm - to enable scene change detection by incorporating a customized persistence filtering as in \cite{davidrosen}. Instead of detecting scene changes such as the removal or addition of objects in the environment, the method is limited to detecting changes in individual map points. Also, the authors conducted experiments only in small environments. The algorithm's performance in complex urban environments is unknown. 

We have conducted tests with SceneChangeDet \cite{guo2018learning} - a monocular change detection neural network. The method failed to detect long-lasting changes as traffic signs and only worked in cases of short-lasting changes such as the presence of certain cars in a parking slot (see Fig. \ref{fig:sub1} and Fig. \ref{fig:sub2}), which is naturally undesirable for our purposes. Compared to the previously mentioned methods on change detection\cite{alcantarilla2018street, palazzolo2017change}, our pipeline requires only a sparse point cloud of the initial state of the environment and excludes the need for complex arrangements of cameras as in \cite{palazzolo2017change}. Moreover, differently from \cite{davidrosen} and \cite{hashemifar}, it detects changes of concrete environmental structures - such as traffic signs - and are specifically designed for complex urban environments. Unlike \cite{guo2018learning}, the detected changes consist solely of long-lasting modifications of the environment. Furthermore, SceneChangeDet \cite{guo2018learning} only focuses on detecting changes without localizing them. 

Zhang \textit{et al.} \cite{zhangchangedetection} proposed the fusion of SLAM-based algorithms with semantic segmentation to generate a semantic point cloud. To detect changes in the environment, the authors proposed to denoise, cluster, and vectorize the point cloud before matching the semantic point clouds from the initial state with that of the current state. A recursive Bayesian depth filter combined with a camera pose estimation from motion sensors (IMU) is also utilized to obtain the 3D positions of points in the point cloud. Since the lateral and height distances are not considered in their work, the estimations of these 3D positions can be negatively affected. Moreover, since instance segmentation is not utilized, it is not possible to detect changes such as the modification of the meaning of a traffic sign present in a certain region.

He \textit{et al.} \cite{diffnet} introduced an end-to-end deep neural network solution - named Diff-Net - for change detection from 2D images. Their approach works by projecting HD map elements - i.e. traffic signs - onto the camera pose creating a rasterized image with such elements. The rasterized and the original images are utilized as inputs to the neural network to infer map changes. However, their method assumes that the camera poses are known a priori. Also, the authors focused on detecting changes in the 2D images. Therefore, their method refrains from localizing changes in the 3D environment. Similarly, in TransCD \cite{transcd}, Wang \textit{et al.} proposed a transformer-based scene detection algorithm to spot changes in pairs of images. Also like DiffNet \cite{diffnet}, changes are not localized. Santos \textit{et al.} \cite{scdmultiscale} approached the problem of detecting changes in pairs of images with a multiscale convolutional neural network architecture.

Yew and Lee  \cite{cityscalecd} proposed a method for change detection by comparing point clouds created from SfM. Since geo-registering the point clouds before comparing their points would result in large errors due to inaccurate geolocation information and possible drifts introduced by the SfM, the authors proposed a deep learning-based non-rigid registration that allowed them to compare the point clouds more accurately. As mentioned in the Introduction, in a crowdsourced visual data case, it is too costly to create point clouds for each new data collection. Therefore, the proposed method by Yew and Lee \cite{cityscalecd} is not appropriate in crowdsourcing. Our method, on the other hand, does not require the reconstruction of the environment each time it is scanned for changes.

 Jo \textit{et al.} \cite{slamcu} created a change detection and localization algorithm based on SLAM and utilizing the Dempster–Shafer evidence theory. The authors reported a detection accuracy above 90\% and sub-meter localization accuracy. However, their system requires the utilization of burdensome additional devices such as RTK positioning system, LiDAR, IMU, wheel speed sensor, steering angle sensor, and radar. Therefore, its use for crowdsourcing is impractical. Our method only requires the utilization of a camera and a GPS device, which can often be found included in common commercial dashboard cameras.

\subsection{Semantic Mapping} 

McCormac \textit{et al.} \cite{semanticfusion} fused semantic information into dense point clouds of indoor environments created with SLAM (simultaneous localization and mapping) algorithms. The authors employ a deconvolutional semantic segmentation network architecture that provides pixel-wise class predictions. Similar to our work, these predictions are projected into the point cloud utilizing the tracked camera poses provided by the SLAM algorithm. Their approach requires RGB-D image sequences, whereas in our work RGB images suffice.

Rosinol \textit{et al.} \cite{kimera} created Kimera, an open-source C++ library for real-time SLAM with semantic information. Kimera uses mono, stereo, and inertial data to generate a semantic and metric dense reconstruction of the environment by incorporating off-the-shelf tools for 2D semantic segmentation of images. The authors only presented results for simulated indoor environments. Therefore, the performance of the system in large-scale outdoor environments is unclear.

Previous works \cite{paz2020probabilistic, maturana2018real} have tried to combine LIDAR point clouds with semantic segmentation on 2D images for detecting and locating landmarks in a 3D environment. In \cite{paz2020probabilistic}, probabilistic methods were used to construct semantic high-definition multi-layer maps. We apply a similar approach to reduce the manual efforts on map data generation. Instead of combining LIDAR and RGB images, our system only requires input from RGB cameras and focuses on an under-explored scenario: automatic change detection.

Nakajima \textit{et al.} \cite{Nakajima2018} focused on enabling real-time incremental semantic point cloud creation at the same time as providing accurate results. Their approach assigns class probabilities to entire portions of the point cloud instead of to each individual surfel, this notably reduces time complexity. In our approach, since the metadata creation is executed offline, there isn't the need for real-time semantic segmentation of point clouds. 

\noindent
\subsection{Monocular Depth Estimation} 

The Monodepth2 method \cite{9009796} estimates depth from a sequence of RGB images. It is a self-supervised training method that is possible to be fine-tuned without a labeled dataset.  The performance of the model trained on the Kitti dataset \cite{geiger2012we} was unsatisfactory since the boundaries of the estimated depth were blurred (see Figure \ref{fig:monodepth2}). The performance after training the model on our datasets has not demonstrated satisfactory results either (see Figure \ref{fig:monodepth3}). 

Lee \textit{et al.} \cite{lee2019big} designed a neural network architecture based on the encoder-decoder scheme to perform depth estimation from monocular images. In their architecture, based on the locally planar assumption, the authors proposed a novel layer - named local planar guidance (LPG) - located in the decoder block of the network. The experiments have shown that their method outperforms previous ones with a significant margin in diverse metrics, providing state-of-the-art results. 

The limitation of such monocular depth estimation methods is that they do not support 3D localization including the height and lateral information. In this paper, we propose an end-to-end 3D localization network to solve this problem. Since \cite{lee2019big} has shown satisfactory results of depth estimation, we modified it by extending the neural network architecture to support lateral and height estimations.

\section{Discussion} \label{discussion}

In this section, we discuss the limitations and potential improvements for future work organized into 4 main topics: crowdsourcing and data augmentation, geo-registration, road topology, and real-time performance.

\textbf{Crowdsourcing and data augmentation.} Our pipeline is composed of several individual components that must work in synergy for accurate change detection results. Each of the components is prone to errors that accumulate through the pipeline affecting negatively the final results. At the beginning of the pipeline, an accurate point cloud generation at Step A.1 requires multiple views of the same objects with sufficient overlap between the views and possibly at different distances to the objects. This requirement is especially difficult to fulfill when reconstructing large-scale environments since the vehicle being is restricted to follow the road, thus generating images from limited viewpoints and distances. In environments with the presence of large buildings, the camera on the vehicle - regardless of its position - is only able to capture part of the building - most likely a plane wall - which is insufficient for an accurate visual-based feature matching. In some situations, even ultra-wide cameras may not be able to capture the scene with sufficient characteristics for a satisfactory feature matching. As future work, we plan to include crowdsourced data from cars, pedestrians and cyclists to increase the number of different perspectives, thus improving the point cloud generation. In addition, with crowdsourcing, we consider including voting from different observers of the objects in the scene that will be weighted to determine if the change at a particular timestamp is present.

The object detection, semantic segmentation, and pixel-wise object localization methods are powered by deep learning, which are data-hungry algorithms. Therefore, the lack of rich datasets available for their training directly impacts their performance. Both semantic segmentation and object detection neural networks were trained on Mapillary and, even though it is undoubtedly the most complete dataset for our purposes, its creators \cite{8237796} point that it is still insufficient to train an end-to-end neural network and requires some extra tuning. To improve on this, large synthetic datasets can be employed together with domain adaptation \cite{cyclegan}. As for the case of the pixel-wise object localization, since it is trained with data directly extracted from the point cloud generated at Step A.1, for a better generalization other solutions than increasing the number of views with overlapping regions include: 1) training monocular neural network observing a scene from multiple perspectives for better generalization; 2) having an accurate geo-registration since the scale of the SfM model directly affects the scale of the localization predictions in meters; 3) and having a larger amount of reconstructed point cloud data, by mapping larger area of the environment and by densifying the point cloud.

\textbf{Geo-registration} is also a part of the 3D reconstruction and is essential for accurate change detection. It consists of two steps. First, conversion from geodetic coordinates to Cartesian ones is performed. This procedure inevitably introduces errors - especially when the geodetic height is unspecified. Then, a similarity transformation is executed whose parameters are defined such as to minimize the alignment error between the real world and the model’s coordinates. This alignment error can grow higher in large reconstructed areas with limited viewpoints. To alleviate this problem, we plan to perform multiple geo-registration procedures, each of which is executed independently of the other for a segment of the model. Also, the inclusion of the topology of the region can be utilized to improve the conversion between geodetic and Cartesian coordinates. 

\textbf{Road topology.} It has been challenging to detect a change in case a traffic sign has been moved within a short distance to another location without changing the facing direction. This is because the current design of Step B.3 sets a distance threshold and assumes that two traffic signs with the same type detected on different days but close enough to each other (i.e. below the distance threshold) are considered to be identical. In case a sign is shifted from one side of a road to the other, adding the road boundaries and markings to the attributes of metadata may help solve the problem. 

\textbf{Deployment at the edge of Internet.} We plan to deploy the system in a distributed manner that the initial point clouds are created in the cloud while the change detection and map update are conducted at the edge of Internet, such as the computing nodes co-located with cellular base stations or road side units. By moving computation closer to vehicles, the amount of traffic going through the core network would drop, and more importantly, the transmission latency would decrease, which could help reduce the delay of change detection. 

\section{Conclusions} \label{conclusion}

In this paper, we presented a system for creating and updating a multi-layer map for autonomous driving. Our system is partially built on top of other existing methods, e.g., SfM, semantic segmentation, and object detection. Nonetheless, the system brings new functionalities and addresses a number of challenges to enable crowdsourced-based change detection and localization in rapidly changing urban environments. Our solution is able to spot changes in the environment with accuracy above 85\% by analyzing the current state of the environment with its previous one having traffic signs as the objects of interest. The results could be further improved in the future by increasing the performance of background technologies in use.

\section*{Acknowledgments} 
This work was supported in part by the European Union’s Horizon 2020 Research and Innovation Programme under Grant 825496; in part by the Academy of Finland under Grant 317432 and Grant 318937. The authors would like to thank Kari Tammi and Risto Ojala at Aalto University for the provided equipment for data collection. (Corresponding author: Yu Xiao)

\bibliography{mainwiley}

\end{document}